\documentclass{svjour3}                     

\smartqed 
\usepackage[utf8]{inputenc}
\usepackage{amsmath}
\usepackage{amsfonts}
\usepackage{amssymb}
\usepackage{listings}
\usepackage{mathrsfs}
\usepackage{times}
\usepackage{float}
\usepackage{color}
\usepackage{setspace}
\usepackage{color,soul}

\usepackage{amsmath,amsfonts,amsthm,eucal, bm}
\usepackage{enumerate}
\usepackage[letterpaper,margin=3cm]{geometry}
\usepackage{float}
\usepackage[title]{appendix}
\usepackage{color}
\usepackage{comment}
\usepackage{multirow}
\usepackage{caption}
\usepackage{enumitem}

\usepackage[
pdfstartview=XYZ,
bookmarks=true,
colorlinks=true,
linkcolor=blue,
urlcolor=blue,
citecolor=blue,
pdftex,
bookmarks=true,
linktocpage=true,   
hyperindex=true
]{hyperref}

\usepackage{lipsum}
\usepackage{lineno}
\usepackage[FIGBOTCAP,TABTOPCAP,bf,tight]{subfigure}

\usepackage[numbers, sort&compress]{natbib}

\usepackage{graphicx}
\usepackage{pstool}
\usepackage{psfrag}
\usepackage{epstopdf}
\usepackage{booktabs} 
\usepackage{lineno}
\usepackage{import}
\usepackage[export]{adjustbox}

\usepackage{longtable}

\graphicspath{{figures/}}

\newcommand\gbf[1]{\hbox{\boldmath${#1}$\unboldmath}}

\newcommand{\Ds}{\displaystyle}

\newcommand{\FR}[0]{{F\!R}}

\newcommand{\Ve}{{\mathbf{e}}}
\newcommand{\Vv}{{\mathbf{v}}}
\newcommand{\Vu}{{\mathbf{u}}}

\newcommand{\TK}{{\mathbf{K}}}

\usepackage{algorithm}
\usepackage[noend]{algpseudocode}

\newcommand{\Vxi}  {\gbf{\xi}}

\journalname{My-journal}
\begin{document}

\title{Deep learning-aided inverse design of porous metamaterials}

\author{Phu Thien Nguyen$^{1}$
        \and
        Yousef Heider$^{1}$
        \and
        Dennis M. Kochmann$^{2}$
        \and
        Fadi Aldakheel$^{1,\ast}$
}

\institute{$^{\ast}$ Corresponding author: Fadi Aldakheel.\\ 
\email{fadi.aldakheel@ibnm.uni-hannover.de}\\[2mm]
$^1$Institute of Mechanics and Computational Mechanics (IBNM), Leibniz University Hannover, Appelstr. 9A, 30167 Hannover, Germany\\[1mm]
{$^2$Mechanics \& Materials Laboratory, Department of Mechanical and Process Engineering, ETH Zürich, 8092 Zürich, Switzerland}
}

\authorrunning{Nguyen et al.}

\date{Received: date / Accepted: date}

\maketitle

\begin{abstract}
The ultimate aim of the study is to explore the inverse design of porous metamaterials using a deep learning-based generative framework. Specifically, we develop a property-variational autoencoder (pVAE), a variational autoencoder (VAE) augmented with a regressor, to generate structured metamaterials with tailored hydraulic properties, such as porosity and permeability. 
While this work uses the lattice Boltzmann method (LBM) to generate intrinsic permeability tensor data for limited porous microstructures, a convolutional neural network (CNN) is trained using a bottom-up approach to predict effective hydraulic properties. This significantly reduces the computational cost compared to direct LBM simulations.
The pVAE framework is trained on two datasets: a synthetic dataset of artificial porous microstructures and CT-scan images of volume elements from real open-cell foams. The encoder-decoder architecture of the VAE captures key microstructural features, mapping them into a compact and interpretable latent space for efficient structure-property exploration. The study provides a detailed analysis and interpretation of the latent space, demonstrating its role in structure-property mapping, interpolation, and inverse design. This approach facilitates the generation of new metamaterials with desired properties. The datasets and codes used in this study will be made open-access to support further research.

 
\end{abstract}

\begin{center}
  \fbox{ 
    \begin{minipage}{0.9\textwidth}
    \textbf{Abbreviations} 
    \begin{longtable}{ll ll}
    ANN & Artificial Neural Network & BGK & Bhatnagar-Gross-Krook \\
    CNN & Convolutional Neural Network & CT & Computed Tomography \\
    DNN & Deep Neural Network & ELBO & Evidence Lower Bound \\
    FFNN & Feed-Forward Neural Network & GAN & Generative Adversarial Network \\
    KDE & Kernel Density Estimation & KL & Kullback–Leibler \\
    LBM & Lattice Boltzmann Method & MLP & Multi-Layer Perceptron \\
    ML & Machine Learning & MSE & Mean Squared Error \\
    PNM & Pore Network Model & PCA & Principal Component Analysis \\
    pVAE & Property-Variational Autoencoder & RNN & Recurrent Neural Network \\
    RVE & Representative Volume Element & SGD & Stochastic Gradient Descent \\
    {\it slerp} & Spherical Linear Interpolation & SVE & Stochastic Volume Elements \\
    TPM & Theory of Porous Media & VAE & Variational Autoencoder \\
    \end{longtable}
    \end{minipage}
  }
\end{center}
\section{Introduction}
\label{intro}
Porous materials are gaining increasing attention in research and industry due to their unique properties and wide-ranging applications. Their high surface area, tunable porosity and permeability (as seen in porous metamaterials), and lightweight nature make them indispensable for filtration, catalysis, energy storage, and biomedical engineering. Recent advancements in computational modeling and additive manufacturing have further enhanced the ability to precisely design and optimize these materials, paving the way for innovative and highly efficient applications across multiple fields. 
To better understand the behavior of multiphase heterogeneous porous materials, it is crucial to incorporate microscopic information into macroscopic modeling, as the effective macroscopic properties are strongly influenced by the material’s microstructural characteristics. However, applying classical multiscale techniques directly is often impractical due to their high computational costs. The rapid advancements in artificial intelligence, particularly data-driven models, offer a promising solution to accelerating these computational processes, enabling more efficient and scalable material analysis \cite{Aldakheel_2023_CNN_Hetero, chaaban2024CNN, ltknguyen2018DD, HeiderHSSuh2020_ML_offline, NabipourEtAl2025_DataDriven, YANGEtAl2024_DataDrivenMethods_Review}.
%
%
%
Regarding the integration of machine learning (ML) in the analysis of multiphase porous microstructures, three-dimensional (3D) volumetric images have become an essential data source. Represented as a 3D voxel grid, these images provide detailed insights into internal material structures and enable the accurate derivation of physical properties across various industries \cite{johan2024diffusion}. Recently, there has been growing interest in using diffusion models to reconstruct, generate, and mimic natural porous structures, as demonstrated in \cite{chun2020syntheticgeneration, kumar2024MD,phong2022syntheticgeneration, johan2024diffusion, xianrui2024deepgenerative}.
Considering these developments, there is a growing emphasis on material inverse design within the field of material discovery. This approach focuses on unraveling the complex interrelationships between processing methods, material structures, and resultant properties, guiding the development of advanced materials with tailored functionalities \cite{weichen2022datadesign, xianrui2024deepgenerative}. 

Addressing the inverse problem of identifying microstructure topologies that satisfy specific material property requirements has traditionally relied on inefficient trial-and-error strategies. These methods often require the developer to have an intuitive understanding of the complex relationship between structure and properties. Even today, some of the most innovative biomimetic metamaterials with specialized properties are designed using interactive, trial-and-error-based exploration of structure-property maps \cite{plessis2019biodesign}. 
Moreover, while the forward problem, i.e., mapping topological features to the property space, is well defined, the inverse problem is inherently ill-posed, see, e.g., \citet{kumar2020inverse}. This means that multiple topologies can correspond to a single set of material properties, leading to non-uniqueness in the solution space.
In this context, \citet{chakraborti2004GA} introduced the use of genetic algorithms for material design, while \citet{sigmund2003TO, sigmund2013TO} proposed topology optimization as a systematic approach. However, both methods are computationally demanding \cite{kumar2020inverse}. Given these challenges, ML has emerged as an efficient alternative.

Deep generative models, such as generative adversarial networks (GANs) \cite{goodfellow2014GAN} and variational autoencoders (VAEs) \cite{kingma2013VAE, kingma2019VAE}, have attracted significant attention within the materials science community. 
Both GANs and VAEs aim to approximate complex, high-dimensional data by learning a structured latent space from which new samples can be generated. GANs typically excel in producing high-fidelity outputs but lack a well-defined, continuous latent space, making their representations difficult to interpret and control \cite{Goodfellow-et-al-2016}. In contrast, VAEs offer a distinct advantage for tasks that require a coherent and meaningful latent space, enabling smoother and more interpretable mappings.
In this context, \citet{phong2022syntheticgeneration} introduced a novel deep learning method that synthesizes realistic 3D microstructures with controlled structure-property relationships, utilizing a combination of GANs and actor-critic (AC) reinforcement learning, achieving a $5\%$ error margin relative to target values. 
Within design optimization in the automotive industry, the work of \cite{Borse_Stoffel_2024_GANs_RL} uses GANs to expand the computationally expensive datasets. This allows the implementation of reinforcement learning (RL) to automate the exploration of the design parameter.
On a different front, VAEs have been employed to explore design space more deeply by obtaining a low latent space parameterization. This approach has also sparked interest in integrating constitutive physical models into neural network architectures. For instance, \citet{liweiwang2020vae} introduced an innovative data-driven framework for metamaterial design, leveraging deep generative modeling to address the challenges posed by high-dimensional topological design spaces, multiple local optima, and significant computational costs. In their approach, a VAE and a property-prediction regressor are jointly trained on a large metamaterial database to map complex microstructures into a low-dimensional, continuous, and structured latent space. Notably, the latent space captures a distance metric that reflects shape similarity, enabling interpolation between different microstructures and encoding meaningful patterns of geometric and property variation. This allows complex mappings between topology and mechanical properties to be efficiently navigated through simple vector operations within the latent space. 
\citet{lizheng2023trussinv} proposed a generative framework for the inverse design of truss metamaterials. Their approach leverages a low-dimensional latent space derived from a VAE, which allows for a detailed exploration of the mapping between topology and mechanical properties. Their latent space not only encodes complex structural relationships but also facilitates a more intuitive and efficient search for optimal truss designs, addressing the inverse problem by revealing deeper insights into the structure-property relationships of metamaterials.


In the context of porous materials characterized by complex and non-periodic 3D architectures, microstructures resembling those formed via spinodal decomposition have been used as surrogate data for inverse design
\cite{vidyasagar2018spin, soyarslan2018spin, mengting2019spin, kumar2020inverse, Rassloff2025, lizheng2021spin, magnus2022inverse, kumar2024MD}.
Their smooth, non-intersecting surfaces with near-zero mean curvature and unique topological properties make them ideal candidates for such applications \cite{kumar2020inverse}.
For instance, \citet{magnus2022inverse} introduced an inverse design methodology focused on two-phase, anisotropic, random microstructures with tailored diffusivity in multiple directions. By integrating convolutional neural network (CNN) predictions within an approximate Bayesian computation framework, the microstructure design process becomes computationally efficient, allowing for the specification of diffusivity across all three spatial directions. In the efficient design of structures that consider microstructural effects, a significant challenge arises from the impracticality of performing direct numerical simulations (DNS) on the entire microstructure of a component. This difficulty is largely attributed to finite-size effects and spatial correlations, while classical micromechanics primarily focuses on the limitations of average or mean properties. To address these challenges, \citet{JONES2024multiscalelatent} propose a multiscale simulation methodology that integrates the spatial correlations inherent in the underlying microstructure, thus mitigating some of the finite-size effects that complicate the homogenization process in this context. The objective of this approach is to facilitate the design of components that account for the interplay between macroscopic and microstructural scales, particularly under conditions of microstructural uncertainty, by enhancing the forward propagation of such uncertainties. In this framework, a property-Variational Autoencoder (pVAE) is employed to establish the relationship between microstructure and properties of the material. Through an intermediate structure-to-latent encoding, the pVAE provides a compact representation of microstructural features relevant to property predictions. Moreover, the combination of principal component analysis (PCA) with VAEs suggests the plausible existence of a representative latent space, where the structure-to-property mapping demonstrates sufficient regularity. This regularity ensures effective sampling of spatial property correlations driven by structural patterns and inter-property relationships. 

A common limitation of the data-based pVAE framework is the necessity of a structured and sufficiently large database, particularly when aiming to create a continuous and meaningful design space. To address this challenge, several efforts have been made. Notably, \citet{ZHANG2024_DA-VEGAN} introduced a hybrid VAE-GAN model for reconstructing material microstructures from small datasets. A key innovation of this work is the use of $\beta$-VAEs \cite{higgins2017betaVAE}, which facilitate smooth and interpretable latent spaces. Additionally, the application of differentiable data augmentation techniques enables high-quality microstructure generation even with limited data. This approach represents a significant advancement in computational materials science, facilitating data-scarce applications in materials design. \citet{JONES2024multiscalelatent} employed $\beta$-VAEs in conjunction with transfer learning to further reduce the reliance on large numerical datasets. 

\begin{figure}[hbt]
\begin{center}
\includegraphics[width=14.0cm]{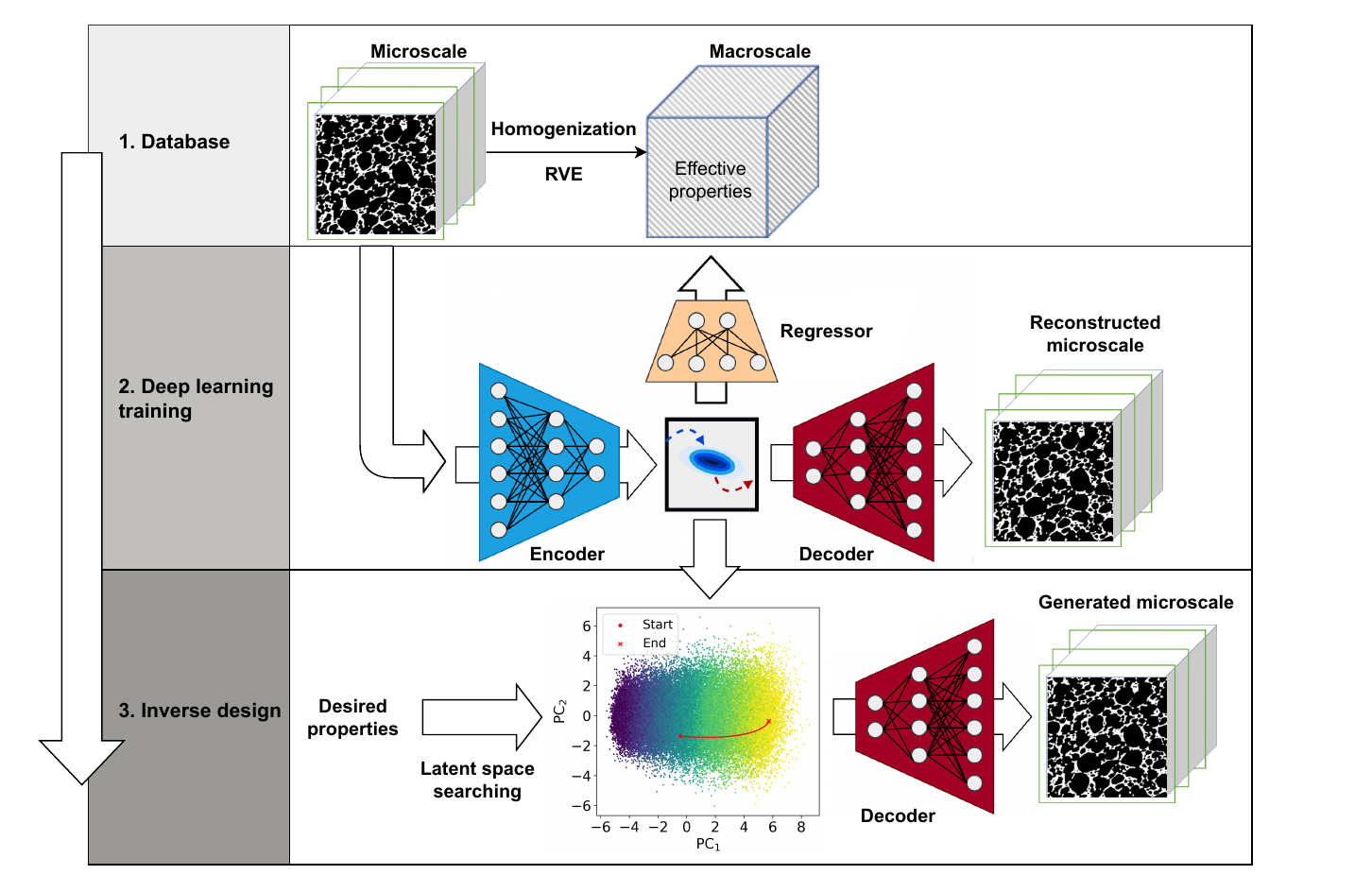}
\end{center}
\captionsetup{skip=-2pt}
\caption{Illustration of the proposed data-driven inverse design framework. It integrates microscale structure generation, macroscopic property evaluation, and deep generative modeling for porous metamaterials. Effective properties, such as porosity $n^F$ and intrinsic permeability $\mathbf{K}^{S}$, are computed via homogenization using the LBM approach. A pVAE is trained to learn a compact latent representation that supports both structure reconstruction and property prediction. This latent space is then exploited to perform inverse design by generating microstructures that satisfy predefined property criteria.}
\label{Fig:graphical}
\end{figure}

The presented contribution utilizes pVAE to encode data related to porous materials into a continuous, lower-dimensional latent space, offering a structured framework for capturing and managing microstructural information in multiscale design applications.
Two distinct datasets are employed: (1) a synthetic porous material dataset with uni-directional flow, generated using Python, serving as a proof of concept of the pVAE framework and to illustrate how microstructural complexity affects the interpretability of the latent space; and (2) CT-scan images of real open-cell foams, providing a more realistic representation of porous structures. The study focuses on porosity and fluid permeability as key properties, guiding the latent space construction to enable efficient structure-property mapping and material design optimization.
In particular, Figure~\ref{Fig:graphical} illustrates the complete workflow used in this study, which comprises three major steps:
(1) microstructural data generation and effective property evaluation;
(2) a deep learning framework using the pVAE; and
(3) the inverse design step.
The process begins by computing the effective hydraulic properties of porous microstructures. These structure-property pairs are then used to train the pVAE model, which captures the relationship between microstructure and macroscopic behavior and learns a compact latent representation that encodes key morphological features. This latent space enables microstructure reconstruction and property prediction. Finally, gradient-based optimization is used to generate new microstructures that meet the desired effective property targets.

%


To provide an overview of the manuscript, Section~\ref{sec:TPM} presents an abstract representation of the effective porous media properties. Section~\ref{sec:DLModel} gives an outline of the ML frameworks used in this research, including a surrogate CNN model for intrinsic permeability computation, pVAE framework, and inverse design approach.
Section~\ref{sec:pVAESynthetic} presents the first pVAE model, applied to synthetic data. This includes database preparation, CNN-based intrinsic permeability prediction, pVAE model evaluation, interpolation in the latent space, and gradient-based inverse design in the latent space.
Section~\ref{sec:pVAErealFoam} presents the second pVAE application in connection with real $\mu$-CT microstructures of open-foam material. This section also includes an evaluation of the trained pVAE model and the inverse design in the presence of a small database.
This is followed in Section~\ref{sec:Conclusions} by
concluding remarks and a discussion of future aspects.
\section{Macroscopic aspects of porous media mechanics} 
\label{sec:TPM}


The macroscopic effective hydraulic properties of porous materials play a vital role in the inverse design process as they govern the material’s behavior within the porous media framework. To discuss these properties within the framework of the Theory of Porous Media (TPM), we consider a fully saturated porous material that consists of a single pore fluid and a compressible solid matrix. The macroscopic TPM framework \cite{Ehlers2002, EhlersWagner2019}  is usually employed to model microscopically heterogeneous porous materials on the large scale.
Within this framework, homogenization is applied to a representative volume element (RVE) composed of distinct constituents $\phi^{\alpha}$, where $\alpha \in \{S, F\}$ denotes the solid and fluid phases, respectively. Each macroscopic material point is assumed to be simultaneously occupied by statistically distributed, interacting, and superimposed solid and fluid continua, such that $\phi = \phi^{S} \cup \phi^{F}$. 
For each phase $\phi^{\alpha}$, the volume fraction $n^{\alpha}:=dv^{\alpha}/dv$ is defined as the ratio of the partial volume $dv^{\alpha}$ to the total volume element $dv$ of the RVE, where $0<n^{\alpha}<1$\,. A fully saturated condition is enforced through the saturation constraint:
\begin{equation}
    \sum_{\alpha} n^{\alpha} = n^{S} + n^{F} = 1, \quad \text{with }
    \begin{cases}
    n^{S} & \text{: solid volume fraction (solidity)} \\
    n^{F} & \text{: fluid volume fraction (porosity)\,.}
    \end{cases}
    \label{eq:saturation}
\end{equation}

In porous media flow modeling, Darcy’s law is a widely adopted constitutive relation for describing fluid flow under fully saturated conditions. It is particularly applicable to incompressible, steady-state, and laminar flow regimes. Under these assumptions, the fluid velocity is primarily driven by the pressure gradient, while inertial and frictional effects are negligible \cite{Maea10}. 
In its simplified form, Darcy’s law defines a linear relationship between the seepage velocity $\mathbf{w}^{F}$ and the pressure gradient $\nabla p$, expressed as:
\begin{equation}
    \nabla p = -\mu^{F} \left( \mathbf{K}^{S} \right)^{-1} \mathbf{w}^{F}.
    \label{eq:darcy}
\end{equation}
Here, $\mu^{F}$ denotes the dynamic viscosity of the fluid, and $\mathbf{K}^{S}$ is the intrinsic permeability tensor, which is a key material property that quantifies the porous medium’s ability to transmit fluids independently of the fluid viscosity. It also reflects geometric characteristics of the porous microstructure, such as pore size, shape, and connectivity. Consequently, $\mathbf{K}^{S}$ can be directly linked to the micromorphology of the porous medium, as captured by techniques like $\mu$-CT. Further details and references on the TPM and homogenization framework can be found in, e.g., \cite{markert2007constitutive, chaaban2020upscaling, PhuEtAl2023_PAMM, chaaban2024CNN,ALDAKHEEL2020102517}.

In this work, we relate the pVAE approach to two key effective properties, i.e. \(n^F\) and \(\mathbf{K}^{S}\). In this, \(n^F\) is directly computed from the binary image representation of the microstructure, as the ratio of total void volume to the total volume. In contrast, evaluating \(\mathbf{K}^{S}\) is computationally more demanding. Darcy’s law is employed as an inverse technique to estimate \(\mathbf{K}^{S}\), based on data obtained from lattice Boltzmann method (LBM) simulations performed on mesoscale RVEs. Specifically, the geometry of each porous microstructure serves as input to a single-phase LBM simulation, as detailed in Appendix~\ref{appx:OneFluid_LBMTheor}.
In particular, the objective of the LBM simulations is to compute the average fluid velocity in response to a prescribed pressure gradient applied across the porous domain. Simulations are performed independently along the three principal directions, \(\mathbf{x}_1\), \(\mathbf{x}_2\), and \(\mathbf{x}_3\), corresponding to pressure gradients \(\nabla p_1\), \(\nabla p_2\), and \(\nabla p_3\), respectively. Based on the resulting velocity fields, the intrinsic permeability tensor \(\mathbf{K}^{S}\), expressed in lattice units \(\big[\mathrm{l.u.}\big]\), is calculated for each representative volume element (RVE). Additional computational details are provided in Appendix~\ref{appx:InrinsicPermComputation} and in~\cite{chaaban2020upscaling, PhuEtAl2023_PAMM, chaaban2024CNN}.

For large datasets, performing LBM simulations on all samples becomes computationally impractical. To address this, and inspired by the work of~\citet{HeiderDakheelEhlers2024CNN}, a surrogate ML model based on a 3D convolutional neural network (3D-CNN) is employed to predict the intrinsic permeability efficiently, as will be explained in later sections. 

\section{Deep learning frameworks}
\label{sec:DLModel}
This section presents details of the two deep learning (DL) frameworks, i.e., CNN and pVAE models, considered in this work and implemented in Python 3.11.8 using Keras \cite{chollet2015keras} with TensorFlow 2.15.0 \cite{abadi2016tensorflow} as the backend. Both implementation and training are done in a Jupyter Notebook environment under Anaconda. Training is performed using a NVIDIA H100 GPU with 80 GB of memory. 
The first subsection covers the application of a CNN model for predicting the intrinsic permeability \(\mathbf{K}^{S}\) of porous materials. The CNN architecture and optimization procedures are briefly described, with further details available in \cite{Aldakheel_2023_CNN_Hetero, HeiderDakheelEhlers2024CNN}. Once trained and validated, the CNN enables a reliable and efficient generation of a large dataset for training the pVAE model.
The second subsection introduces the core concepts, architecture, and extension techniques of the pVAE model \cite{liweiwang2020vae}, which is established as a general framework for {\it structure-property mapping}. Specifically, the pVAE model processes 3D pixel tensors constructed from stacked artificial binary images or $\mu$-CT images. It maps these inputs to a low-dimensional, continuous latent space, allowing both reconstruction of the 3D pixel tensors and prediction of porosity and intrinsic permeability components. 
In connection with this, the third subsection discusses the inverse optimization neural network framework, which allows the reconstruction of microstructures that satisfy desired effective properties.
\subsection{Surrogate CNN model for intrinsic permeability computation}
\label{subsec:CNN}
\begin{figure}[!ht]
\begin{center}
\includegraphics[width=14.0cm]{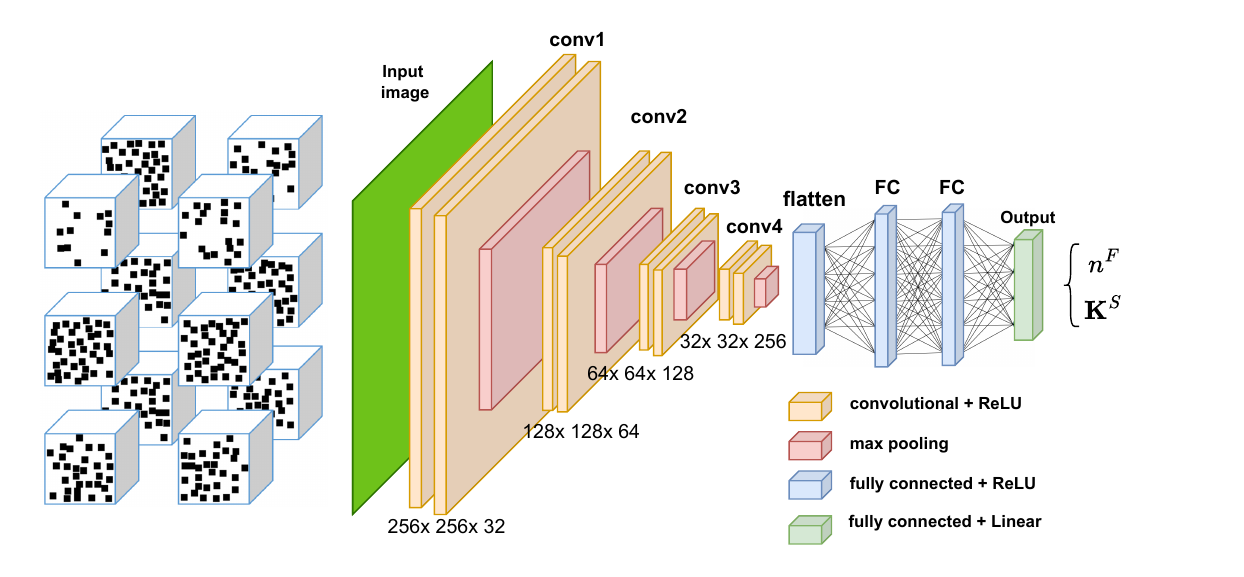}
\end{center}
\caption{Illustration of the 3D CNN architecture and its information flow, including input data, convolutional layers, pooling operations, flattening, and fully connected Multi-Layer Perceptron (MLP) layers. This model predicts the prosity \(n^{F}\) and intrinsic permeability tensor \(\mathbf{K}^{S}\), which include only one component (\(K^{S}_{11}\)) for the uni-directional flow.}
\label{Fig:CNN_StrucIni}
\end{figure}

CNN excels at processing grid-structured data, making it highly effective for image classification, object recognition, and text analysis. More recently, they have been applied to predict effective properties from microscale image data, expanding their role in materials science and engineering. For detailed reviews and applications, refer to \cite{GU2018CNN, EIDEL2023_CNN, Aldakheel_2023_CNN_Magnetostatics, Dhillon_Verma_2020_reviewCNN, Aldakheel_2023_CNN_Hetero, Tandale_2023_CNN_RNN, BisharaEtAl2023_Review_ML_Multiscale, Tragoudas2024}.
In this study, the CNN model is adapted from \cite{Aldakheel_2023_CNN_Magnetostatics, HeiderDakheelEhlers2024CNN}, specifically optimized for predicting hydraulic properties. 
An illustration of the 3D CNN architecture is given in Figure~\ref{Fig:CNN_StrucIni}.

The network architecture in Figure~\ref{Fig:CNN_StrucIni}, which is illustrated for uni-directional flow, consists of the following key components:
%
\noindent
\begin{itemize}
    \item[$\bullet$] Data Preparation: Input images are reshaped for compatibility with the 3D CNN architecture. Data indices are shuffled, and the output properties ($n^F$, $K^{S}_{11}$) are normalized using MinMaxScaler \cite{pedregosa2011scikit}.
    \item[$\bullet$] Model Architecture: The CNN comprises four convolutional blocks with increasing filter sizes (32 to 256) and kernel sizes (3$\times$3$\times$3 to 7$\times$7$\times$7), using \textit{ReLU} activation and \textit{MaxPooling3D} for spatial down-sampling. Fully connected layers follow, with 64 and 32 units, and the final layer outputs two regression targets with linear activation.
\item[$\bullet$] \textbf{Optimization and loss function:} The Adam optimizer is employed with a learning rate of $0.00001$. During training, the CNN model minimizes the Mean Squared Error (MSE) loss, defined as  
\begin{equation}\label{eq:LossMSE}
    \mathcal{D}_\text{MSE} = \frac{1}{m} \sum_{i=1}^{m} \left[ \left(n^{F,p}_i - n^{F,t}_i\right)^2 + \left(K^{S,p}_{11,i} - K^{S,t}_{11,i}\right)^2 \right]\,,
\end{equation}
where $m$ denotes the number of output data points.
    \item[$\bullet$] Training: Training is conducted for up to 175 epochs with a batch size of 16. Callbacks, including \textit{ReduceLROnPlateau}, \textit{ModelCheckpoint}, and \textit{EarlyStopping}, are implemented for performance monitoring.
\end{itemize}
The size of the input 3D images can be adjusted without altering the architecture of the proposed model. For more detailed information, refer to \cite{HeiderDakheelEhlers2024CNN}.
Note that the CNN model is applied exclusively to unidirectional flow in synthetic porous materials. For real porous materials, the LBM approach was used due to the limited size of the available dataset.

\subsection{Property-variational autoencoder (pVAE)}
\label{subsec:pVAEArch}
\begin{figure}[hbt]
\begin{center}
\includegraphics[width=12.0cm]{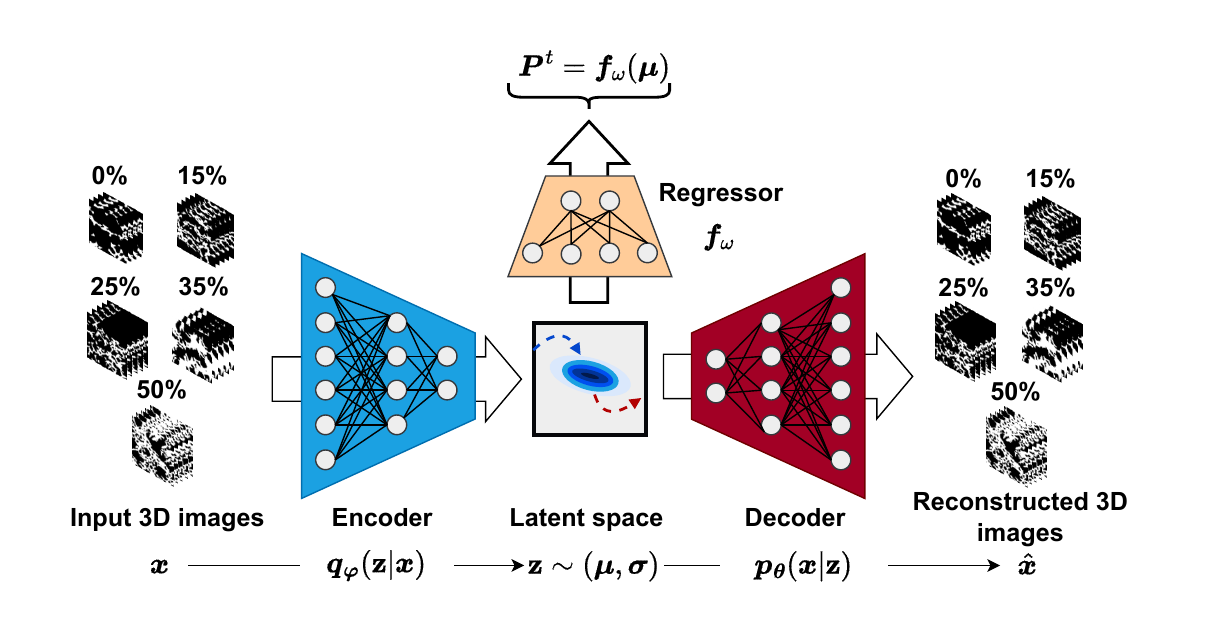}
\end{center}
\captionsetup{skip=-12pt}
\caption{Schematic of the property-variational autoencoder (pVAE). The encoder maps input 3D microstructures of porous material $\boldsymbol{x}$ to a latent space $\boldsymbol{z} \sim (\boldsymbol{\mu}, \boldsymbol{\sigma})$. A regressor predicts target properties $\boldsymbol{P}^t$, here, \{$n^F$, $\mathbf{K}^{S}$\}, from $\boldsymbol{\mu}$, while the decoder reconstructs porous media microstructures $\boldsymbol{\hat{x}}$, cf. \cite{lizheng2023trussinv,JONES2024multiscalelatent}.}
\label{Fig:3DCVAE_FM}
\end{figure}

Aiming to map complex microstructures into a lower-dimensional latent space, \citet{liweiwang2020vae} introduced a data-driven framework for the inverse design of metamaterial microstructures. This approach leverages a VAE combined with a FFNN regressor to connect the latent space with the material’s effective properties. This approach, later referred to as pVAE by \citet{JONES2024multiscalelatent}, utilizes the VAE’s ability to generate a compact representation of high-dimensional data while enabling efficient sampling from the latent space. 
Specifically, the framework consists of a parametric encoder and a decoder, trained simultaneously to optimize a variational lower bound on the data likelihood. Enforcing a structured latent representation enhances the scalability and interpretability of metamaterial design. The learned latent space provides a continuous and structured representation, where similar microstructures are mapped to nearby points, facilitating smooth interpolation and exploration of new designs.
Thus, each input microstructure, here, binarized images of synthetic and real open-cell porous materials, is represented by a tensor $\boldsymbol{x}$ with an unknown true distribution $p^{*}(\boldsymbol{x})$ (marginal likelihood). Accordingly, the encoder maps $\boldsymbol{x}$ to a continuous distribution in latent space rather than to a discrete set of points that is as close as possible to the true distribution.
%
Thus, in line with \cite{lizheng2023trussinv,JONES2024multiscalelatent}, the proposed pVAE  model architecture is illustrated in Figure~\ref{Fig:3DCVAE_FM}.
%
In detail, the VAE model establishes a stochastic mapping between the observed data space $\boldsymbol{x}$ and a latent representation $\boldsymbol{z}$.
This mapping corresponds to a directed probabilistic model with a joint distribution $p_{\boldsymbol{\theta}}(\boldsymbol{x}, \boldsymbol{z})$, defined over both the observed variable $\boldsymbol{x}$ and the latent variable $\boldsymbol{z}$ as
\begin{equation}
\label{eq:Jointdistribution}
p_{\boldsymbol{\theta}}(\boldsymbol{x},\boldsymbol{z}) = p_{\boldsymbol{\theta}}(\boldsymbol{x}|\boldsymbol{z})p_{\boldsymbol{\theta}}(\boldsymbol{z}),   
\end{equation}
where $\boldsymbol{\theta}$ is the vector of decoder parameters, $p_{\boldsymbol{\theta}}(\boldsymbol{z})$ is the parameterized prior distribution of latent variables and $p_{\boldsymbol{\theta}}(\boldsymbol{x}|\boldsymbol{z})$ is the approximated distribution of $\boldsymbol{x}$ conditioned on $\boldsymbol{z}$. The conditioned distribution $p_{\boldsymbol{\theta}}(\boldsymbol{x}|\boldsymbol{z})$, parameterized by the decoder, provides almost arbitrary flexibility for the marginal distribution $p_{\boldsymbol{\theta}}(\boldsymbol{x})$ with a relatively simple predefined prior distribution $p_{\boldsymbol{\theta}}(\boldsymbol{z})$.  
However, computing the marginal distribution directly, which is essential for obtaining the likelihood function during training, is generally intractable. To address this intractability, the VAE introduces an encoder (inference model) $q_{\boldsymbol{\varphi}}(\boldsymbol{z}|\boldsymbol{x})$ to map $\boldsymbol{x}$ back to the latent vector $\boldsymbol{z}$, serving as an approximation of the posterior distribution $p_{\boldsymbol{\theta}}(\boldsymbol{z}|\boldsymbol{x})$. By coupling the encoder and decoder networks, the VAE provides an explicit representation of the likelihood function, which is then approximated during training using the evidence lower bound (ELBO), i.e.
\begin{equation}
\label{eq:ELBO}
\mathcal{L}(\boldsymbol{\theta}, \boldsymbol{\varphi}; \boldsymbol{x}) = \mathbb{E}_{q_{\boldsymbol{\varphi}}(\boldsymbol{z}|\boldsymbol{x})} \left[ \log \frac{p_{\boldsymbol{\theta}}(\boldsymbol{x}, \boldsymbol{z})}{q_{\boldsymbol{\varphi}}(\boldsymbol{z}|\boldsymbol{x})} \right] \\
= 
\underbrace{
\mathbb{E}_{q_{\boldsymbol{\varphi}}(\boldsymbol{z}|\boldsymbol{x})} \left[ \log p_{\boldsymbol{\theta}}(\boldsymbol{x}|\boldsymbol{z}) \right] }_{\text{(a)}}
- 
\underbrace{
\mathbb{E}_{q_{\boldsymbol{\varphi}}(\boldsymbol{z}|\boldsymbol{x})} \left[ \log \frac{q_{\boldsymbol{\varphi}}(\boldsymbol{z}|\boldsymbol{x})}{p_{\boldsymbol{\theta}}(\boldsymbol{z}|\boldsymbol{x})} \right]}_{\text{(b)}}.
\end{equation}
Here $\mathbb{E}$ is the expectation operator and (\ref{eq:ELBO})$_\text{(a)}$ represents the reconstruction loss, which helps to accurately reconstruct the input data. 
The term (\ref{eq:ELBO})$_\text{(b)}$ represents the Kullback-Leibler (KL) divergence, alternatively expressed as $D_{\mathrm{KL}} \left[ q_{\boldsymbol{\varphi}}(\boldsymbol{z}|\boldsymbol{x}) \, \| \, p_{\boldsymbol{\theta}}(\boldsymbol{z}|\boldsymbol{x}) \right]$. This acts as a regularizer by forcing the latent distribution $q_{\boldsymbol{\varphi}}(\boldsymbol{z}|\boldsymbol{x})$ to stay close to the prior $p_{\boldsymbol{\theta}}(\boldsymbol{z})$.
The ELBO enables the use of the efficient stochastic gradient descent method (SGD) for the simultaneous training of the encoder and decoder. Specifically, VAE assumes $p_{\boldsymbol{\theta}}(\boldsymbol{z}) \sim N(0,\boldsymbol{I})$ and adopts a Gaussian distribution for approximated posterior distribution as
\begin{equation}
\label{eq:Posteriordistribution}
q_{\boldsymbol{\varphi}}(\boldsymbol{z},\boldsymbol{x}) = N(\boldsymbol{\mu},\boldsymbol{\sigma})\,.   
\end{equation}
Here, $\boldsymbol{\mu}$ (mean vector) and $\boldsymbol{\sigma}$ (standard deviation) are predicted by the encoder. By assigning $\boldsymbol{z} = \boldsymbol{\mu} + \boldsymbol{\sigma} \odot \boldsymbol{\varepsilon}$, where $\boldsymbol{\varepsilon} \sim N(0, \boldsymbol{I})$ and $\odot$ denotes the element-wise product (i.e., an operation performed between two vectors or matrices of the same dimensions, where each element of one is multiplied by the corresponding element of the other), the usual Monte Carlo estimator for ELBO used in SGD is reduced to
\begin{equation}
\label{eq:ELBO2}
\begin{aligned}
\mathcal{L}(\boldsymbol{\theta}, \boldsymbol{\varphi}; \boldsymbol{x}) 
&
= \mathbb{E}_{q_{\boldsymbol{\varphi}}(\boldsymbol{z}|\boldsymbol{x})} \left[ \log p_{\boldsymbol{\theta}}(\boldsymbol{x}|\boldsymbol{z}) \right] 
- 
D_{\mathrm{KL}} \left[ q_{\boldsymbol{\varphi}}(\boldsymbol{z}|\boldsymbol{x}) \, \| \, p_{\boldsymbol{\theta}}(\boldsymbol{z}) \right] \\
&
= 
\underbrace{\frac{1}{L} \sum_{l=1}^{L} \log p_{\boldsymbol{\theta}}(\boldsymbol{x}|\boldsymbol{z})}_{\text{(a)}} 
- 
\underbrace{\frac{1}{2} \sum_{j=1}^{J} \left( 1 + \log(\sigma_{j}^{2}) - \sigma_{j}^{2} - \mu_{j}^{2} \right)}_{\text{(b)}}.
\end{aligned}
\end{equation}
Here, $J$ and $L$ are the dimensions of the latent space and number of samples, respectively. The symbols $\mu_{j}$ and $\sigma_{j}$ represent, respectively, the mean and standard deviation of the $j$th element of a latent vector with $J$ dimensions, while $l$ denotes the $l$th sample out of $L$ samples.
Therefore, we can optimize the parameters of the encoder and decoder by performing SGD to solve:
\begin{equation}
\label{eq:minLoss}
\min_{\boldsymbol{\theta}, \boldsymbol{\varphi}} \big[-\mathcal{L}(\boldsymbol{\theta}, \boldsymbol{\varphi}; \boldsymbol{x})\big]\,.
\end{equation}
Relying solely on reconstruction loss (\ref{eq:ELBO2})$_\text{(a)}$ during training leads the model to overfit by memorizing the training data, effectively reducing it to a classical autoencoder. This results in "dead zones" in the latent space, i.e., regions the decoder cannot map to realistic samples. To address this, the VAE incorporates a sampling process during training, introducing random noise into the latent variables. This promotes better generalization and adds a regularization term based on KL divergence to the loss function. The KL divergence enforces normalization, encouraging the encoder to map neighboring regions of the latent space to similar microstructures. Consequently, the encoder is forced to create a continuous and semantically meaningful latent space. 
This property facilitates precise control of complex geometries and supports an efficient, data-driven design framework for metamaterial microstructures and multiscale systems.
%
%
%

For porous media metamaterial design, it is crucial to establish a structure-property mapping that links the latent space to the effective hydraulic properties, i.e., the permeability and porosity in this work. To achieve this, the property-variational autoencoder (pVAE) introduces a third component, a regressor network, that maps the latent variables $\boldsymbol{\mu}$ to the effective properties of the structure, defined as $\boldsymbol{P}^{t} = [n^{F}, K_{ii}^{S}], \, i = 1, 2, 3$, through the function $\boldsymbol{f}_{\boldsymbol{{\omega}}}$, where $\boldsymbol{P}^{t}$ denotes the true effective properties of the porous microstructure and $w$ represents the parameters of the regressor network.
The complete architecture is illustrated in Figure~\ref{Fig:3DCVAE_FM}. The number of predicted properties in the mapping can be adjusted depending on the specific design objectives. To enable the simultaneous learning of both geometrical features and effective material properties, the optimization process of the VAE is extended to include a regression loss term. This modification allows the model to be trained jointly for accurate microstructure reconstruction and reliable property prediction, formulated as:
\begin{equation}
\label{eq:pVAELoss}
\boldsymbol{\theta}, \boldsymbol{\varphi}, \boldsymbol{{\omega}} \leftarrow \min_{\boldsymbol{\theta}, \boldsymbol{\varphi}, \boldsymbol{{\omega}}} \bigg[ -\mathcal{L}(\boldsymbol{\theta}, \boldsymbol{\varphi}; \boldsymbol{x}) 
+ \big\| \boldsymbol{P}^{t} - \boldsymbol{f}_{\boldsymbol{\omega}}(\boldsymbol{\mu}) \big\|^{2} \bigg]\,.
\end{equation}
We design the encoder and decoder using convolutional layers, with the dimension of the latent space tested across various values to balance low dimensionality and generation quality. To mitigate the impact of random noise introduced by the reparameterization trick, the regressor operates solely on the mean value $\boldsymbol{\mu}$, while the decoder uses latent variables sampled from the approximated posterior distribution. 
A key challenge in this setting arises from the difficulty of accurately mapping porous microstructure geometries, which are represented as pixelated tensors and are sensitive to inherent noise. To address this, the trade-off among the three primary tasks of the pVAE, i.e. geometry reconstruction, latent space regularization, and property prediction, can be effectively managed using the $\beta$-VAE framework \cite{higgins2017betaVAE}. In this framework, a hyperparameter $\beta$ is introduced to the KL divergence term in (\ref{eq:ELBO}), allowing its contribution to the total loss to be adjusted. By increasing $\beta$, the pVAE places greater emphasis on regularization, thereby improving generalization.

Building on this idea, \citet{JONES2024multiscalelatent} introduced an additional parameter, $\lambda$, in the regression term of (\ref{eq:pVAELoss}), ensuring that the predictor more accurately captures the relationship between latent representations and microstructure properties. Meanwhile, \citet{lizheng2023trussinv} proposed incorporating a parameter into the reconstruction loss term of (\ref{eq:ELBO}) to prioritize accurate reconstruction in truss architectures.
In this study, due to the inherent heterogeneity of the porous microstructures, the approach of \citet{JONES2024multiscalelatent} is adopted, as it effectively guides the latent space to better support property prediction. 
The resulting final loss function of the pVAE can be expressed as 
\begin{equation}
\label{eq:finalpVAEloss}
\begin{aligned}
\mathcal{L}_\text{pVAE}(\boldsymbol{\theta}, \boldsymbol{\varphi}; \boldsymbol{x}) 
&= \mathbb{E}_{q_{\boldsymbol{\varphi}}(\boldsymbol{z}|\boldsymbol{x})} \left[ \log p_{\boldsymbol{\theta}}(\boldsymbol{x}|\boldsymbol{z}) \right] 
- \beta D_{\mathrm{KL}} \left[ q_{\boldsymbol{\varphi}}(\boldsymbol{z}|\boldsymbol{x}) \, \| \, p_{\boldsymbol{\theta}}(\boldsymbol{z}) \right] + \lambda \big\| \boldsymbol{P}^{t} - \boldsymbol{f}_{\boldsymbol{w}}(\boldsymbol{\mu}) \big\|^{2}.
\end{aligned}
\end{equation}

%
Table~\ref{tab:archit_pVAE} presents the different parameters of the pVAE model. The latent space dimension is set to $N_\ell = 250$, providing a balance between effective dimensionality reduction and the preservation of critical information in the bottleneck, which is essential for high-quality reconstruction. The package Optuna~\cite{akiba2019optuna} is used to determine an appropriate latent dimension, performing 50 trials with 50 epochs each. $N_{\boldsymbol{P}}$ is the number of output properties for regressor mapping from the latent space.
\begin{table}[h!]
\centering
\scriptsize 
\begin{adjustbox}{width=\textwidth, center}
\begin{tabular}{lclclclclcl}
\toprule
\multicolumn{3}{c}{\textbf{Encoder}} & \multicolumn{2}{c}{\textbf{Latent}} & \multicolumn{3}{c}{\textbf{Decoder}} & \multicolumn{3}{c}{\textbf{Regressor}} \\
\cmidrule(lr){1-3} \cmidrule(lr){4-5} \cmidrule(lr){6-8} \cmidrule(lr){9-11}
\textbf{Layer} & \textbf{Output} & \textbf{Activation} & \textbf{Layer} & \textbf{Output} & \textbf{Layer} & \textbf{Output} & \textbf{Activation} & \textbf{Layer} & \textbf{Output} & \textbf{Activation} \\
\midrule
Input   & $100^3 \times 1$  &      & Dense          & $ 2 \times N_\ell$ & Input           & $N_\ell$          &              & Input & $N_\ell$             &      \\
Conv3D  & $50^3 \times 48$  & ReLU & Sampling layer & $N_\ell$           & Dense           & $128000$          & ReLU         & Dense & 16                   & ReLU \\
Conv3D  & $25^3 \times 48$  & ReLU &                &                    & Reshape         & $10^3 \times 128$ &              & Dense & 16                   & ReLU \\
Conv3D  & $22^3 \times 192$ & ReLU &                &                    & Conv3DTranspose & $22^3 \times 128$ & ReLU         & Dense & $4$                  & ReLU \\
Conv3D  & $10^3 \times 128$ & ReLU &                &                    & Conv3DTranspose & $25^3 \times 192$ & ReLU         & Dense & $N_{\boldsymbol{P}}$ &      \\
Flatten & $128000$          &      &                &                    & Conv3DTranspose & $50^3 \times 48$  & ReLU         &       &                      &      \\
        &                   &      &                &                    & Conv3DTranspose & $100^3 \times 48$ & ReLU         &       &                      &      \\
        &                   &      &                &                    & Conv3DTranspose & $100^3 \times 1$  & SteepSigmoid &       &                      &      \\
\bottomrule
\end{tabular}
\end{adjustbox}
\caption{Details of the pVAE model architecture, showing the layer configurations and activation functions of the encoder, latent space, decoder, and regressor.}
\label{tab:archit_pVAE}
\end{table}
The number of layers in the encoder-decoder architecture was determined through extensive experiments involving a separate training of VAEs, with the number of neurons optimized using Optuna \cite{akiba2019optuna}. To efficiently downsample the input 3D pixelated tensor, strides were applied within the 3D convolutional layers, eliminating the need for pooling layers. The hidden layers employed the Rectified Linear Unit (ReLU) activation function.
%
The output layer of the decoder employs a $\mathrm{SteepSigmoid}$ function, defined as
\begin{equation}
\label{eq:SteepSig}
\mathrm{SteepSigmoid}(x; k) = \frac{1}{1 + e^{-kx}}, \quad k > 0\, .
\end{equation}
The decoder is designed to reconstruct a 3D binary pixelated tensor, with voxel values approximating 0 or 1. To promote this binarization, the $\mathrm{SteepSigmoid}$ function with $k = 5$ is applied to the decoder's output layer, producing sharper transitions between low and high values. Additionally, the regressor's output layers utilize the ReLU activation function to ensure non-negative predictions of material properties.

The model employs the Adam optimizer, an adaptive variant of stochastic gradient descent (SGD), along with a learning rate scheduler that reduces the learning rate when the validation loss plateaus over a specified number of epochs. Initially, the regressor, composed of linear dense layers, is trained using only the mean vector $\boldsymbol{\mu}$ from the latent space, while keeping both the encoder and decoder parameters frozen. This pre-training stage enables the regressor to effectively learn the mapping from latent representations to target properties without interference from latent space updates.

Simultaneously, the VAE is trained independently with a small learning rate (\(1.2 \times 10^{-4}\)) to ensure stable convergence, especially given the depth and complexity of the 3D convolutional architecture. Joint training from the beginning proved suboptimal due to the conflicting learning dynamics, as the regressor benefits from a higher learning rate for faster convergence, while the VAE requires a significantly smaller rate to maintain latent space stability. Pre-training each component separately allows for appropriate tuning of these learning rates and provides well-initialized weights. After this stage, joint fine-tuning of the full model is performed to further improve the coupling between the latent representation and property prediction.
Once satisfactory accuracy is achieved from the VAEs, transfer learning integrates the encoder-decoder architecture with the regressor to map structural inputs to target properties. During this phase, the decoder parameters remain frozen to maintain consistency with the pre-trained VAE loss, while the regressor parameters are optimized using the MSE loss to enhance property prediction accuracy. This strategy ensures that the decoder adheres to the VAE objective, while the regressor effectively learns the structure-property relationships.
To prevent overfitting in the regressor, early stopping is applied based on validation loss, monitored using a checkpoint callback to save the best-performing model during training. The dataset is efficiently loaded using a pipeline implemented with the TensorFlow library, addressing memory capacity limitations. A batch size of 16 is employed to ensure computational efficiency. The dataset is divided into training, validation, and test sets, with proportions of 80\%, 10\%, and 10\%, respectively, facilitating effective model training and evaluation.

\subsection{Inverse design optimization framework}
\label{subsec:InvDesignFramework}
\begin{figure}[!ht]
\begin{center}
\includegraphics[width=11.0cm]{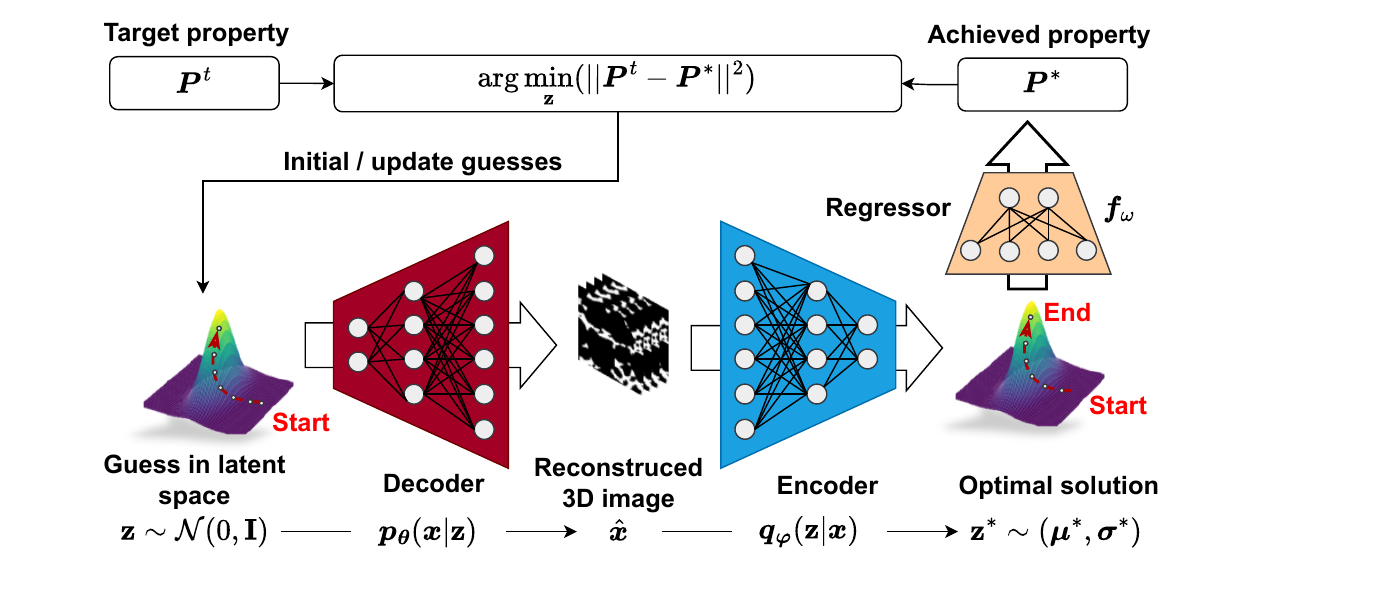}
\end{center}
\caption{A schematic representation of an inverse design framework, where the latent space gradient-based optimization strategy iteratively refines porous media microstructure to align with desired target hydraulic properties.}
\label{Fig:invers_design_framework}
\end{figure}

The continuous latent space generated by the pVAE provides a base for gradient-based optimization to design microstructures with desired properties and to explore regions beyond the initial training domain. Solving the inverse problem in our work is challenging due to the inherent one-to-many relationship, where multiple porous microstructures can correspond to similar effective hydraulic properties. 
In the search for a candidate porous microstructure with properties such as permeability close to the desired values, this work implements a systematic approach as introduced by \citet{lizheng2023trussinv}. 
As shown in Figure~\ref{Fig:invers_design_framework}, the inverse design process begins with an initial guess in the latent space, which is iteratively refined through optimization. A decoder reconstructs microstructures from latent representations, while an encoder extracts latent variables to guide further refinement. A property predictor evaluates the achieved properties, ensuring convergence towards microstructures that align with target characteristics while maintaining physical realism.
%


%

In this work, the generative framework is tested by designing microstructures with random  values for effective permeability and porosity as the target properties.
A set of random target values is used to initialize the process. The initial latent guesses are selected from the training dataset based on proximity to the target properties. This selection provides a realistic starting point for the decoder, which ensures credible reconstructions before the encoding step.
Acknowledging the one-to-many nature of the mapping, a reference-based approach is also employed, inspired by \citet{lizheng2023trussinv}. Here, all microstructures in the training dataset are evaluated relative to the target permeability, and the 100 closest matches are identified as initial guesses. These candidates are then optimized using gradient-based techniques, and the solutions are assessed by examining the predicted properties from the property predictor. This step provides a computationally efficient way to evaluate permeability during optimization.
Ultimately, the framework identifies multiple viable microstructure candidates with similar permeability values, offering flexibility in design. These options can then be refined based on additional criteria such as sustainability, weight, or other desired attributes, providing a robust solution for inverse design challenges in porous material development.


\section{Model investigations using synthetic data}
\label{sec:pVAESynthetic}
The performance of the pVAE is analyzed to achieve two primary objectives: (1) to assess its reconstruction accuracy, and (2) to evaluate the regressor's effectiveness in predicting material properties.
Thereafter, we explore the latent space, highlighting its ability to interpolate and its potential to utilize probabilistic integration to generate diverse families of microstructure geometries. Lastly, we apply gradient-based optimization to randomly selected target properties to test the framework’s capability for the inverse design of microstructures.
\subsection{Database preparation}
\label{subsec:Aca_data}
\begin{figure}[!ht]
\begin{center}
\includegraphics[width=11.0cm]{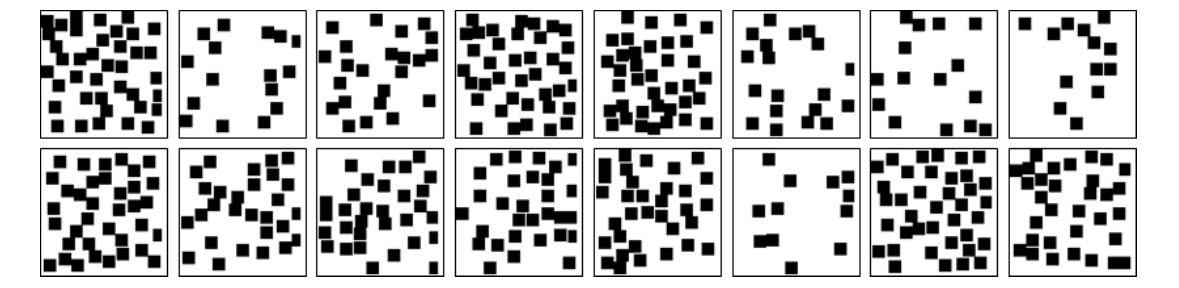}
\end{center}
\caption{Illustration of the synthetic academic dataset, where each 3D sample consists of 100 binary images, each with a resolution of 100×100 pixels.}
\label{Fig:SyntheAcaData}
\end{figure}

This example features a synthetic academic dataset generated using Python. The dataset comprises stacks of $100^2$ binary images, each containing randomly distributed square pores. While these pores are prevented from overlapping, they may share boundaries.
%
With a total of 48,831 3D samples, it provides a comprehensive and reliable foundation for analysis and model development, as illustrated in Figure~\ref{Fig:SyntheAcaData}.
The dataset in this example considers square pores of size 10$^2$ pixels, with the number of pores per image ranging from 10 to 40. Two key random variables are considered: the spatial distribution of pores and the total number of pores, both determined using the \textit{randint} function from the NumPy library \cite{harris2020array}. A constraint is imposed to prevent pore overlapping, i.e., pores may share boundaries but cannot intersect.
Consequently, multiple 3D samples exhibit the same porosity yet differ in pore position. Despite having identical porosities, their hydraulic properties can vary due to the formation of larger, connected pores through boundary sharing. This phenomenon illustrates a common challenge in inverse design, where many samples share the same porosity and permeability values despite geometric differences, complicating structure-property mapping.

Table~\ref{tab:DescrStat_DataFrame} presents descriptive statistics of the synthetic dataset, highlighting the distribution of key parameters. This includes essential statistical measures, such as the mean, standard deviation, minimum, and maximum values, along with quartiles (25\%, 50\%, and 75\%) for each parameter.
\begin{table}[ht]
\centering
{\small
\begin{tabular}{ |p{2.0cm}|c|c|c|c|c|c|c|c| }
\hline
\textbf{Parameter} & \textbf{Count} & \textbf{Mean} & \textbf{Std} & \textbf{Min} & \textbf{25$\%$} & \textbf{50$\%$} & \textbf{75$\%$} & \textbf{Max} \\
\hline
$n^F$ & 48831 & 0.24795 & 0.08894 & 0.1 & 0.17 & 0.25 & 0.32 & 0.4 \\
$K^{S}_{11} \, \big[\mathrm{l.u.}\big]$ & 48831 & 0.98113 & 0.38528 & 0.34291 & 0.64685 & 0.96275 & 1.29225 & 2.02872 \\
\hline
\end{tabular}
\caption{Summary statistics of key parameters in the synthetic dataset, presenting central tendencies and dispersion measures for the analyzed variables.}
\label{tab:DescrStat_DataFrame}
}
\end{table}

\begin{figure}[!ht]
\begin{center}
\includegraphics[width=9.0cm]{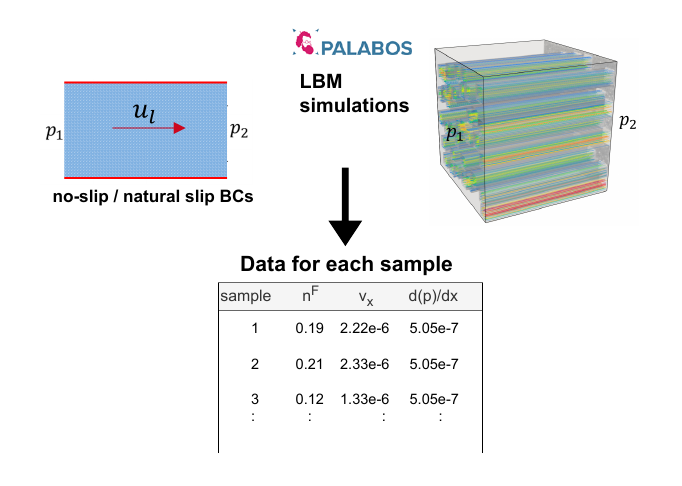}
\quad
\includegraphics[width=5.5cm]{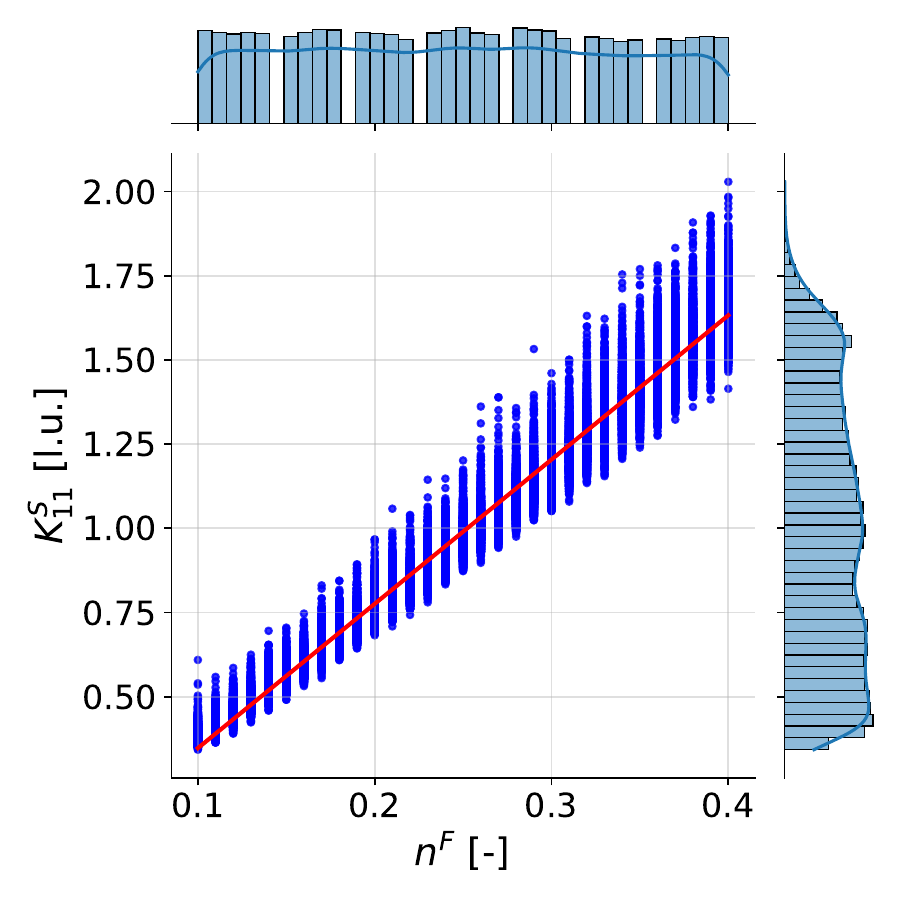}
\end{center}
    \caption{Workflow for permeability computation and analysis using LBM simulations. The left schematic illustrates the pressure-driven flow setup using the LBM. In this setup, a pressure difference $p_1$ and $p_2$ is applied at opposite boundaries of a porous medium, leading to an average velocity field ($u_l$), simulated using the \textbf{PALABOS} package. An example of the simulation results is shown in the table. On the right, the correlation between porosity $n^F$ and permeability $K^S_{11}$ is visualized, as indicated by the red regression line.}
\label{Fig:JointPairwise}
\end{figure}

The synthetic dataset is large enough for the pVAE approach. The values of $n^F$ range from 0.1 to 0.4, and the values of  $K^{S}_{11}$ range from approximately 0.3 to 2.0. The TPM-LBM approach presented in Appendix~\ref{appx:OneFluid_LBMTheor} is applied to this synthetic dataset, where uni-directional flow is considered.
Figure~\ref{Fig:JointPairwise} shows the workflow for permeability computation and analysis using LBM simulations. It also visualizes the regression relationship between porosity and intrinsic permeability. A linear relationship consistent with Darcy's law is evident. Additionally, although the variance does not cover a wide range, porosity values correlate with varying permeability.

\subsection{CNN-based intrinsic permeability prediction}
\label{sec:ReCNN}
\begin{figure}[!ht]
\begin{center}
\includegraphics[width=6.7cm]{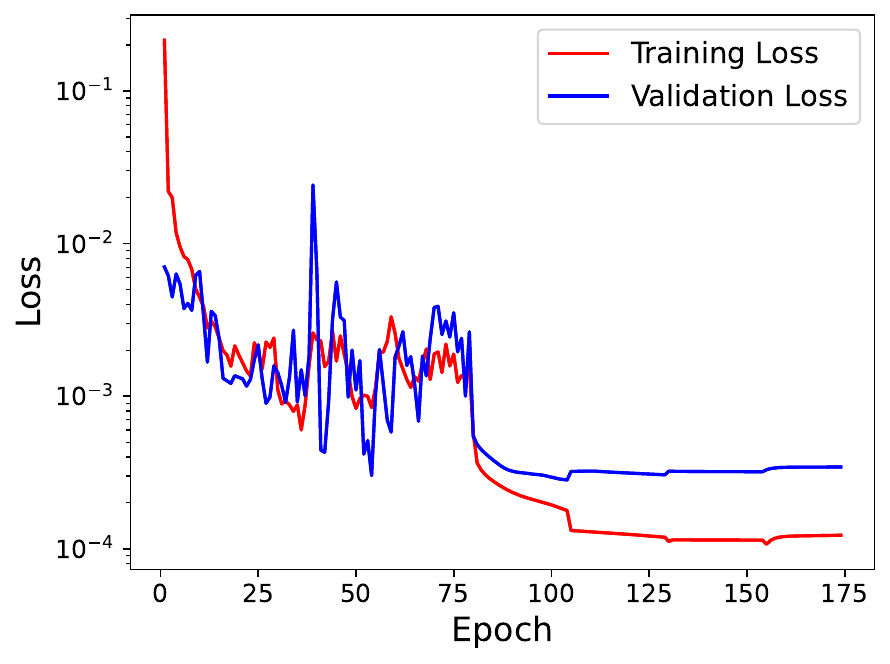}~~~~
\includegraphics[width=6.7cm]{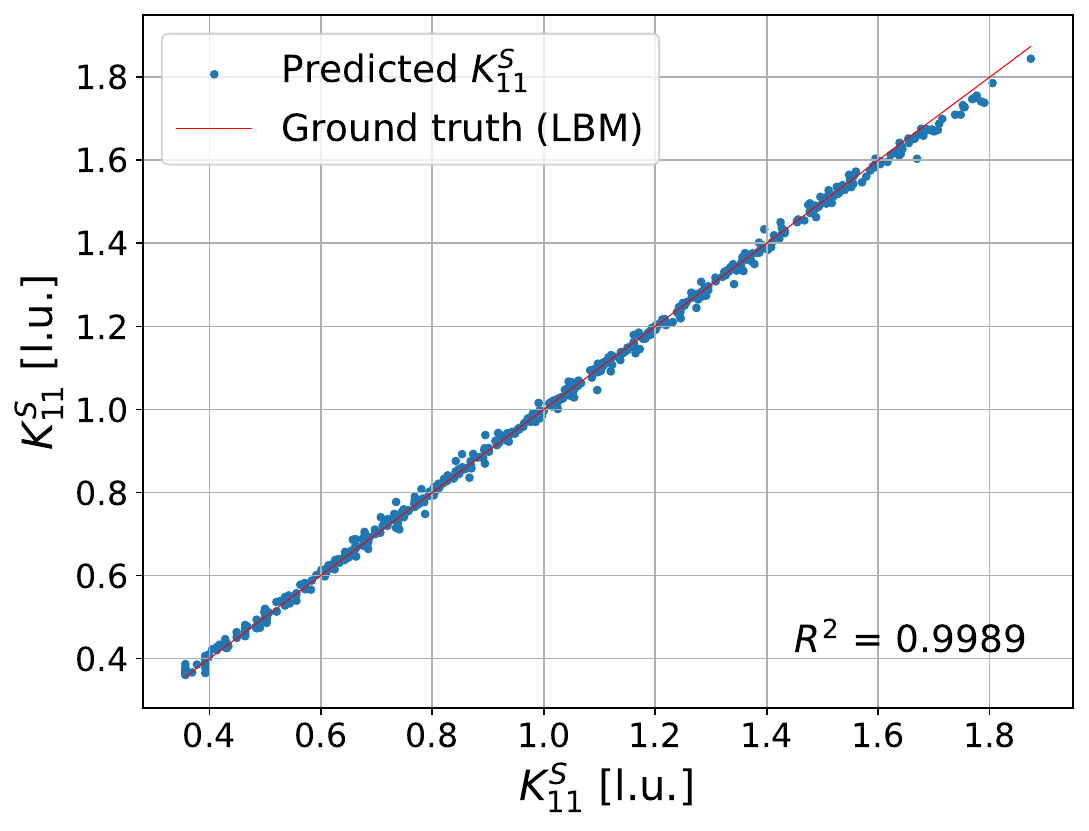}
\end{center}
\caption{Synthetic data: Left: Training and validation loss function values over the number of weight updates (Epoch). The validation loss (blue curve) initially follows a similar trend to that of training loss (red curve) but starts to diverge after approximately 80 epochs, suggesting  overfitting beyond this point. Right: CNN-predictions vs. ground-truth values of the intrinsic permeability component $K^{S}_{11}$ for unseen test dataset.
}
\label{Fig:CNNloss-Model1Pred}
\end{figure}

In the following, we apply the CNN model, as described in Section \ref{subsec:CNN}, to the synthetic dataset to predict the permeability of 3D microstructures with unidirectional flow. This approach generates a sufficiently large dataset to construct a design space for the pVAE method.
The training process, which involves 2800 samples, is monitored by tracking both training and validation losses over multiple epochs, as illustrated in Figure~\ref{Fig:CNNloss-Model1Pred} (left). Initially, both losses decrease rapidly, indicating effective learning. However, as training progresses, the training loss reaches a lower value than the validation loss, signaling the onset of overfitting. Despite various hyperparameter adjustments, overfitting could not be fully mitigated for this dataset. As a result, the model state after approximately 100 epochs, where the loss stabilizes around $5\times10^{-4}$ is considered optimal and used for testing on unseen data.
The scatter plots in Figure~\ref{Fig:CNNloss-Model1Pred} (right) compare the predicted values with the ground truth values for the output variable \(K^{S}_{11}\). This plot illustrates the model’s ability to accurately predict intrinsic permeability values and corresponding porosity.
The model's performance can quantitatively be assessed using the \(R^2\)-score%
\footnote{
The \(R^2\) score, typically between 0 and 1, is calculated using the formula:
\begin{eqnarray}\label{Eq:R2}
R^2 = 1 - \frac{\sum_{i=1}^n (y_i - h_i)^2}{\sum_{i=1}^n (y_i - \bar{y})^2}
\quad\text{with}\quad
\left\{\begin{array}{ll}
y_i & \text{:True value for the \(i\)-th data point}, \\
h_i & \text{:Predicted value for the \(i\)-th data point}, \\
\bar{y} & \text{:Mean of the true values}, \\
n & \text{:Number of data points}. 
\end{array}\right.
\end{eqnarray}
%
}.
The \(R^2\)-score achieved in this model is $\approx\,0.9977$, demonstrating a high level of accuracy in predicting the output variables based on unseen data.
%
%
%
\subsection{pVAE model evaluation}
\label{subsubsec:RepVAE_aca}
As outlined in Section~\ref{sec:DLModel}, the VAE is initially trained, independent of the properties, to ensure accurate reconstruction capabilities. 
Following this, transfer learning is employed within pVAE, where the latent space is mapped to the properties $n^F$ and $K^{S}_{11}$. 
To accommodate the case of synthetic data, in which all cross-sectional slices are identical, the pVAE architecture in  Table~\ref{tab:archit_pVAE} is modified to use 2D convolutional layers instead of 3D ones. The number of layers and the regressor architecture remain unchanged.

%

\begin{figure}[!ht]
\begin{center}
\includegraphics[width=8cm]{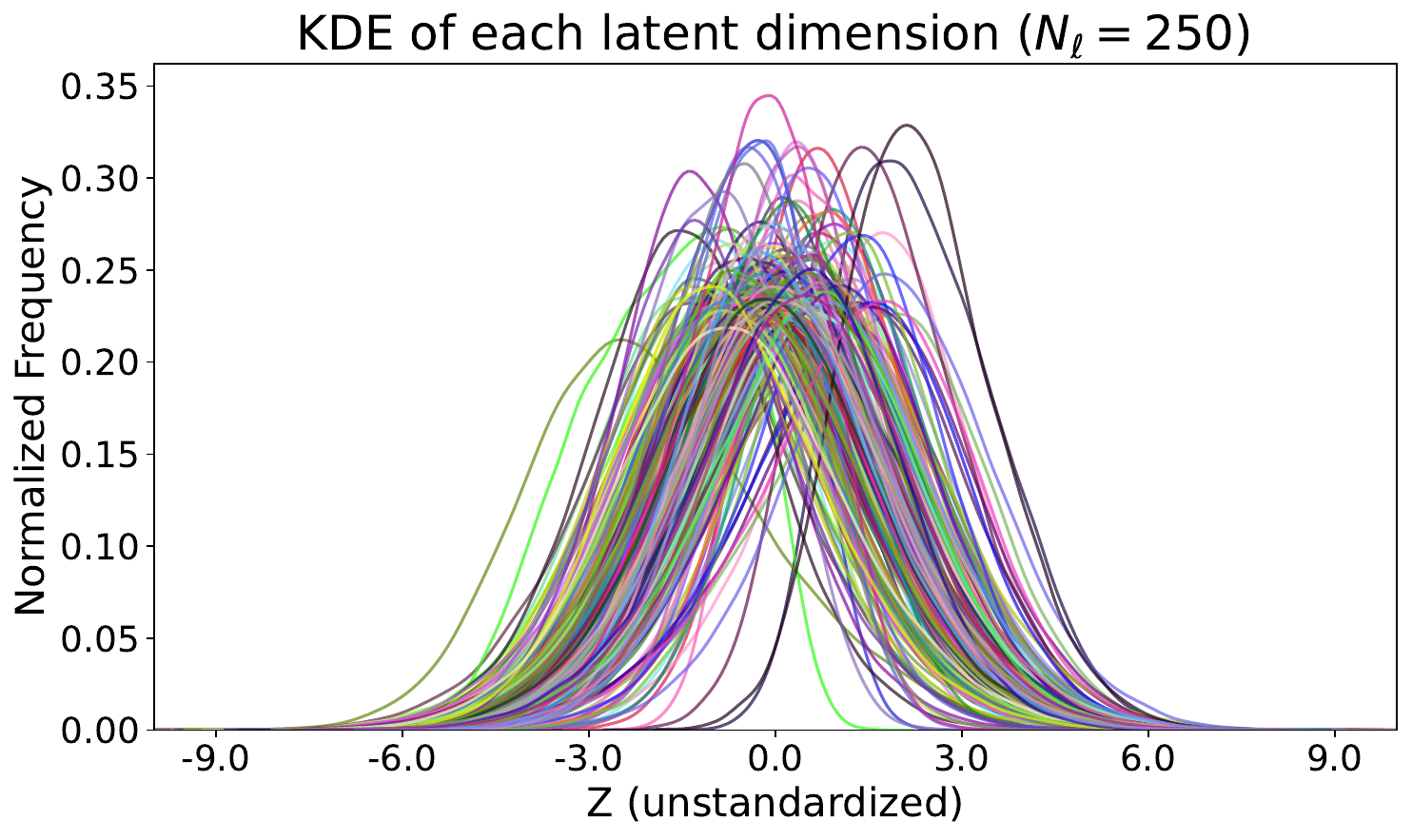}
\end{center}
\caption{Synthetic dataset: 
Kernel Density Estimation (KDE) plots for each of the 250 latent dimensions in the autoencoder, illustrating the distribution of encoded microstructures. Each curve represents the probability density function of a single latent dimension across the dataset, showing variations in spread and shape.
}
\label{Fig:acaKDE}
\end{figure}

%
Figure~\ref{Fig:acaKDE} presents the Kernel Density Estimation (KDE) curves for each of the 250 latent dimensions in the autoencoder, computed from the encoding of 48,831 microstructures. These KDE plots illustrate the distribution of latent values across different dimensions, revealing variations in shape and spread. While most distributions exhibit a near-Gaussian profile with minor deviations in mean and standard deviation, the variational regularizer is designed to enforce normalization and encourage a structured latent space. However, some latent dimensions show asymmetry or deviations from normality, which could be attributed to the high-dimensional encoding, data variability, or latent dependencies reflecting complex microstructural features.
%

\begin{figure}[!ht]
\begin{center}
\includegraphics[width=4.577cm]{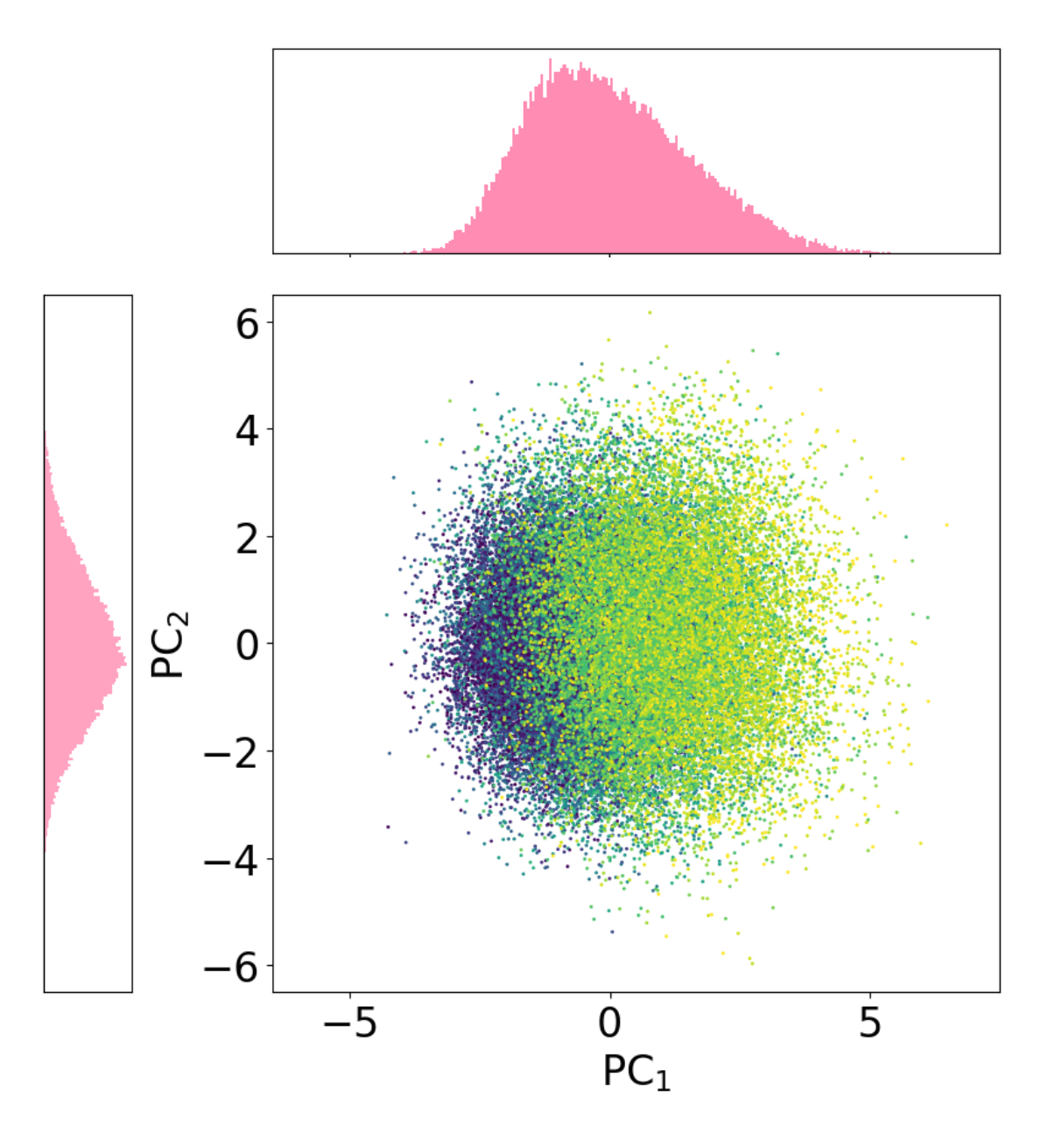}
\includegraphics[width=4.577cm]{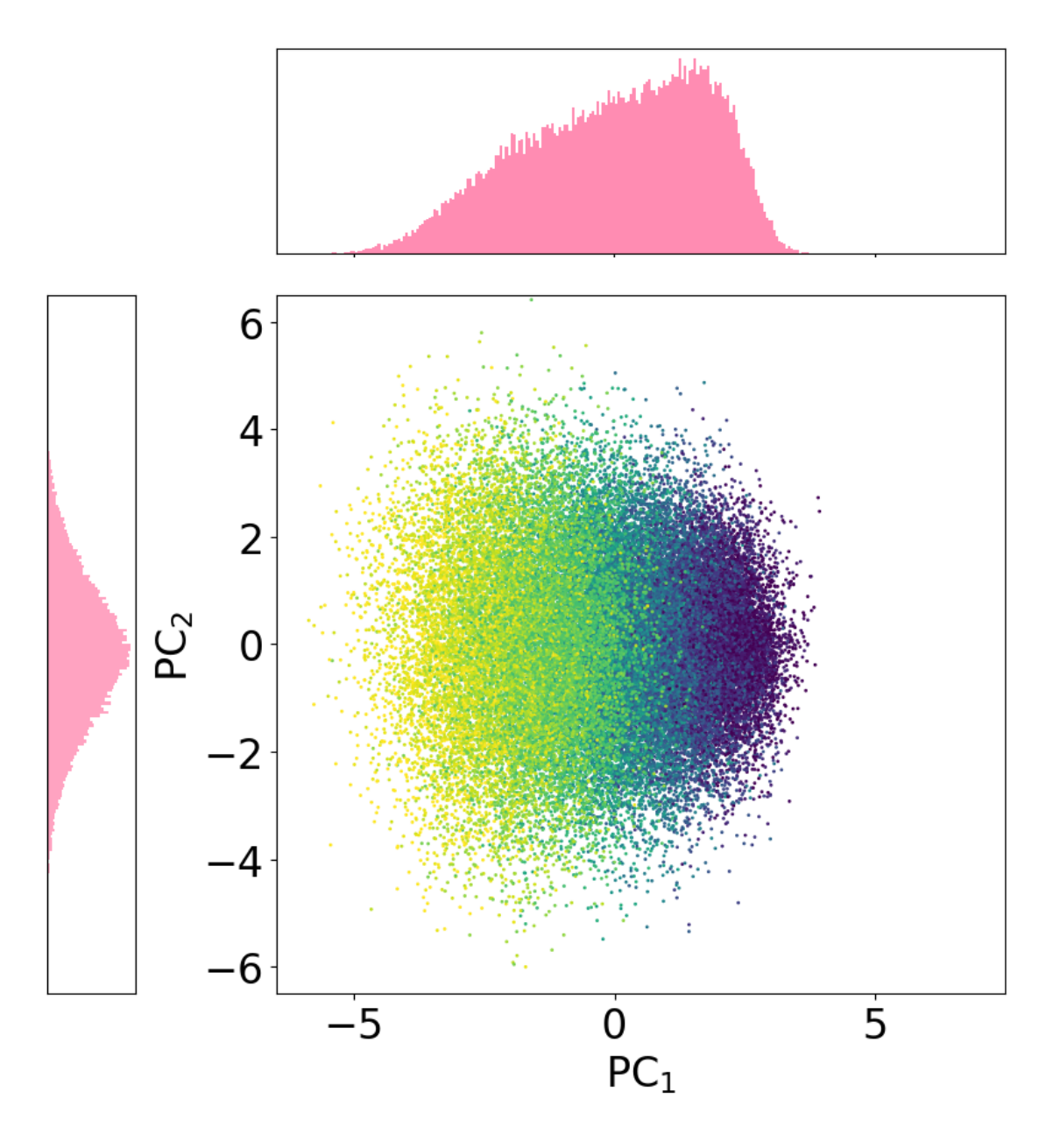}
\includegraphics[width=5.615cm]{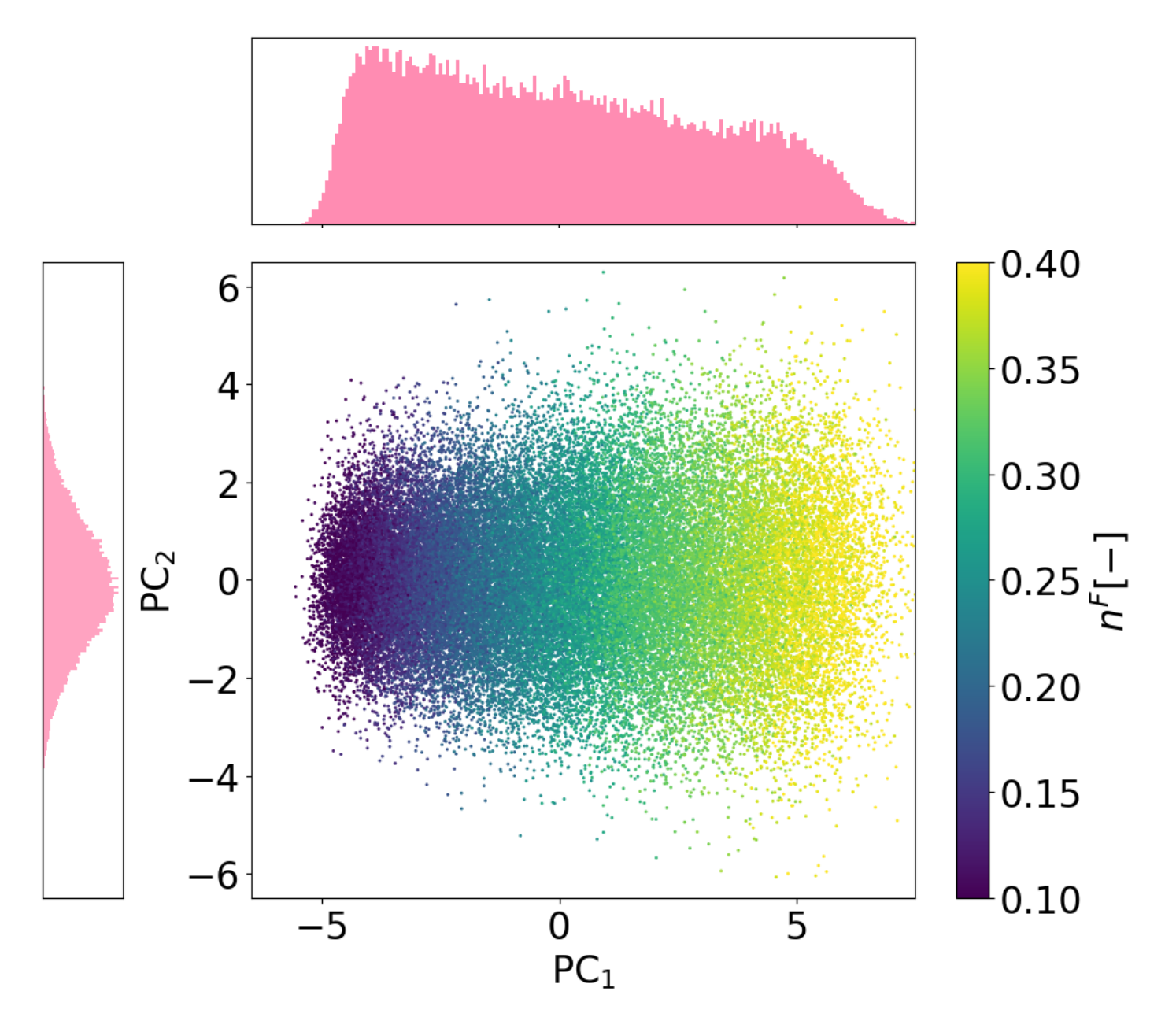}
\end{center}
\caption{Synthetic dataset: Visualization of the latent space of VAE mapped with regressor using the first two PCA components colored by the porosity. The original VAE (left), the $n^F$ mapping only (middle), and the combined $n^F$ and $K^S$ mapping (right). The x-axis and y-axis represent the first and second principal components, respectively.}
\label{Fig:acaLatentCompared}
\end{figure}

The Principal Component Analysis (PCA) is applied to the latent vectors of the entire dataset to visualize the information captured within the pVAE's latent space. To gain a deeper insight into how the latent space responds under the guidance of hydraulic properties, Figure~\ref{Fig:acaLatentCompared} shows how the latent space adapts as the model is trained to predict hydraulic properties. By projecting the latent representations onto the first two principal components, Figure~\ref{Fig:acaLatentCompared}, left, shows unorganized samples encoding from the original VAE, Figure~\ref{Fig:acaLatentCompared}, middle, shows how samples are organized with respect to porosity. Joint training on both porosity and permeability leads to a shift in the latent distribution, making it more structurally dispersed across the latent space along the first principal component (PC1), as shown in Figure~\ref{Fig:acaLatentCompared} (right). This shift is consistent with the observed linear correlation between porosity and permeability (Figure~\ref{Fig:JointPairwise}), suggesting that the latent space captures shared information between these properties.

\begin{figure}[!ht]
\begin{center}
\includegraphics[width=15cm]{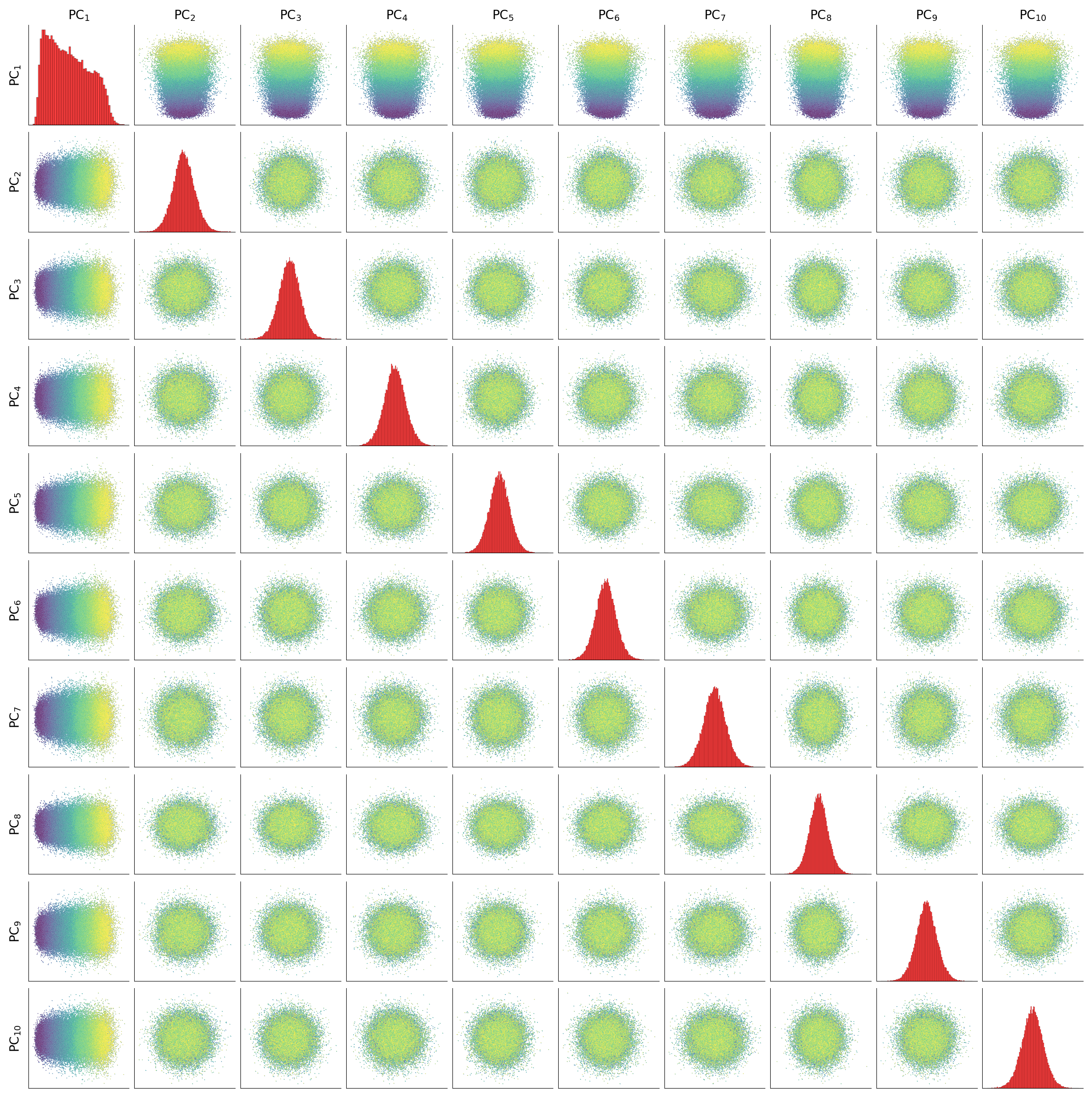}
\end{center}
\caption{Synthetic dataset:
Pairwise distribution of PCA components in the latent space. The diagonal plots (red histograms) show the distribution of individual PCA components, while the off-diagonal scatter plots illustrate the relationships between different PCA components colored by the porosity ($n^F$). The circular shape of most scatter plots suggests a minimal correlation between components, indicating that PCA effectively captures independent modes of variation in the latent space.
}
\label{Fig:acalatentPCAcomponents}
\end{figure}

\begin{figure}[!ht]
\begin{center}
\includegraphics[width=13cm]{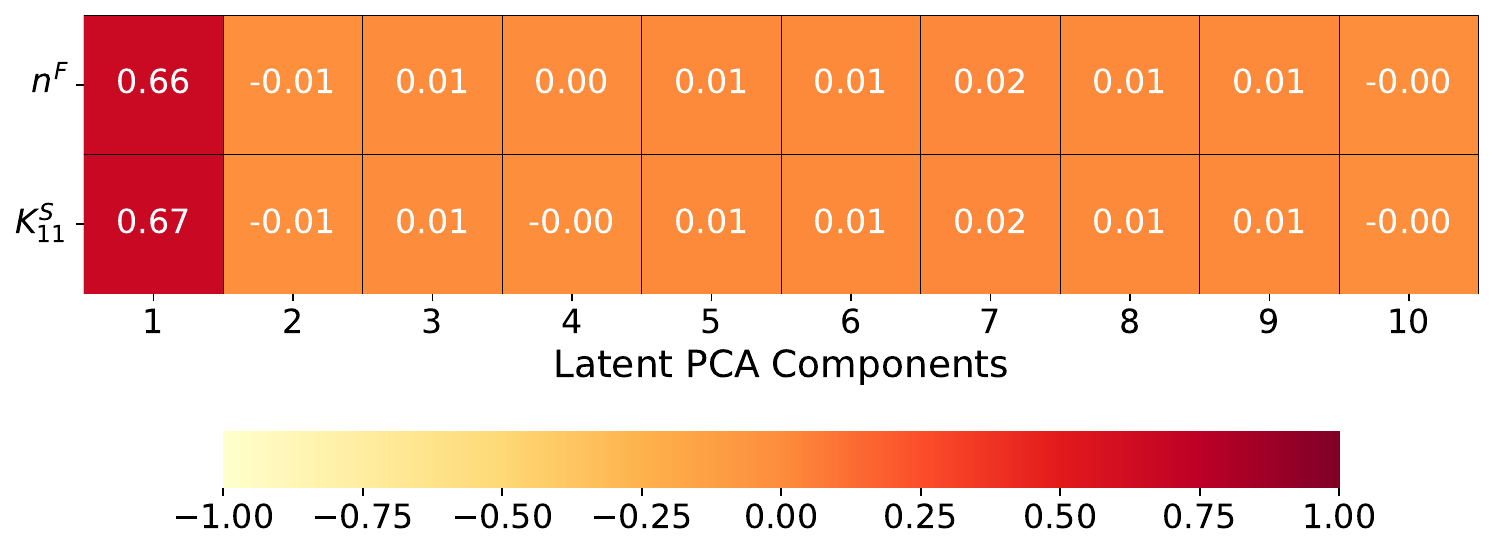}
\end{center}
\caption{Synthetic dataset: 
Heatmap of latent-property correlations with PCA components. The color intensity represents the strength of the correlation between the first 10 PCA components and two key properties; $n^F$ and $K^S_{11}$. 
}
\label{Fig:acaDiffMatrix}
\end{figure}

Figure~\ref{Fig:acalatentPCAcomponents} presents the first 10 PCA components in the latent space, providing insights into the variance captured by each component. The diagonal histograms reveal that the first principal component (PC1) exhibits the widest distribution, approximately Gaussian, with a higher standard deviation, indicating its dominant role in capturing latent space variability. In contrast, the higher-order components (PC2, PC3, ...) display narrower, more concentrated distributions, suggesting they contribute less variance.
%
%
The importance of PC1 in encoding key structural information is further emphasized in Figure~\ref{Fig:acaDiffMatrix}, which displays the correlation between PCA components and the two latent properties $n^F$ and $K^S_{11}$. This reflects the ability of the model to encode relevant structural variations. PC1 exhibits the strongest correlation with both properties ($n^F=0.66$, $K^{S}_{11}=0.67$), confirming its dominant role in capturing relevant variations. In contrast, higher-order PCA components show much weaker correlations and exhibit minimal influence on latent property representation. This aligns with their narrower distributions in Figure~\ref{Fig:acalatentPCAcomponents}, reinforcing the idea that PC1 encodes the primary structural variations, while the remaining components contribute to finer, less significant details.

\begin{figure}[!ht]
\begin{center}
\includegraphics[width=15.0cm]{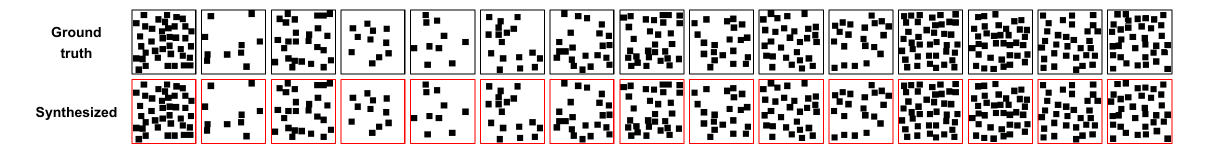}
\end{center}
\caption{Synthetic dataset: 
Reconstruction comparison using prior sampling from the latent space of pVAE model. The ground truth microstructures (top rows) are compared with their synthesized counterparts (bottom rows, highlighted in red) generated by the decoder. The visual correspondence between the two sets provides insights into the model’s reconstruction accuracy and its ability to learn a meaningful latent representation.}
\label{Fig:acapriorSampling}
\end{figure}

Figure~\ref{Fig:acapriorSampling} provides the reconstruction accuracy by sampling randomly from the latent space for the decoder. The synthesized images are visually compared to the ground truth images.

\begin{figure}[!ht]
\begin{center}
\includegraphics[width=7.5cm]{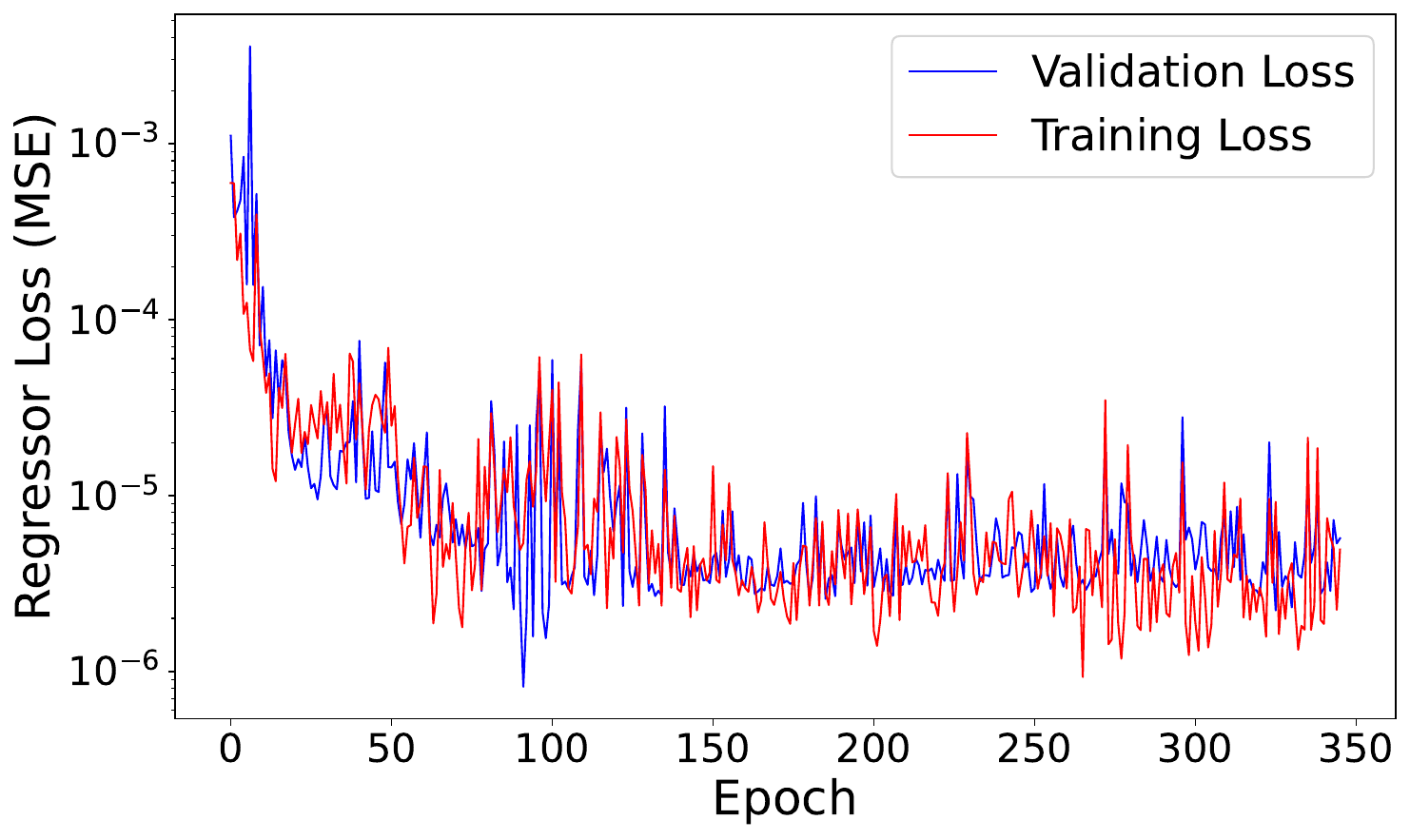}
\includegraphics[width=7.5cm]{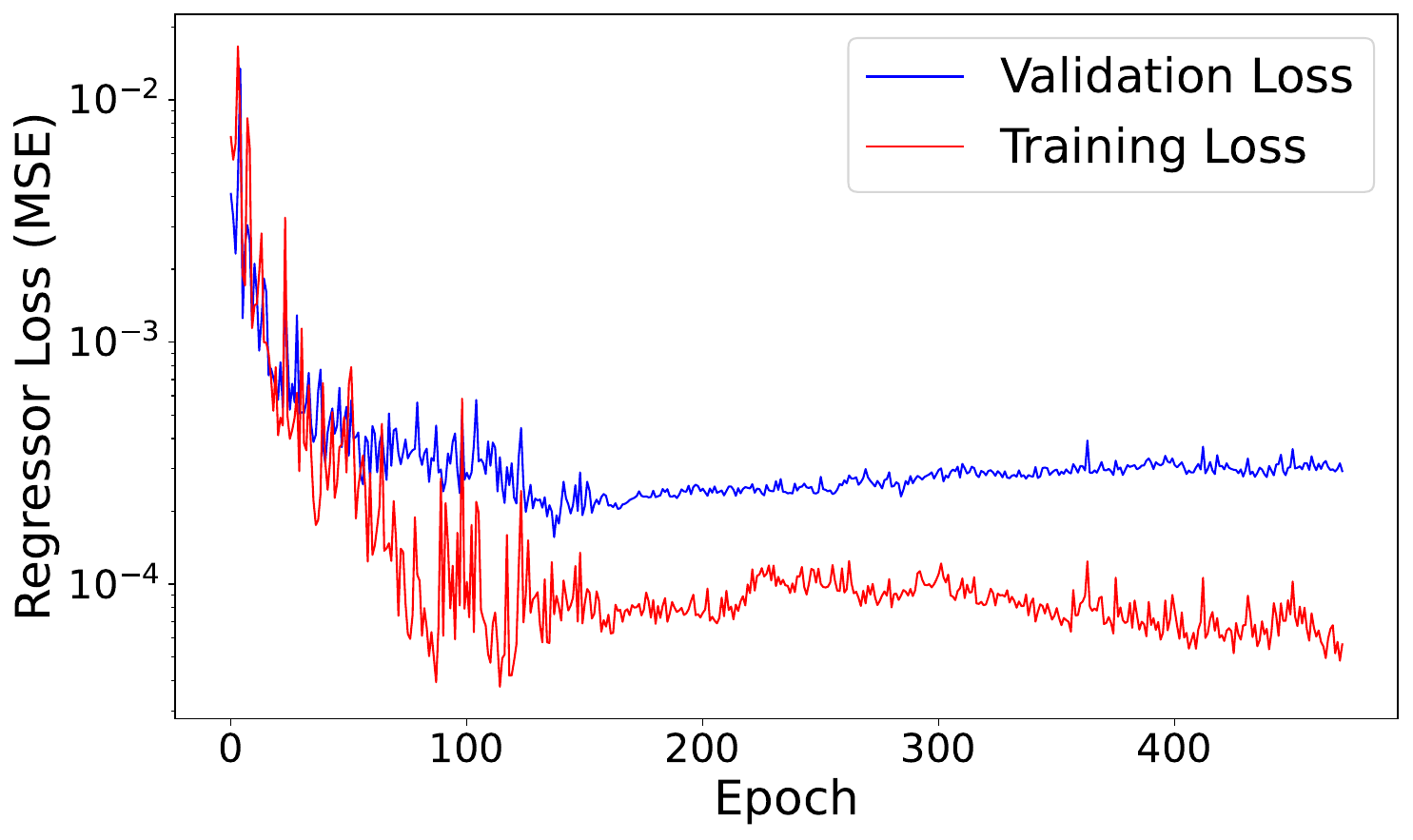}
\end{center}
\caption{Synthetic dataset: Training and validation loss (MSE) curves for the pVAE regressor. The left plot shows the model trained to predict porosity $n^F$ only, while the right plot corresponds to joint training on both $n^F$ and $K^S_{11}$.}
\label{Fig:acaPredLoss}
\end{figure}

\begin{figure}[!ht]
\begin{center}
\includegraphics[width=7.5cm]{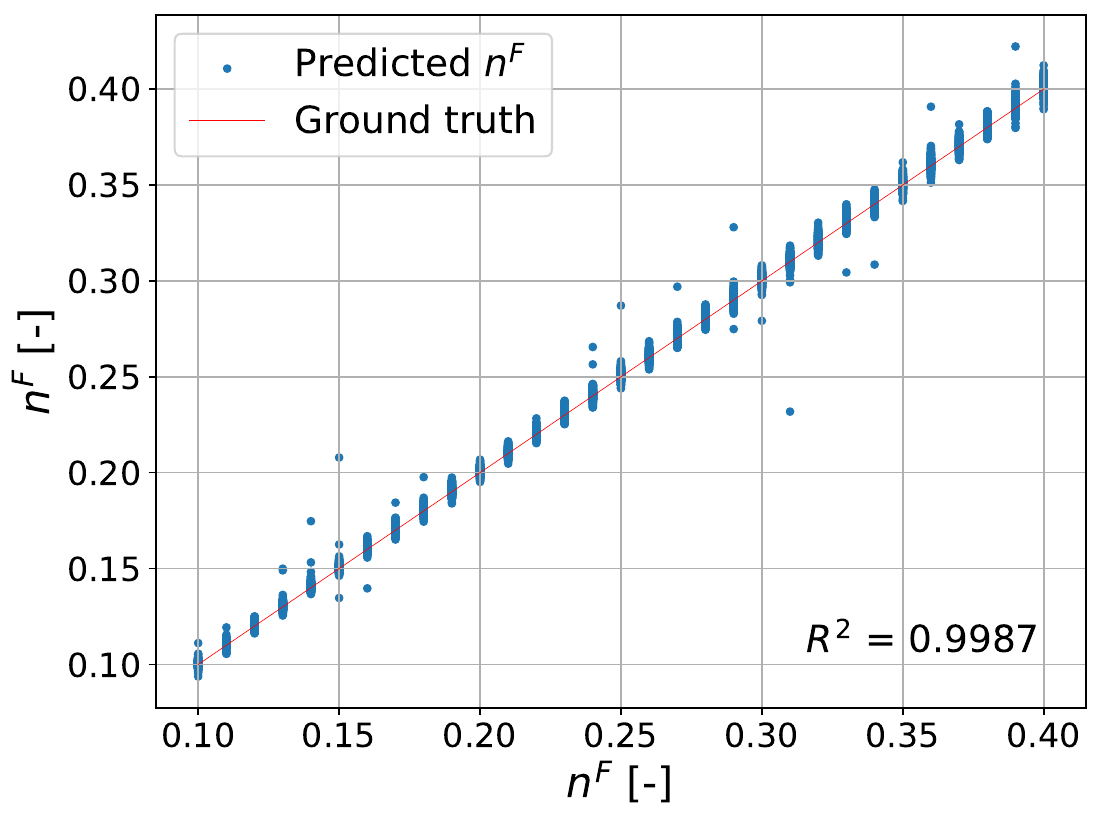}
\includegraphics[width=7.5cm]{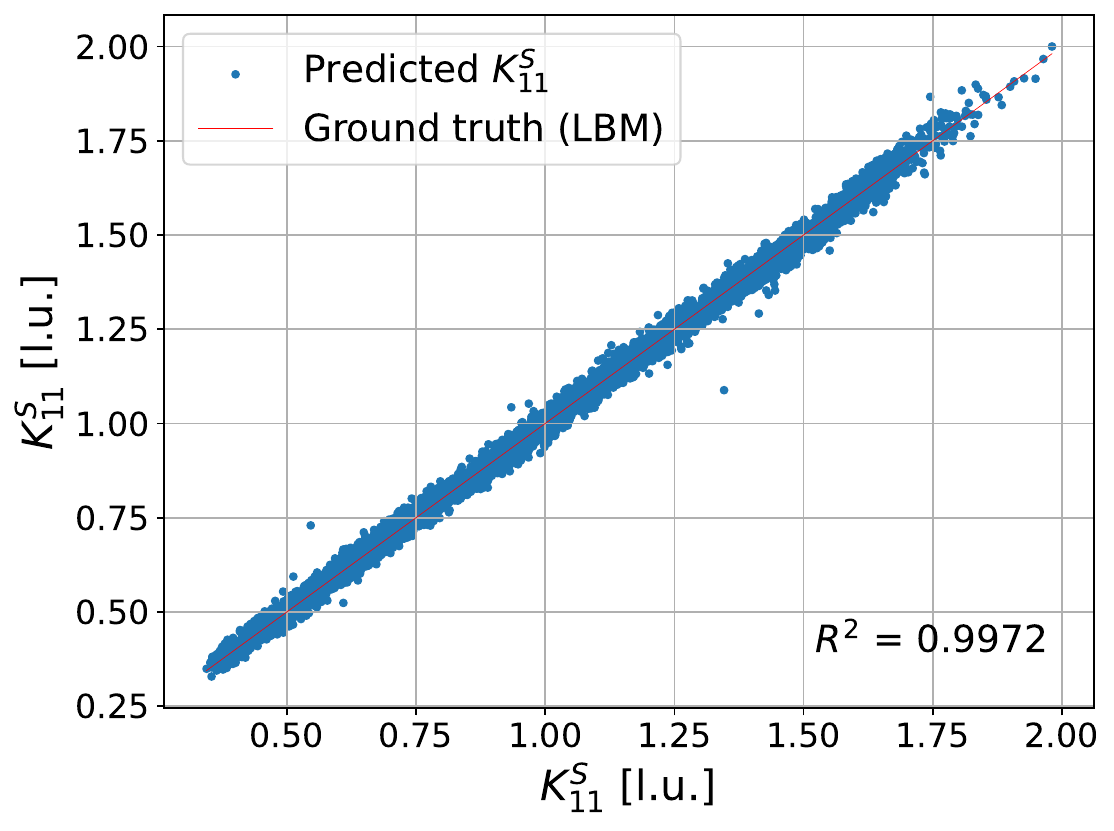}
\end{center}
\caption{Synthetic dataset: Evaluation of the prediction accuracy of the property predictor model. Predicted vs. true components of the porosity $n^F$ and $K^{S}_{11}$ in the test dataset. The predicted properties are computed using the regressor ${\bf f}_{\omega}$. The solid red lines represent the ground-truth values with zero intercept and unit slope; the corresponding $R^2$-scores are indicated.}
\label{Fig:acaR2accuracy}
\end{figure}

Figure~\ref{Fig:acaPredLoss} illustrates the regressor loss (MSE) of the pVAE model. The left plot corresponds to mapping solely to $n^F$, while the right plot includes both $n^F$ and $K^{S}_{11}$ as target properties. The pVAE model trained with only $n^F$ achieves a lower loss compared to the model trained on both $n^F$ and $K^{S}_{11}$. The presence of different microstructures with the same porosity but different permeability, leading to non-unique solutions, explains this observation.
The evaluation of the regressor is conducted on the test dataset, which constitutes $15\%$ of the entire dataset. Figure~\ref{Fig:acaR2accuracy} shows that the $R^2$ values are approximately 0.998 for both $n^F$ and $K^{S}_{11}$, demonstrating good prediction accuracy.
The strong performance of the structure-property mapping establishes a continuous and meaningful latent space that effectively represents the microstructure geometry in a low-dimensional space. This enables interpolation and gradient-based optimization for design purposes.
\subsection{Interpolation in the latent space}
\label{subsec:aca-interpolation}
In this work, spherical linear interpolation ({\it slerp}) is employed instead of linear interpolation ({\it lerp}) in the latent space to ensure smooth and realistic transitions between points. Unlike {\it lerp}, which assumes a straight-line interpolation in Euclidean space, {\it slerp} follows the shortest geodesic path on a hypersphere, preserving the intrinsic structure of the latent space. This is particularly important in non-Euclidean latent spaces, where linear interpolation can lead to inconsistent or unrealistic transitions due to the curvature of the data distribution.
By maintaining a constant interpolation speed along the hyperspherical path,  {\it slerp} better preserves the continuity and coherence of the latent representations, resulting in smoother transitions in both geometry and property variations. This method effectively reduces artifacts and ensures a more meaningful interpolation within the learned latent space. For further details, refer to \cite{white2016sampling, bombarelli2018chemicalInv, lizheng2023trussinv}.

%

\begin{figure}[!ht]
\begin{center}
\includegraphics[width=6.5cm]{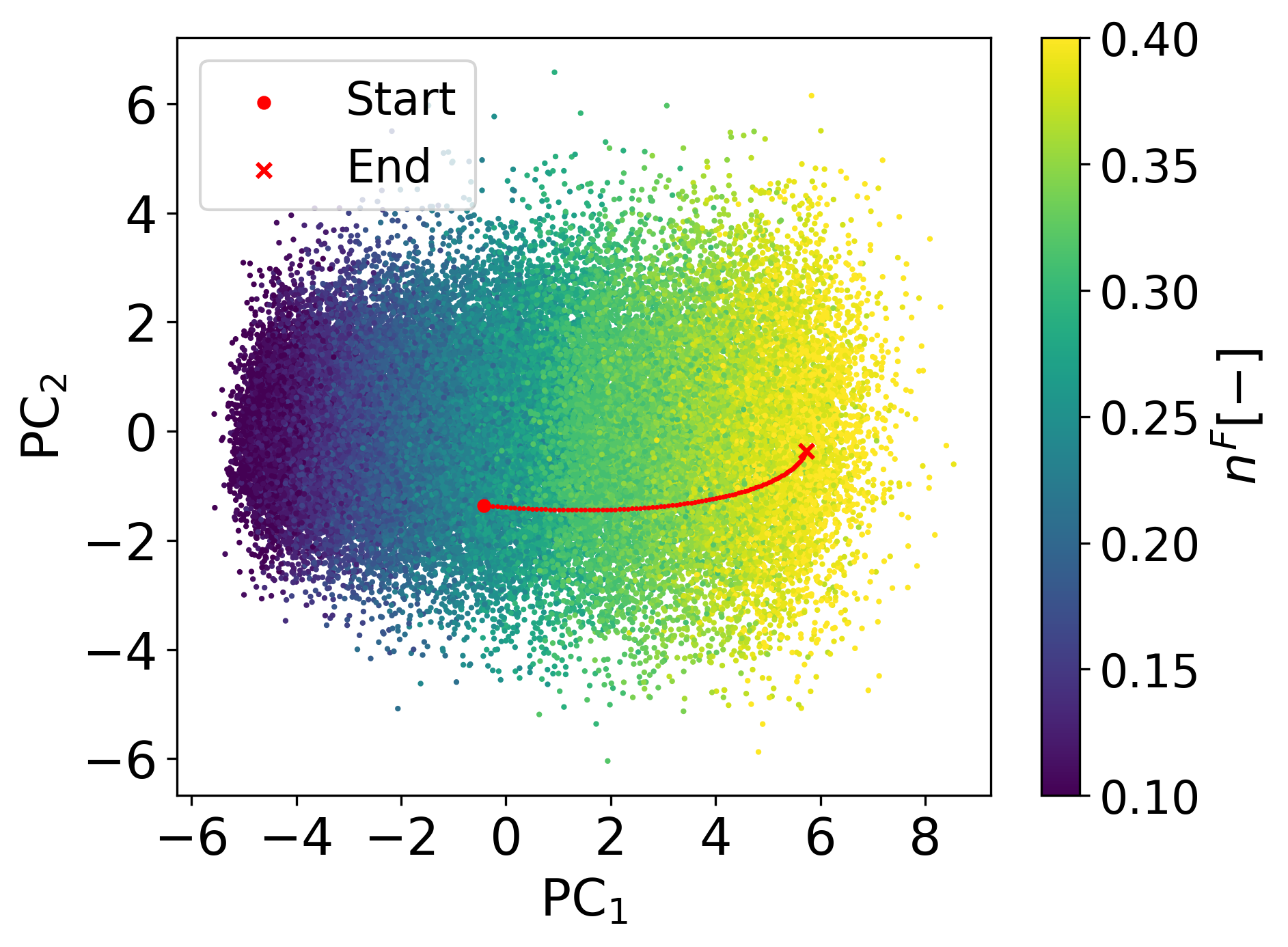}~~~~
\includegraphics[width=8.0cm]{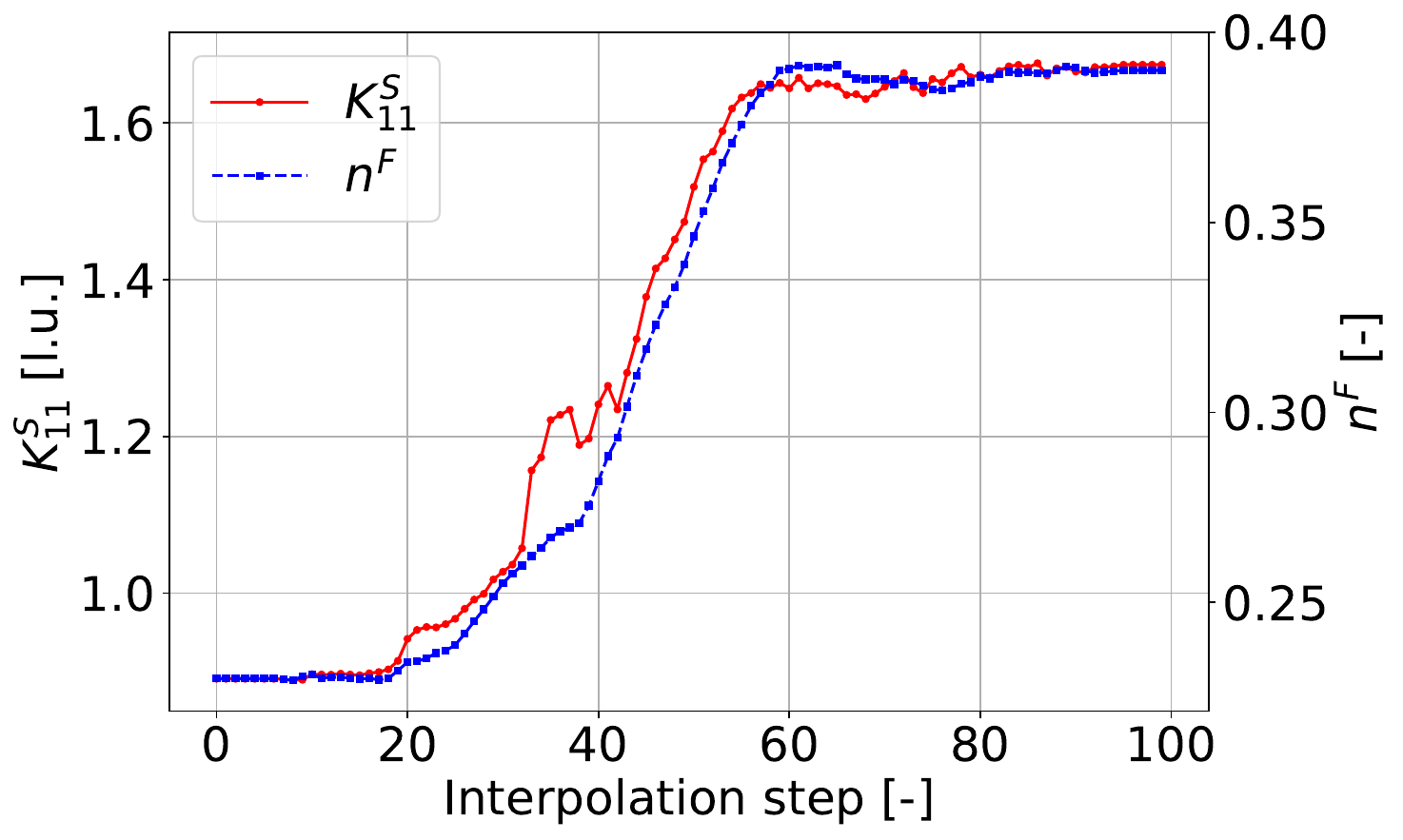}
\includegraphics[width=13.0cm]{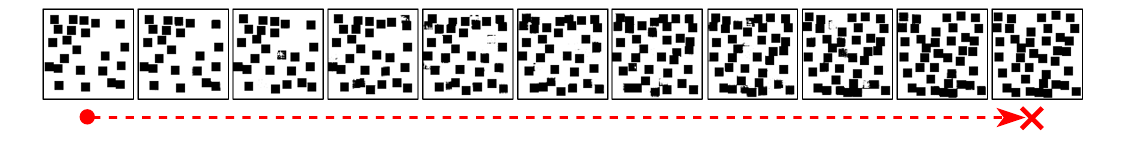}
\end{center}
\caption{
Synthetic dataset: {\it Slerp} in Latent Space. The top-left plot shows the latent space representation using the first two principal components (PC1 \& PC2), with data points color-coded by $n^F$. A smooth trajectory (red curve) represents the interpolation path between two latent points (start and end). 
The top-right figure shows the evolution of $n^F$ and $K_{11}^S$ along the interpolation path in latent space.  
The bottom row shows the corresponding interpolated microstructures, indicating a gradual and realistic transformation between the two endpoints.
}
\label{Fig:acaLatentInterpolation}
\end{figure}

Figure~\ref{Fig:acaLatentInterpolation} presents an example of {\it slerp} between two distinct porous microstructures over 100 interpolation steps, transitioning from low to high porosity $n^F$. The bottom row of Figure~\ref{Fig:acaLatentInterpolation} illustrates eleven representative microstructure images sampled along the interpolation trajectory, showing a gradual densification of square pores as porosity increases. 
The top-right plot of Figure~\ref{Fig:acaLatentInterpolation}, generated using a CNN-based surrogate model (see, section~\ref{subsec:CNN}),  shows $n^F$ and $K^{S}_{11}$ of the microstructures along the interpolation path, highlighting the model’s ability to capture meaningful structure-property relationships in the latent space.
The smooth and continuous latent space ensures that, despite the significant differences in volume fraction and hydraulic properties between the start and end  micro-structures, the transition of structural geometries is seamless. This capability opens new opportunities for designing continuous families of porous microstructures with property gradients, effectively bypassing complex optimization algorithms that operate in high-dimensional, discrete design spaces.

These results emphasize the significance of latent space mapping in capturing the structure-property relationship, demonstrating that microstructures with similar properties naturally cluster within the latent space. This organization enables efficient sampling within the latent space to enrich porous microstructure datasets for homogenization or multiscale modeling purposes. Furthermore, the framework supports gradient-based optimization to identify microstructures corresponding to desired properties, providing an efficient solution for inverse design problems.
%


%
The dataset in this study is treated as a Statistical Volume Element (SVE), introducing microscale uncertainties that result in effective properties with inherent variations, making standard deviations an essential consideration for macroscopic homogenization.
With a continuous and well-structured latent space, targeted sampling can be performed to generate microstructures with desired effective properties. By selecting latent points near the target region, this approach ensures that the synthesized microstructures align with expected macroscopic behaviors, facilitating traditional homogenization. In the following section, we present the inverse design result that deterministically identifies microstructures satisfying predefined property constraints. This result highlights key challenges in the inverse design of porous materials represented by pixels as well as voxels, particularly in simultaneously guiding the latent space of a VAE to conform to specific volume fraction and hydraulic property distributions through a regression model.

\subsection{Gradient-based optimization in the latent space}
\label{subsec:Gen-3Dstructure-based-P}
\begin{figure}[!ht]
\begin{center}
\includegraphics[width=9.0cm]{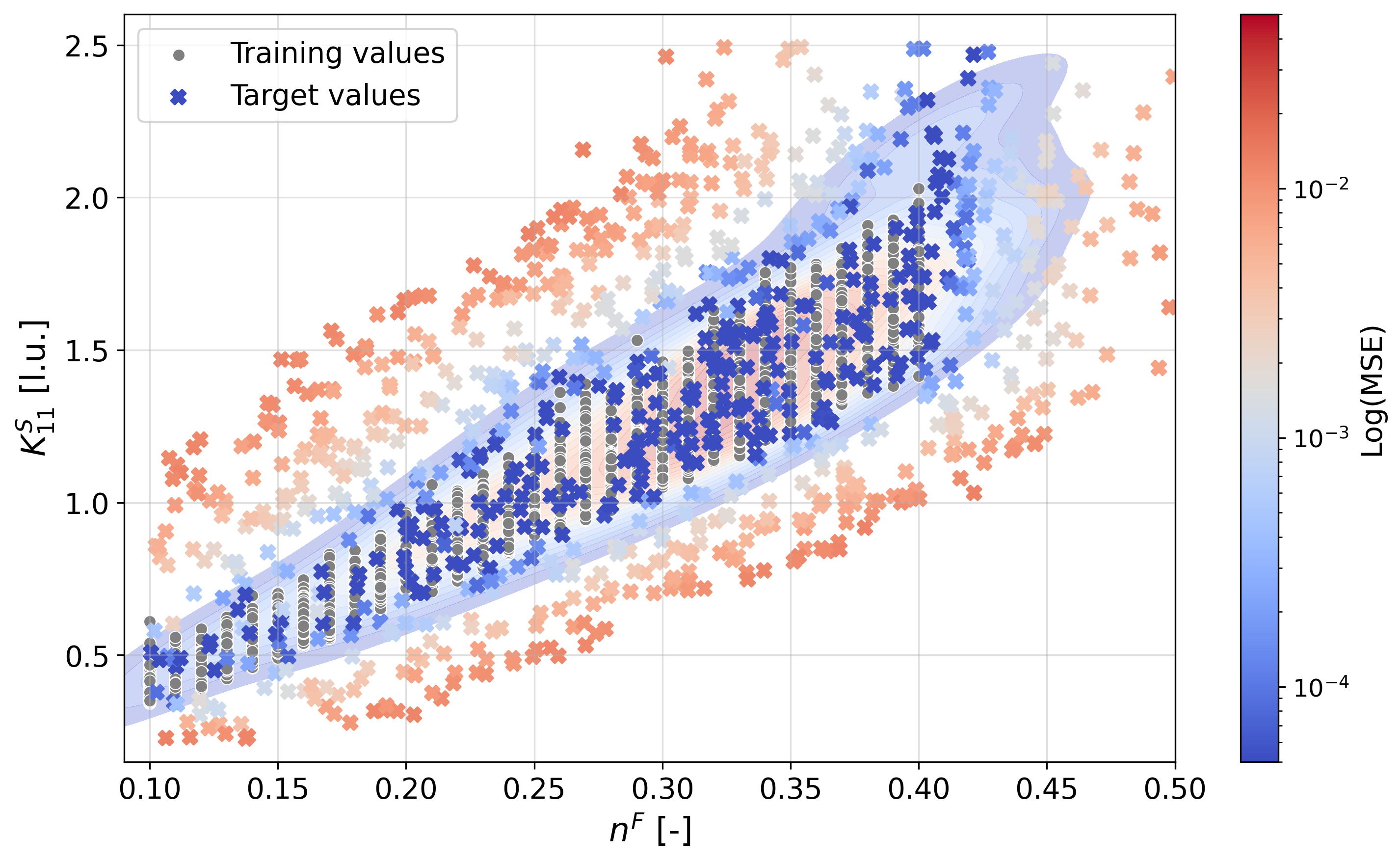}
\includegraphics[width=10.0cm]{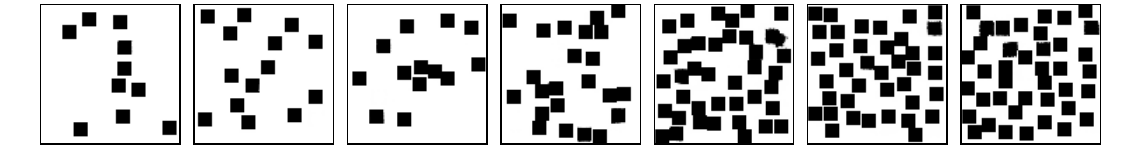}
\end{center}
\caption{Synthetic dataset: 
Random sampling of target properties for employing gradient-based optimization in the latent space. The scatter plot displays the target properties $n^F$ and $K^{S}_{11}$, with colors representing the logarithm of the MSE (sum of the MSE for porosity and intrinsic permeability) of the generated microstructures. A slightly smoothed colored area in the background is derived from the probability density function (PDF) of a continuous variable using KDE with the Seaborn library. The bottom row presents representative examples of the designed microstructures.
}
\label{Fig:acaTargetValues}
\end{figure}

To showcase the capabilities of the pVAE model in leveraging a continuous and smooth latent space, an inverse design framework, as introduced in Subsection~\ref{subsec:InvDesignFramework}, is applied. 
A uniform random distribution of 1000 target property samples is generated, with $n^F$ ranging from 0.1 to 0.5 and $K^{S}_{11}$ from 0.2 to 2.5. Each target property set is then fed into the inverse framework, where gradient-based optimization is performed to generate corresponding microstructures. The optimization process, integrated with stochastic gradient descent (SGD), is terminated when the average $L_2$ loss falls below $10^{-5}$ or when no further improvement is observed after 400 epochs.
The designed microstructures are subsequently evaluated using the CNN surrogate model (section~\ref{subsec:CNN}), which predicts their hydraulic properties. These predicted values are compared with the target properties to assess the performance of the deep learning framework. 
Figure~\ref{Fig:acaTargetValues} illustrates the distribution of target property samples alongside the corresponding designed microstructures, highlighting the effectiveness of the Deep Neural Network (DNN) framework in generating porous structures that match the desired properties ($n^F$ and $K^{S}_{11}$). The color map represents the logarithm of the MSE, \textcolor{black}{computed as the logarithm of the sum of the MSE for porosity and intrinsic permeability}, providing insight into the accuracy of the generated structures across different target property values. Due to the wide variation in $L_2$  errors, the values are presented on a logarithmic scale to enhance interpretability.
Additionally, a slightly colored area in the background highlights regions where the MSE value is small, derived using KDE.

The inverse design framework also enables the generation of novel microstructures beyond those present in the training dataset. Low-error regions correspond to areas within the trained dataset and follow the expected linear relationship between $n^F$ and $K^{S}_{11}$. However, samples with target properties that deviate significantly from this trend exhibit higher error values, as indicated by the color scale. Representative microstructure images, generated using the inverse framework for specific target properties, are displayed below the scatter plot in Figure~\ref{Fig:acaTargetValues}. These examples highlight the generative capability of the pVAE model in producing diverse porous structures while striving to match the desired target properties.

\begin{figure}[!ht]
\begin{center}
\includegraphics[width=7.4cm]{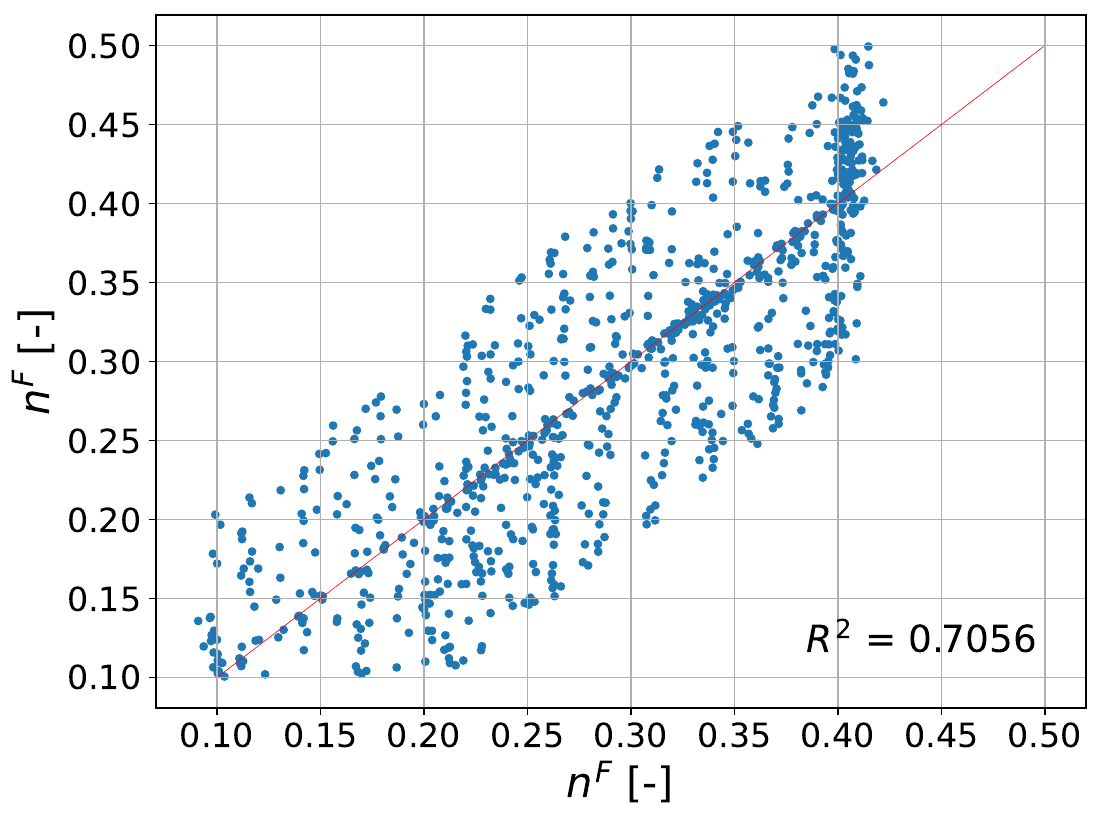}
\includegraphics[width=7.5cm]{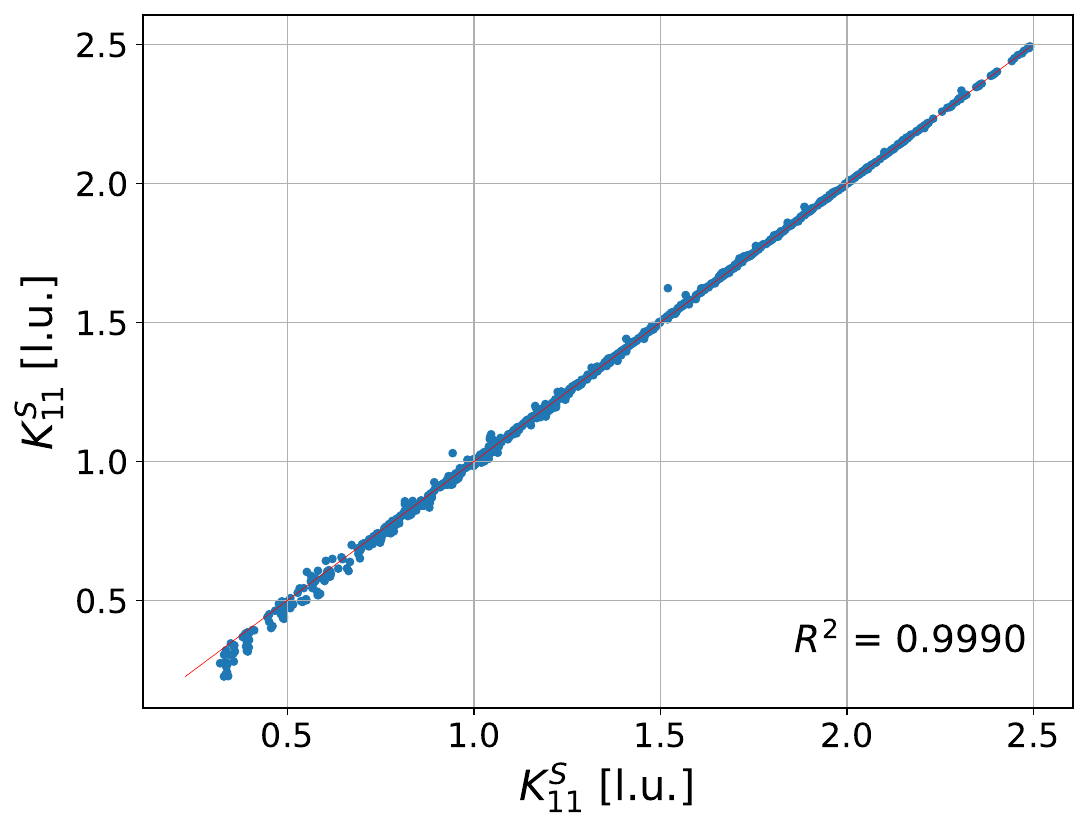}
\end{center}
\caption{Synthetic dataset: 
Evaluation of the inverse design framework by comparing designed vs. target values of porosity ($n^F$) and intrinsic permeability ($K^{S}_{11}$). The properties are computed using the surrogate CNN model based on the generated microstructure images via pVAE. The red lines represent the ground truth. The corresponding $R^2$-scores indicate the accuracy of the inverse design process, demonstrating a strong correlation for $K^{S}_{11}$ and a moderate correlation for $n^F$.
}
\label{Fig:acaInvaccuracy}
\end{figure}

For a more detailed evaluation, the accuracy is assessed separately for each property. The properties of the designed microstructures are evaluated using the coefficient of determination ($R^2$ values). Figure~\ref{Fig:acaInvaccuracy} illustrates the accuracy of the inverse design process. The red line represents the target properties fed into the inverse framework. The effective permeability ($K^{S}_{11}$) achieves a high $R^2$-score of 0.9990, indicating strong predictive accuracy. However, the porosity ($n^F$) exhibits a noticeable deviation from the target values, as reflected by a lower $R^2$-score of 0.7056. This discrepancy in $n^F$ contributes to the $L_2$ error distribution observed in Figure~\ref{Fig:acaTargetValues}.  
This phenomenon arises because the pVAE model prioritizes the mapping of effective permeability over porosity when jointly encoding both properties. As a result, the latent space becomes constrained and reorganized, with its structure predominantly influenced by $K^{S}_{11}$. Consequently, during optimization, the latent vector selected aligns more closely with the target permeability rather than the target porosity. This behavior is influenced by the inherent linear relationship between $n^F$ and $K^{S}_{11}$, dictated by Darcy’s law in the straight square pipes forming these microstructures. Target properties that significantly deviate from this linear correlation are physically unrealistic for this type of porous structure with assumed laminar flow. This observation underscores the necessity of exploring the model’s applicability to more heterogeneous and complex-channel porous metamaterials in future research. 

The pVAE is currently biased toward permeability prediction, as reflected by the higher  $R^2$. This bias likely stems from the model's ability to more effectively encode and reconstruct connectivity-driven features in the latent space, which are more directly tied to permeability than porosity. Additionally, structural variations with similar porosity but different flow paths introduce uncertainty in porosity prediction, making it a more ill-posed problem.
%

In the context of the synthetic dataset, the performance of the pVAE framework demonstrates a smooth, continuous, and meaningful latent space. Microstructures with similar hydraulic properties are effectively clustered within specific regions of the latent space, providing an efficient tool for multiscale simulations based on a probabilistic representation of microstructures. The gradient-based optimization within this latent space highlights the framework’s potential for the inverse design of porous microstructures.  To ensure a more robust design space, the dataset used for training the pVAE framework must be large and diverse. Expanding the dataset to include a broader range of material properties would improve the mapping accuracy and impose additional constraints to enhance generalizability.

In the next example, the pVAE framework is evaluated using a real-world dataset obtained from CT scans of viscoelastic foam, as detailed in \cite{PhuEtAl2023_PAMM}. Due to the high cost of data acquisition, the dataset size is limited. Thus, this study focuses on analyzing the performance of the pVAE-based DNN framework in capturing the probabilistic distribution of limited real microstructures in latent space and assessing the effectiveness of this latent representation.


\section{Real open-foam material}
\label{sec:pVAErealFoam}
The second application examines the pVAE and its latent space in the context of real-world open-foam $\mu$-CT data. This section explores the framework’s limitations arising from dataset constraints, addresses key challenges, and discusses potential advancements for improving inverse design in porous metamaterials.

\subsection{Experimental $\mu$-CT data}
\label{subsec:RW_data}
\begin{figure}[!ht]
\begin{center}
\includegraphics[width=12.0cm]{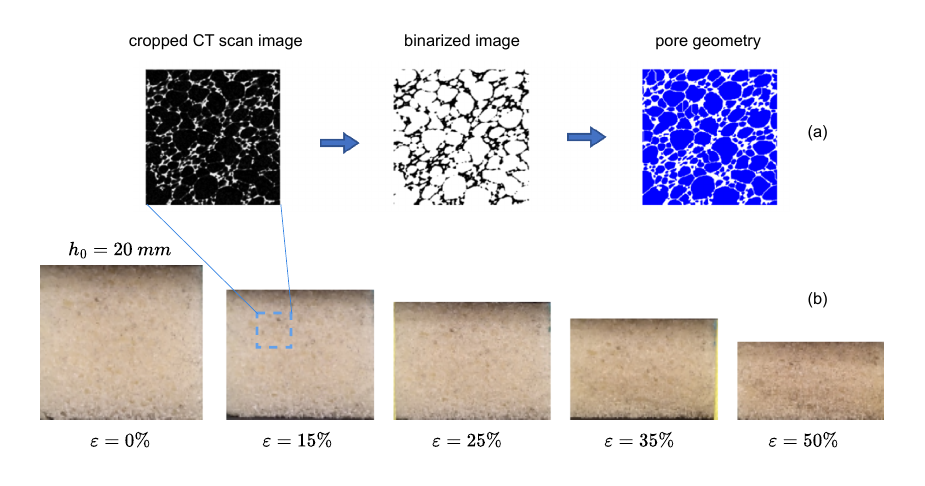}
\end{center}
\caption{(a) Image processing workflow for pore structure extraction from $\mu$-CT scan data, transitioning from cropped images to binarization and pore geometry visualization using the MATLAB function label2rgb(). (b) Compression stages of a  foam specimen under static loading, showing progressive deformation at different strain levels \cite{PhuEtAl2023_PAMM}.}
\label{Fig:SampleCompressCT}
\end{figure}

To evaluate the capabilities and limitations of the proposed pVAE, we analyze real open-foam $\mu$-CT data scanned from an open-cell viscoelastic foam. Figure~\ref{Fig:SampleCompressCT} illustrates the image processing steps used to extract and analyze the pore structures during compression. Specifically, the bottom face of the sample is fixed, all lateral faces are assumed to be constrained against lateral deformation, and a displacement-controlled loading is applied to the top face. The workflow includes segmentation of $\mu$-CT images, binarization, and visualization of the pore geometry. Additionally, the progressive deformation of the foam specimen under static loading is depicted, providing insights into its structural behavior at varying strain levels. Further details about data acquisition are available in \cite{PhuEtAl2023_PAMM}.

\begin{figure}[!ht]
\begin{center}
\includegraphics[width=7.0cm]{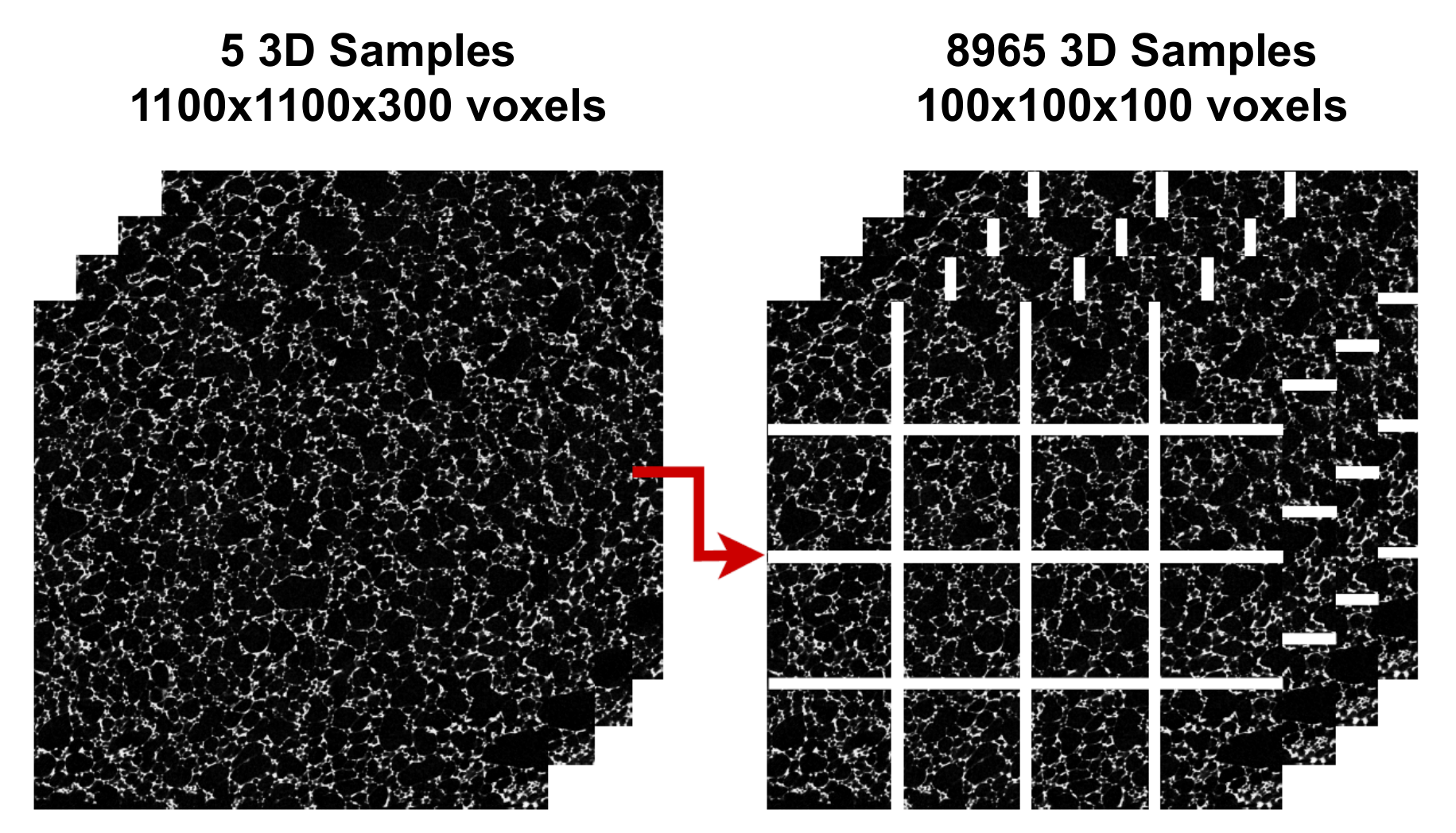}
\end{center}
\caption{Visualization of the data sampling process. The original 3D dataset (left) comprises five large samples of size $1100 \times 1100 \times 300$ voxels. The sampling (right) results in 8965 smaller 3D sub-samples, each measuring $100 \times 100 \times 100$ voxels.}
\label{Fig:Sampling_8965_porosity}
\end{figure}

For database preparation for training, the raw CT images of the 3D samples, originally sized at 1100$\times$1100$\times$300 voxels, undergo a structured sampling process.
This process yields a larger dataset of 8965 3D samples with smaller sizes of 100$\times$100$\times$100 voxels, which ensures a manageable data volume while retaining critical microstructural features. This approach is consistent with the methodology outlined by \citet{HongLiu2020_CNN_Permeability}. An illustration of the data sampling process used to obtain the 8965 3D samples is shown in Figure~\ref{Fig:Sampling_8965_porosity}.
%
%

The binarization process is conducted using thresholding, converting grayscale CT images into binary representations that differentiate between the pore space and the solid matrix. This enables the estimation of the material's porosity $n^F$ at each strain level.
%
The generative deep learning framework can be trained directly on the 3D images without binarization. However, these binary images are critical for the LBM simulations performed in Palabos \cite{Palabos2020}.
The LBM, which is used to simulate microscale flow through the samples of $100 \times 100 \times 100$ voxels, yields the intrinsic permeability components as shown in Appendices~\ref{appx:OneFluid_LBMTheor} and ~\ref{appx:InrinsicPermComputation}. 
These data are then used to train the regressor of the pVAE approach.

\begin{figure}[!ht]
\begin{center}
\includegraphics[width=7.2cm]{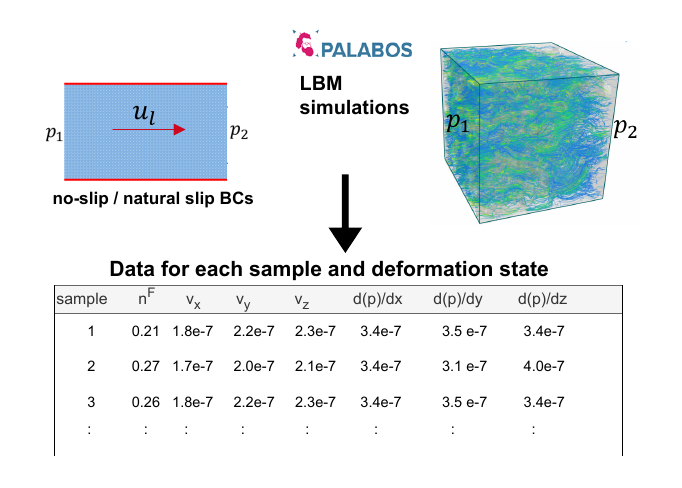}
\quad\quad
\includegraphics[width=7.2cm]{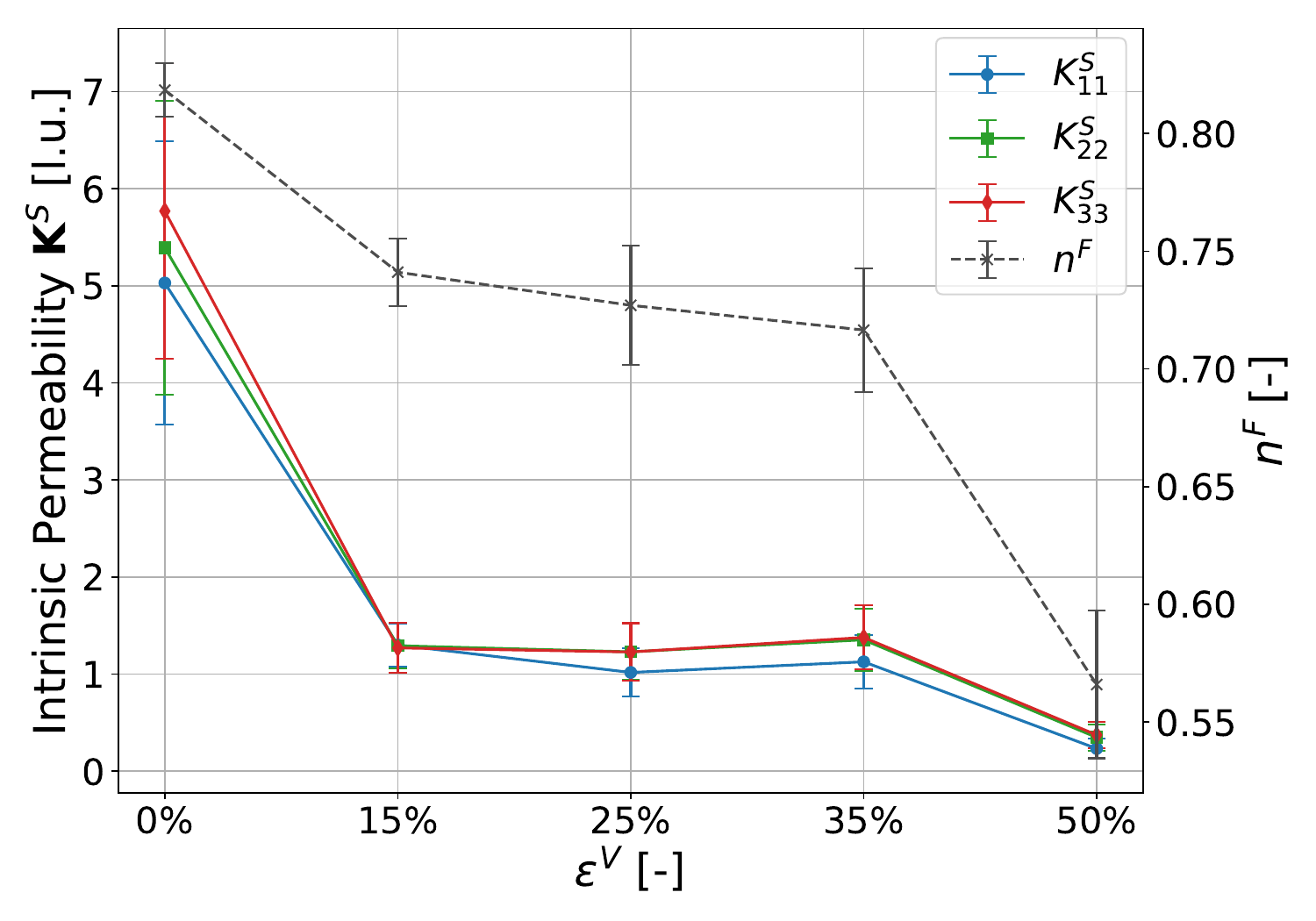}
\end{center}
\caption{Schematic of the generation of data using LBM with pressure drop and no-slip/natural-slip BCs ({left}).
Mean and standard deviation of the porosity $n^{F}$ and the intrinsic permeability components $K^S_{ii}$, $i=1, 2, 3$ for each strain level $\varepsilon^V$ showing the deformation-dependency and anisotropy (right).}
\label{Fig:LBM_illustraion_KS_Diag}
\end{figure}

To give an overview, Figure~\ref{Fig:LBM_illustraion_KS_Diag}, left, illustrates the data generation process using the LBM, while Figure~\ref{Fig:LBM_illustraion_KS_Diag}, right, shows the mean and standard deviation of the diagonal intrinsic permeability components $(K^S_l)_{11},\,(K_l^S)_{22},\,(K_l^S)_{33}$ for each strain level $\varepsilon^V$. This shows the deformation dependency and a slight anisotropy in the flow (different permeability components in different directions).
The macroscopic intrinsic permeability in [$\mathrm{m}^2$] units can be derived using Eq.~\ref{eqnConversion}, as detailed in Appendix~\ref{appx:OneFluid_LBMTheor}.
In the following, we will refer to the permeability tensor as $(\TK^S)$, regardless of whether it is in \(\big[\mathrm{l.u.}\big]\) or [$\mathrm{m}^2$], as this is only a unit change (scaling factor) and does not affect the accuracy of the pVAE algorithm.
\subsection{pVAE model evaluation}
\label{subsec:RealRepVAE}
This section examines the pVAE trained on real-world $\mu$-CT data, characterized by its heterogeneity, non-periodicity, and complex channel structures. As mentioned, the dataset is limited in size. The analysis follows the same methodology used for synthetic academic data, with the key difference that the inverse framework is not applied for microstructure design. 

\begin{figure}[!ht]
\begin{center}
\includegraphics[width=8cm]{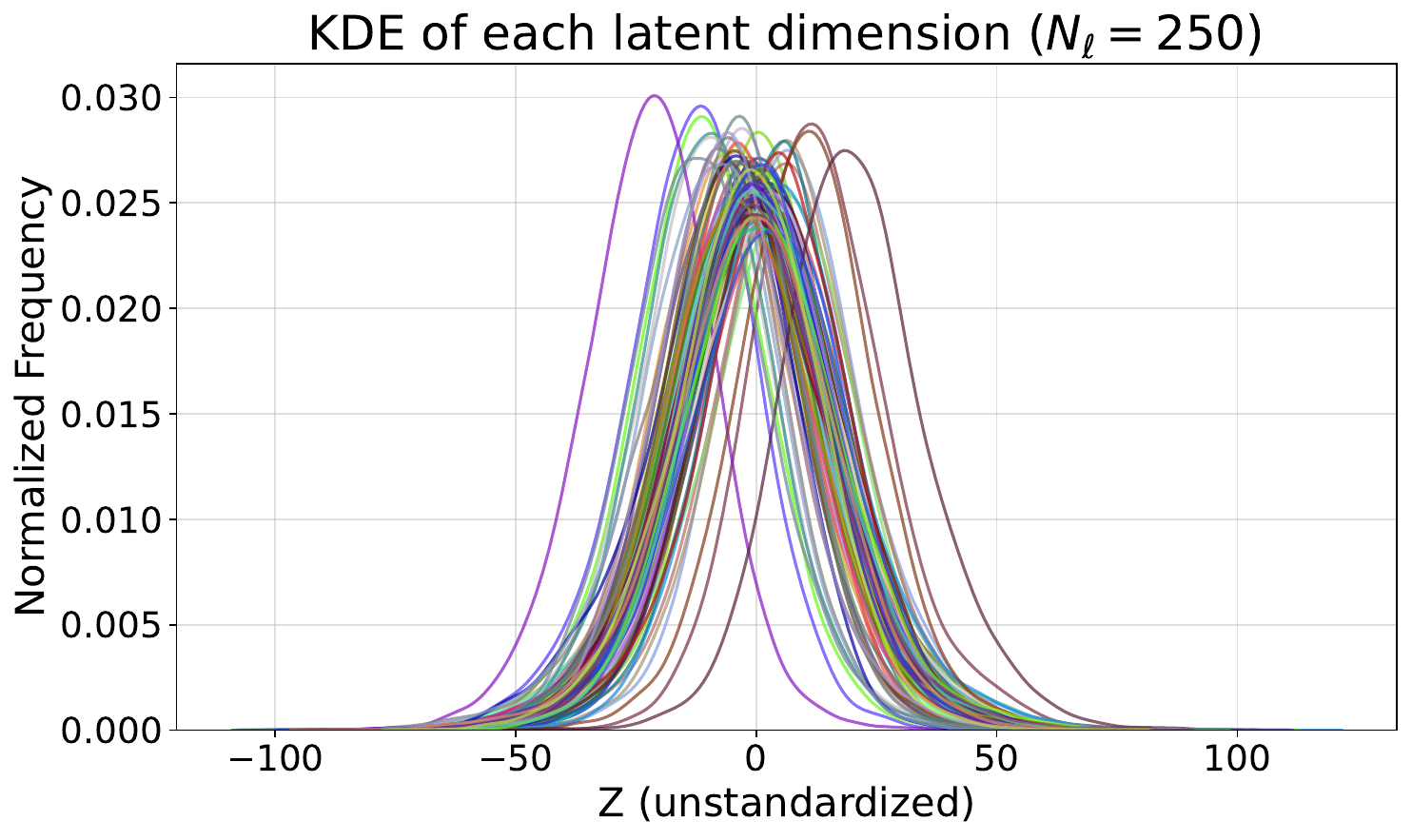}
\end{center}
\caption{Real-world $\mu-$CT data: 
Kernel Density Estimation (KDE) plots for each of the 250 latent dimensions in the autoencoder, illustrating the distribution of encoded microstructures. Each curve represents the probability density function of a single latent dimension across the dataset, showing variations in spread and shape.
}
\label{Fig:realKDE}
\end{figure}

Figure~\ref{Fig:realKDE} presents the KDE plots for each of the 128 latent dimensions in the pVAE, based on the encoding of 8.965 real microstructures. The distributions reveal that while most latent dimensions exhibit an approximately Gaussian shape, their means and standard deviations vary significantly. This variability suggests uncertainty in the mapping process, highlighting the complexity of the latent space representation. This reflects a structural heterogeneity and non-uniform feature importance in the encoded microstructural characteristics.

\begin{figure}[!ht]
\begin{center}
\includegraphics[width=8.0cm]{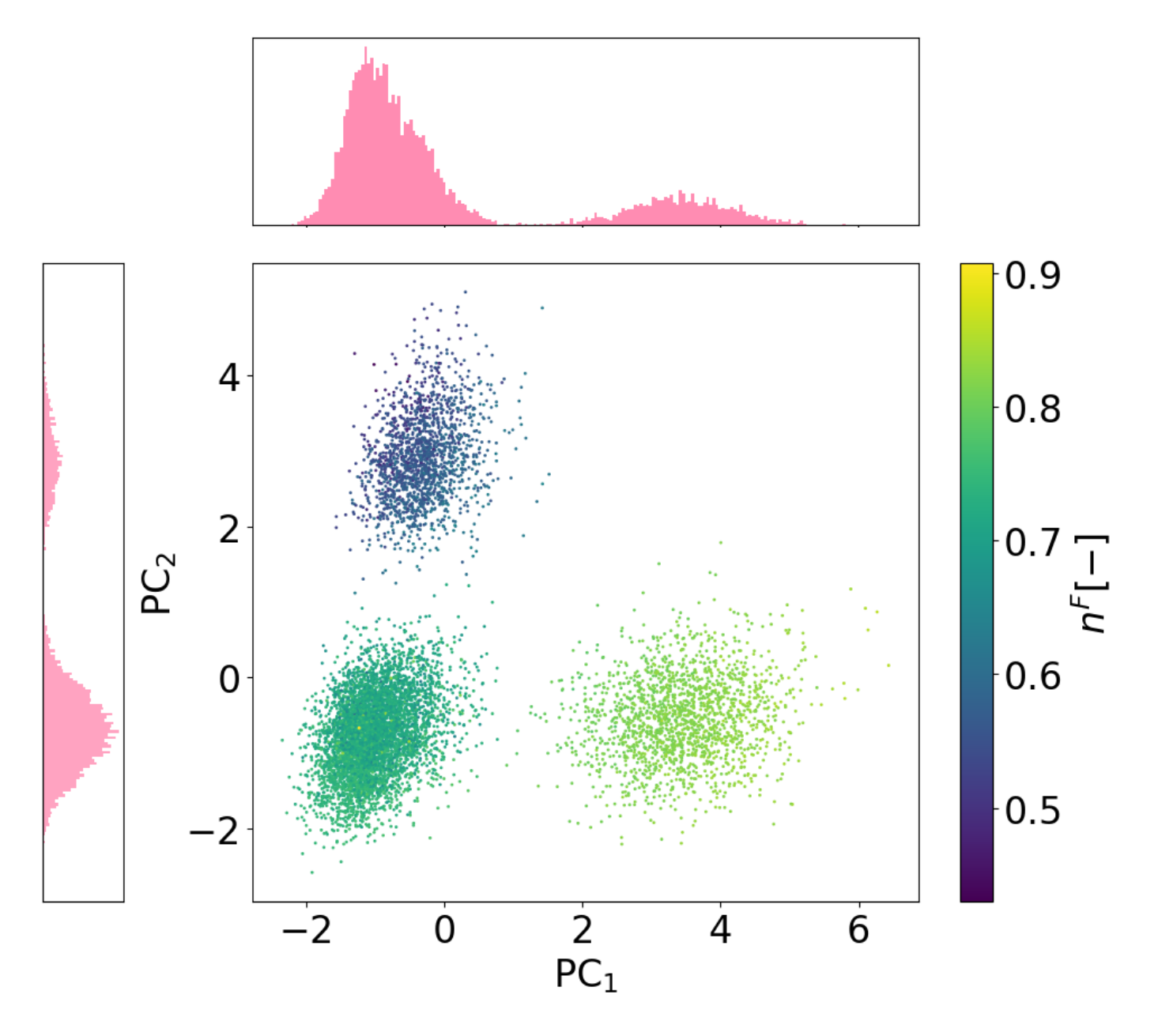}
\end{center}
\caption{Real-world $\mu-$CT data: 
Visualization of the latent space using PCA with points color-coded by  $n^F$. The scatter plot represents the first two principal components, corresponding to the $x$- and $y$-axes, respectively. The observed clustering suggests that the latent space effectively captures variations in porosity.
}
\label{Fig:realLatent}
\end{figure}

\begin{figure}[!ht]
\begin{center}
\includegraphics[width=15cm]
{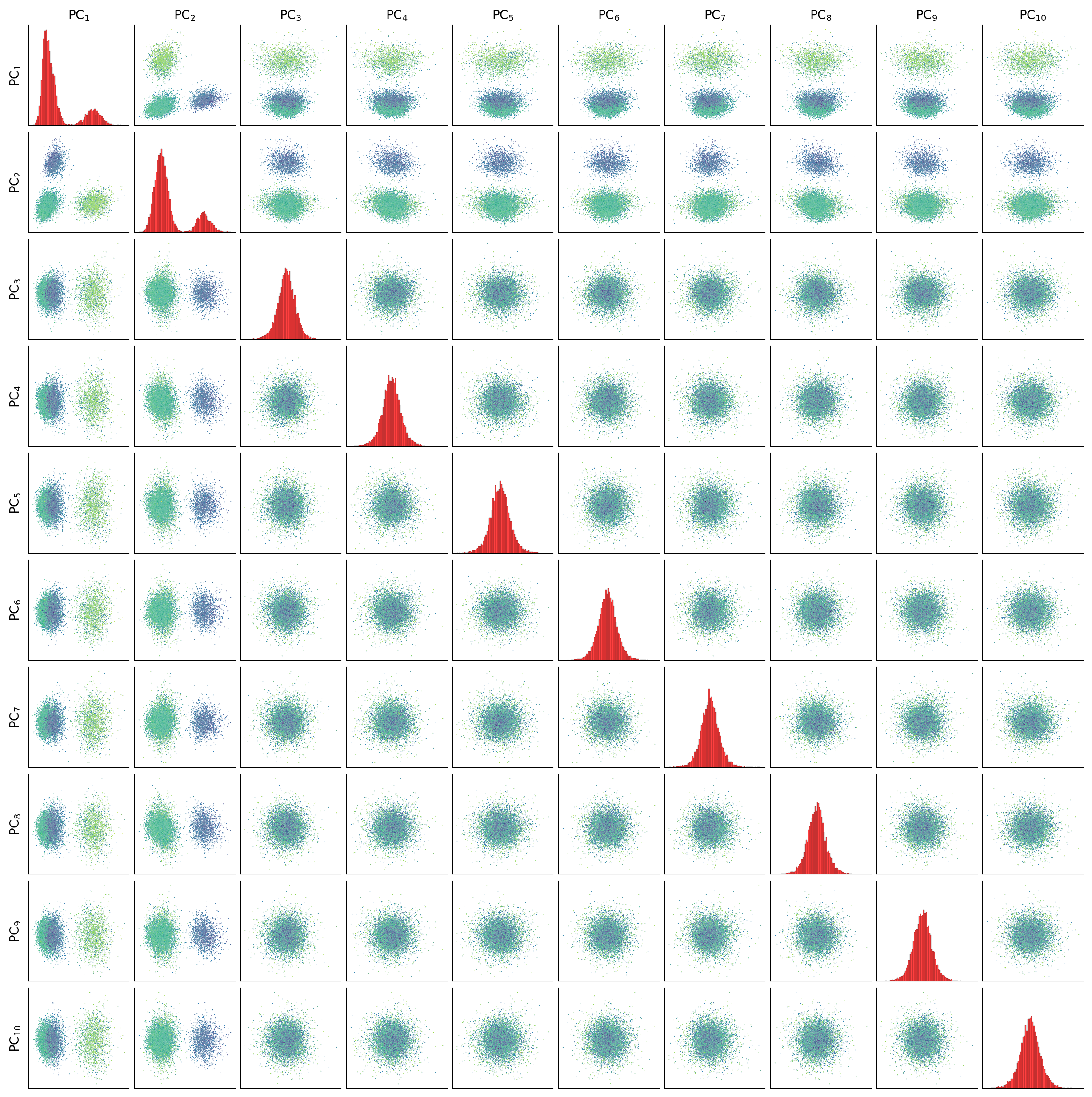}
\end{center}
\caption{Real-world $\mu$-CT data: 
Pairwise distribution of PCA components in the latent space. The diagonal plots (red histograms) depict the distribution of individual PCA components, illustrating their variability. The off-diagonal scatter plots show the pairwise relationships between PCA components, providing insights into potential correlations or clustering patterns within the latent space. 
}
\label{Fig:reallatentPCAcomponents}
\end{figure}

Figure~\ref{Fig:realLatent} presents the latent space visualization using the PCA, where data points are color-coded by $n^F$. The latent space exhibits three distinct clusters, corresponding to 0\%, (15,\,25,\,35)\%, and 50\% compression states, see, Figure~\ref{Fig:SampleCompressCT}.
Microstructures within these compression states are mapped into similar probabilistic distribution regions, leading to higher-density areas in the histogram distributions. This clustering behavior aligns with the dataset analysis shown in Figure~\ref{Fig:LBM_illustraion_KS_Diag}, where both porosity and effective permeability exhibit only slight variations across these compression levels. This indicates a strong correlation between the latent space organization and the physical properties of the microstructures.
Figure~\ref{Fig:reallatentPCAcomponents} illustrates the first 10 principal components of the latent space with points color-coded by  $n^F$. The diagonal histograms show the distribution of each individual PCA component, revealing that the first and second components deviate from a Gaussian distribution, exhibiting skewness or multimodal characteristics. In contrast, the remaining components appear to follow approximately Gaussian distributions, suggesting that they capture less dominant variations in the data.

\begin{figure}[!ht]
\begin{center}
\includegraphics[width=9.0cm]{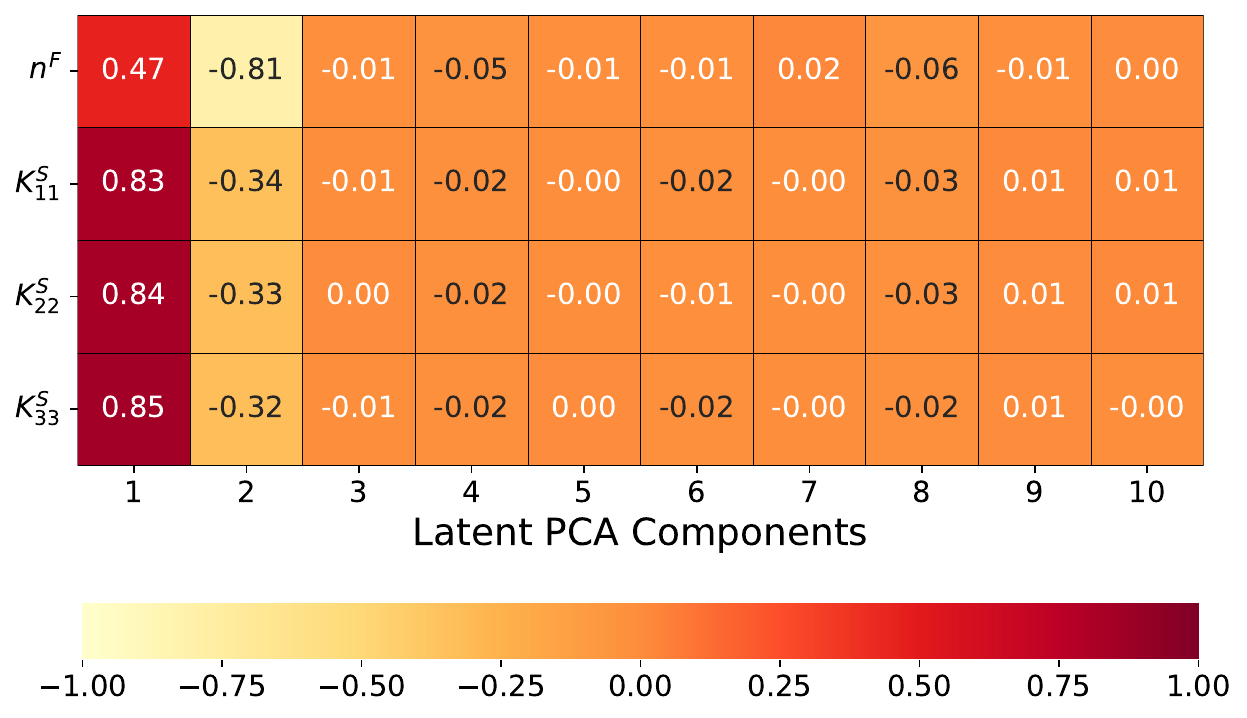}
\end{center}
\caption{Real-world $\mu-$CT data: 
Heatmap showing the Pearson correlation between the first 10 PCA components and key physical properties, including $n^F$ and intrinsic permeability principal components (\( K^S_{11}, K^S_{22}, K^S_{33} \)). Strong correlations are observed in the first two PCA components, with the first component positively correlated with permeability and moderately with porosity, while the second component is inversely correlated with porosity.
}
\label{Fig:realDiffMatrix}
\end{figure}

Figure~\ref{Fig:realDiffMatrix} presents the correlation matrix of the first 10 PCA components derived from the latent space, alongside porosity and intrinsic permeability components. The first two PCA components encode the most relevant information regarding structural and hydraulic variations. The first PCA component exhibits a strong positive correlation with intrinsic permeability ($ K^S_{11}, K^S_{22}, K^S_{33} $) and a moderate correlation with $n^F$. Conversely, the second PCA component is strongly inversely correlated with porosity ($-0.81$) and moderately inversely correlated with permeability (approximately $-0.33$). The remaining PCA components display weak correlations, suggesting that most of the property-related variations are concentrated within the first two principal components.

\begin{figure}[!ht]
\begin{center}
\includegraphics[width=4.9cm]{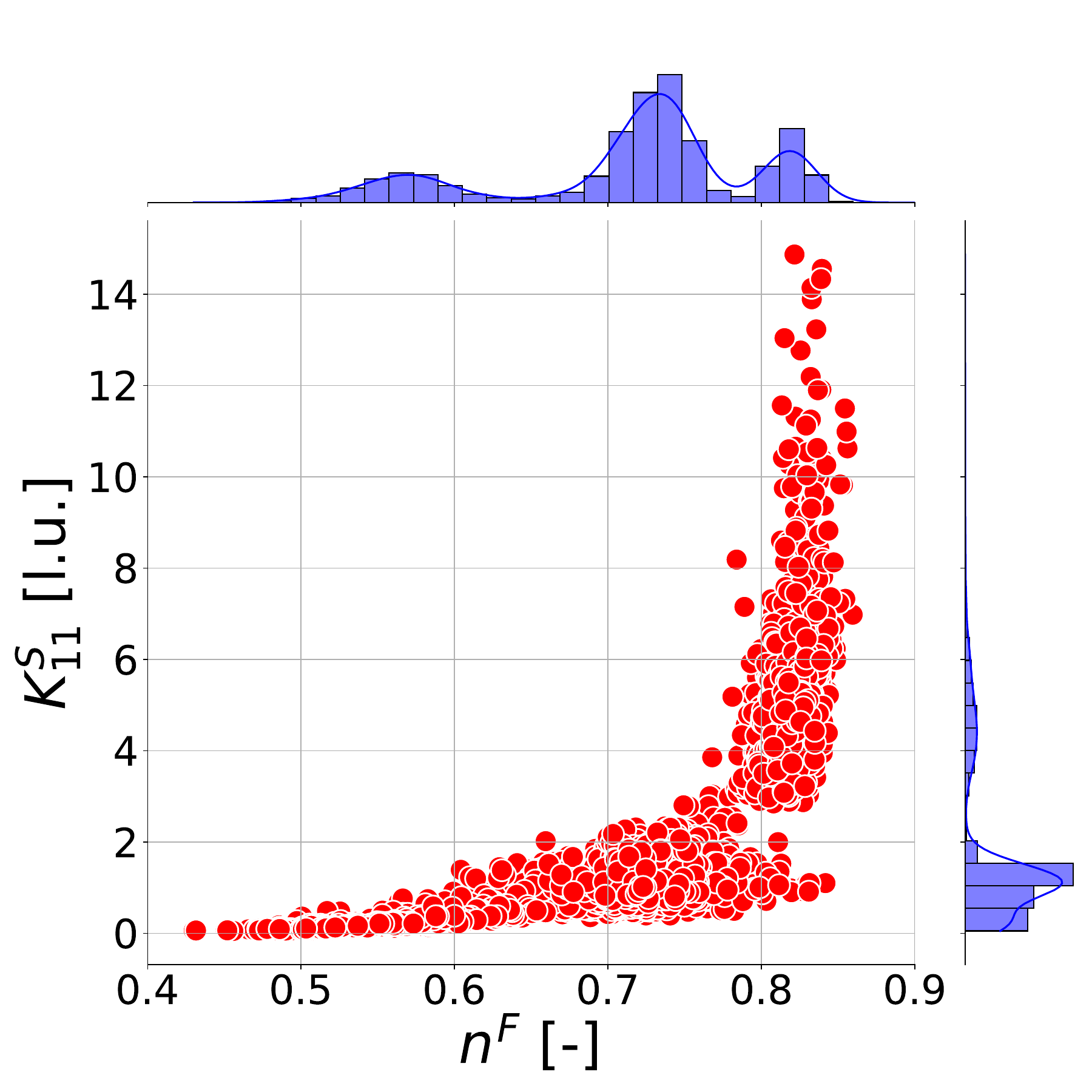}
\quad
\includegraphics[width=4.9cm]{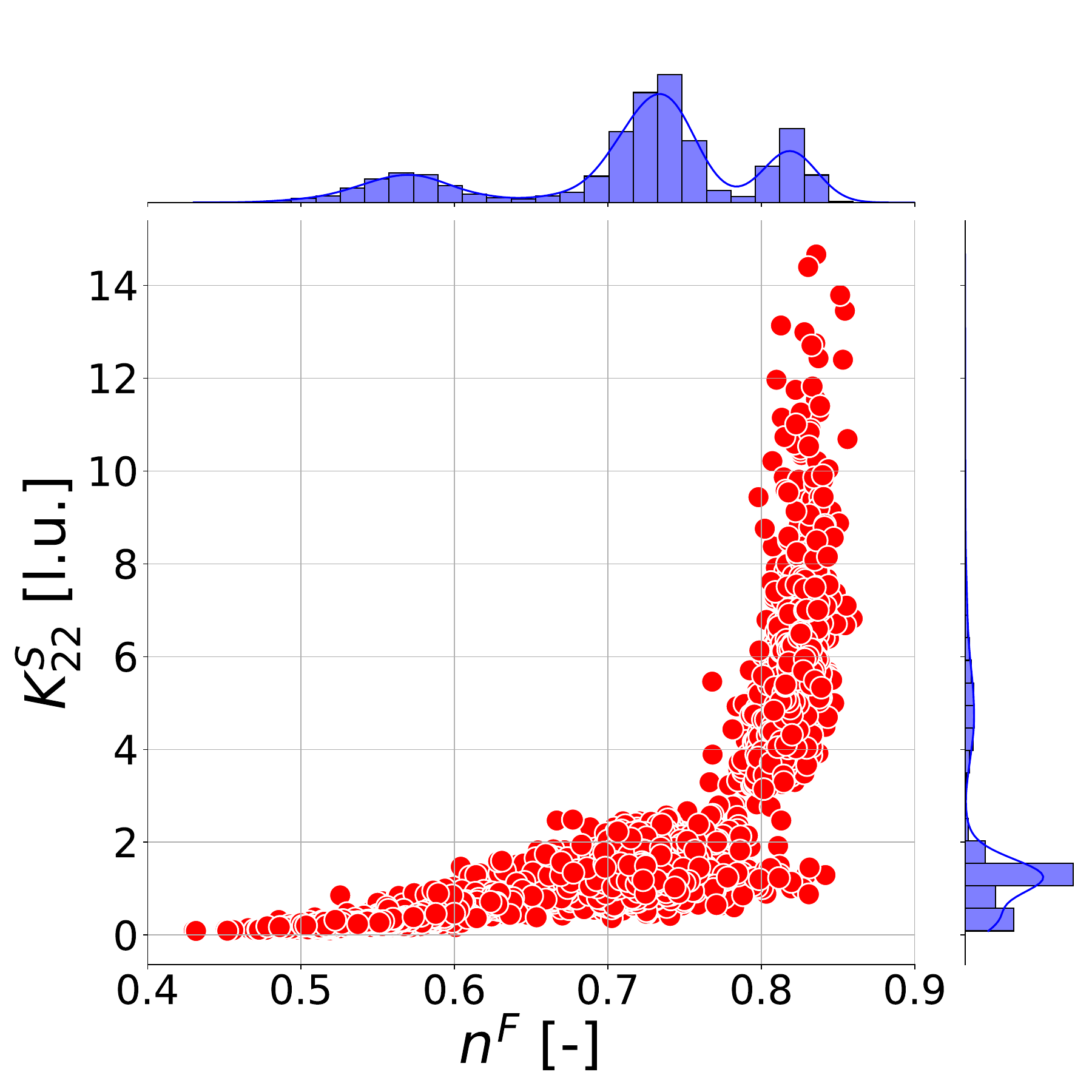}
\quad
\includegraphics[width=4.9cm]{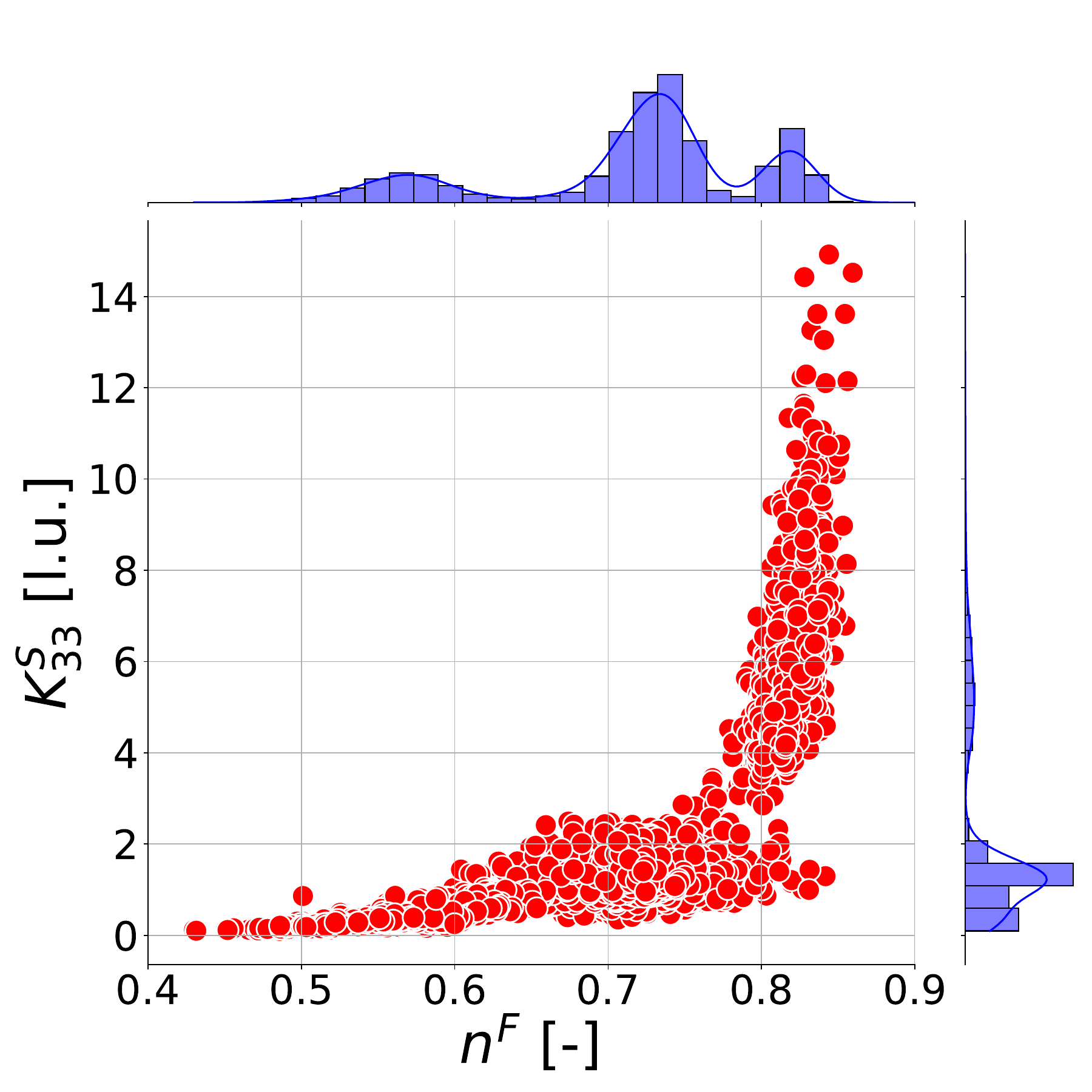}
\end{center}
\caption{Real-world $\mu-$CT data: Scatter plots illustrating the correlation between   and the three principal components of intrinsic permeability ($ K^S_{11}, K^S_{22}, K^S_{33} $). Each plot shows a non-linear relationship, where permeability exhibits an exponential-like increase at higher porosity values.}
\label{Fig:Sampling_8965_Ks}
\end{figure}

This behavior can be explained based on the dataset characteristics.
The scatter plots in Figure~\ref{Fig:Sampling_8965_Ks} illustrate a nonlinear relationship between porosity ($n^F$) and the permeability components ($ K^S_{11}, K^S_{22}, K^S_{33} $). Specifically, permeability remains relatively low for lower porosity values, following a linear trend for $n^F \lesssim 0.7$, but increases sharply beyond this critical threshold ($ n^F \approx 0.7 $), exhibiting a nonlinear correlation as presented in Figure~\ref{Fig:realDiffMatrix}.

\begin{figure}[!ht]
\begin{center}
\includegraphics[width=13.0cm]{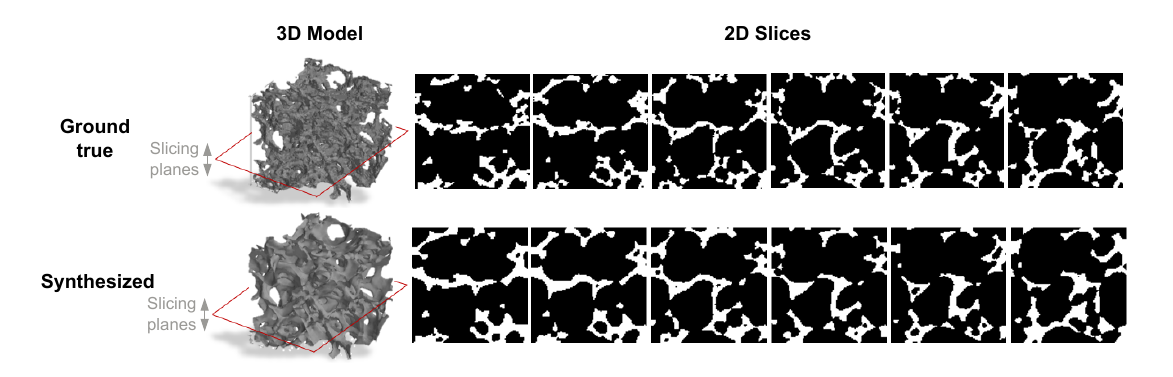}
\includegraphics[width=15.0cm]{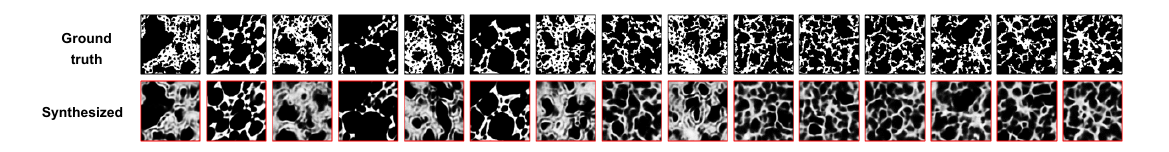}
\end{center}
\caption{Real-world $\mu$-CT data: Example of reconstruction of porous microstructures from latent space sampling. The figure compares ground truth microstructures with their synthesized counterparts generated by the VAE.}
The top row presents 3D reconstructions alongside 2D slices, while the lower rows display additional samples. 
\label{Fig:realpriorSampling}
\end{figure}

Random sampling in the latent space is conducted to evaluate the reconstruction ability of the VAE. Figure~\ref{Fig:realpriorSampling} showcases examples of randomly sampled porous microstructures alongside their synthetic reconstructions, demonstrating that the VAE can closely replicate patterns from the training dataset.
The ground truth data used for training the VAE model is already binarized. In contrast, the synthesized reconstructions generated by the VAE produce values ranging from 0 to 1, using the SteepSigmoid function. While the VAE effectively captures the overall structure of the microstructure, it tends to smooth the boundaries of the solid phase (white voxels), introducing noise (blurred regions) in the transition zones between the solid and void phases. To ensure a fully discrete two-phase structure, a binarization technique must be applied.

In conclusion, the results in this section suggest that while the latent space successfully encodes essential microstructural features, it exhibits clustering and lacks fully continuous and smooth transitions, likely due to the limited diversity and quantity of training data. Nevertheless, the generative deep learning framework demonstrates the ability to reconstruct critical characteristics of porous microstructures. To further analyze the representation quality of the latent space and its interpolation capabilities, additional evaluations are presented in the following section.
\subsection{Interpolation in the latent space}
\label{subsec:Real-interpolation}

\begin{figure}[!ht]
\begin{center}
\includegraphics[width=5.7cm]{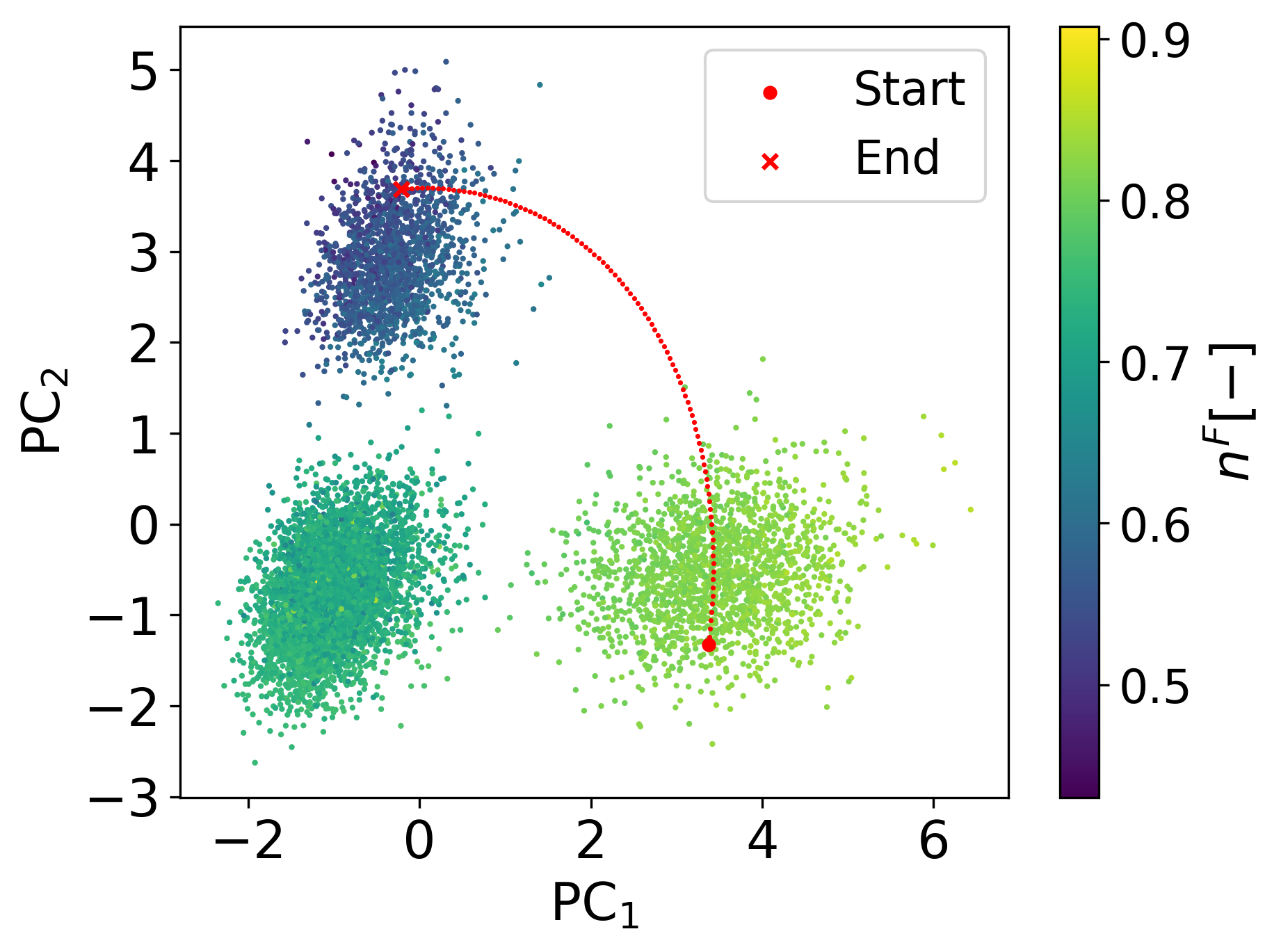}~~~
\includegraphics[width=7.0cm]{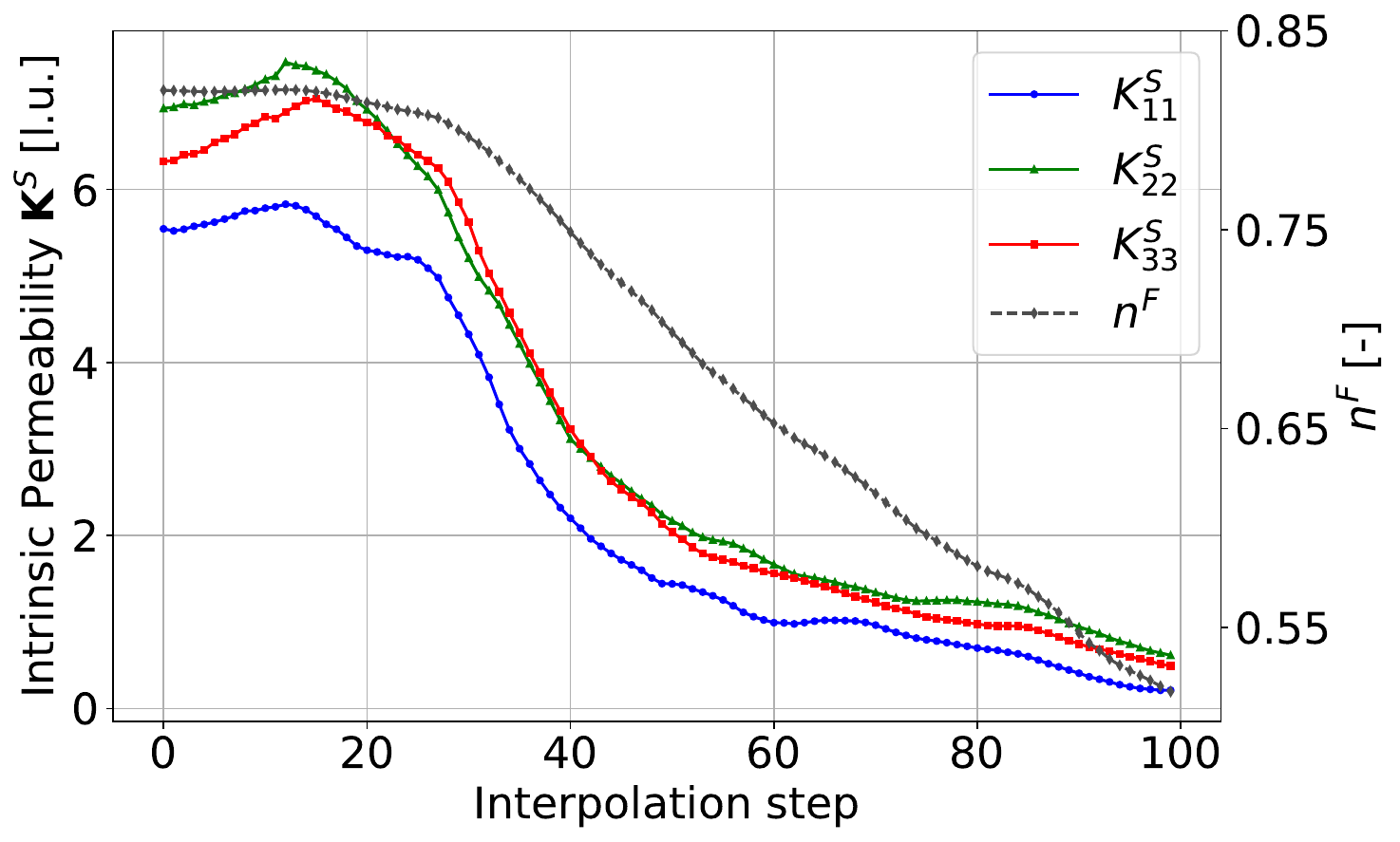}
\end{center}
    \caption{
    Real-world $\mu$-CT data: {\it Slerp} in Latent Space. The left plot shows the latent space representation using the first two principal components (PC1 \& PC2), with data points color-coded by $n^F$. A trajectory (red curve) represents the interpolation path between two latent points (start and end). 
    The right figure shows the evolution of $n^F$, $K^{S}_{11}$, $K^{S}_{22}$, and $K^{S}_{33}$ along the interpolation path in latent space.  
    }
\label{Fig:realLatentInterpolation}
\end{figure}

This section explores interpolation in the latent space using ({\it slerp}) between two distinct microstructures, one with high porosity and one with low porosity. The interpolation consists of 100 steps, smoothly transitioning between the two extremes. The trajectory in the latent space is visualized using PCA projection, as shown in Figure~\ref{Fig:realLatentInterpolation}, left. 
To assess the impact of latent space interpolation on material properties, all 100 generated microstructures are evaluated using TPM-LBM simulations (see, Appendices~\ref{appx:OneFluid_LBMTheor} and \ref{appx:InrinsicPermComputation}) to compute the intrinsic permeability tensor. Figure~\ref{Fig:realLatentInterpolation}, right, presents the variation of permeability components \(K^{S}_{11}\), \(K^{S}_{22}\), and \(K^{S}_{33}\), alongside thee porosity \( n^F \) over the interpolation steps. The results indicate a gradual and smooth decrease in permeability and porosity, confirming that interpolated microstructures follow a physically meaningful evolution.
The continuous path demonstrates the pVAE's ability to generate realistic microstructures even in less densely populated regions (dead zones) of the latent space.

\begin{figure}[!ht]
\begin{center}
\includegraphics[width=15.5cm]{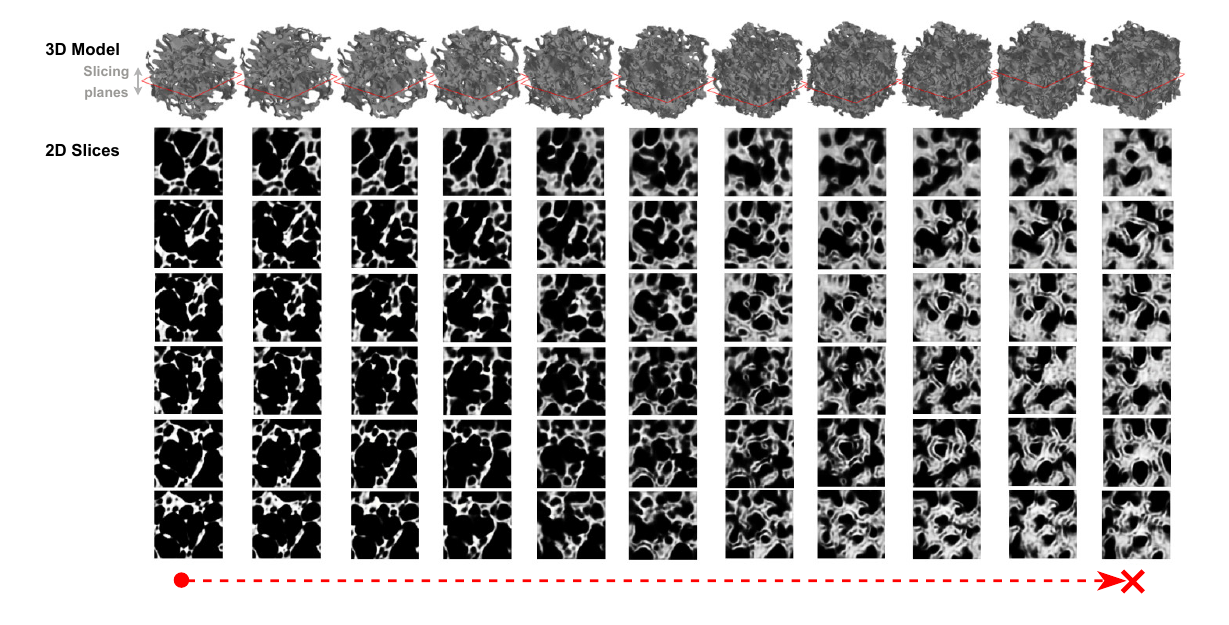}
\end{center}
    \caption{
    Generated 3D microstructures along the interpolation path. The top row displays the 3D porous geometries at selected interpolation steps, while the bottom rows show corresponding 2D slices. The gradual transition highlights the pVAE's capability to generate smooth, continuous variations in porous structures.}
\label{Fig:realInterpolationMicrostructure}
\end{figure}

A subset of 11 representative 3D microstructures along the interpolation path is displayed in Figure~\ref{Fig:realInterpolationMicrostructure}, illustrating the gradual transformation of porous structures. This confirms that the pVAE effectively captures the smooth transition between microstructures without abrupt artifacts.

\begin{figure}[!ht]
\begin{center}
\includegraphics[width=4.8cm]{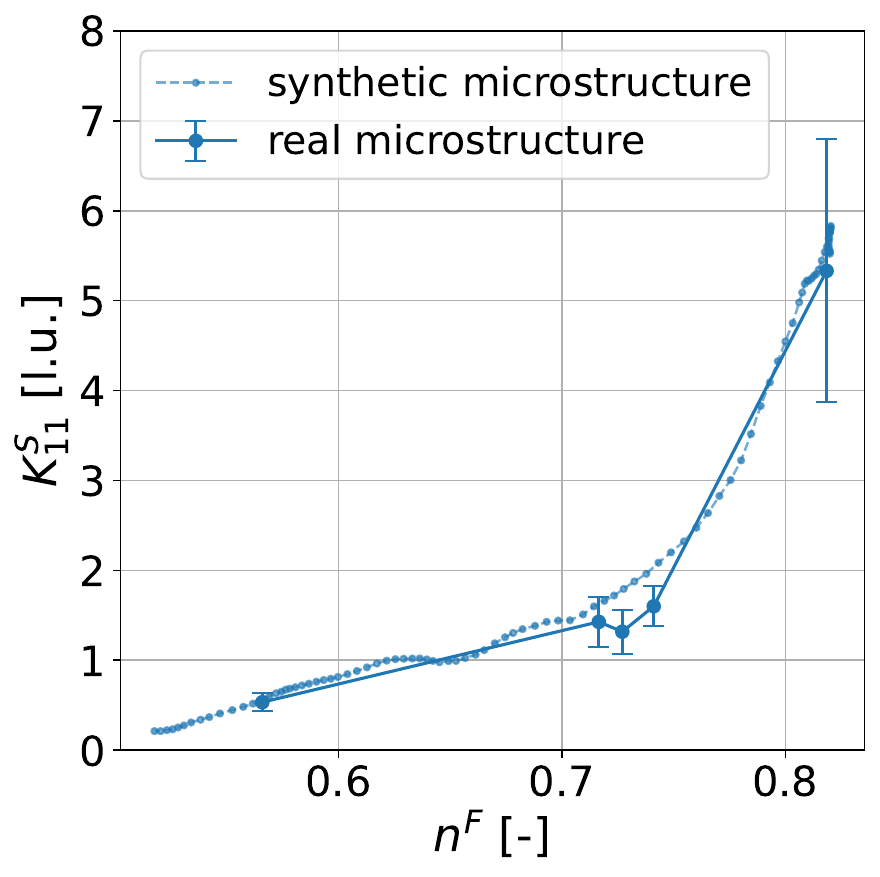}~~~
\includegraphics[width=4.8cm]{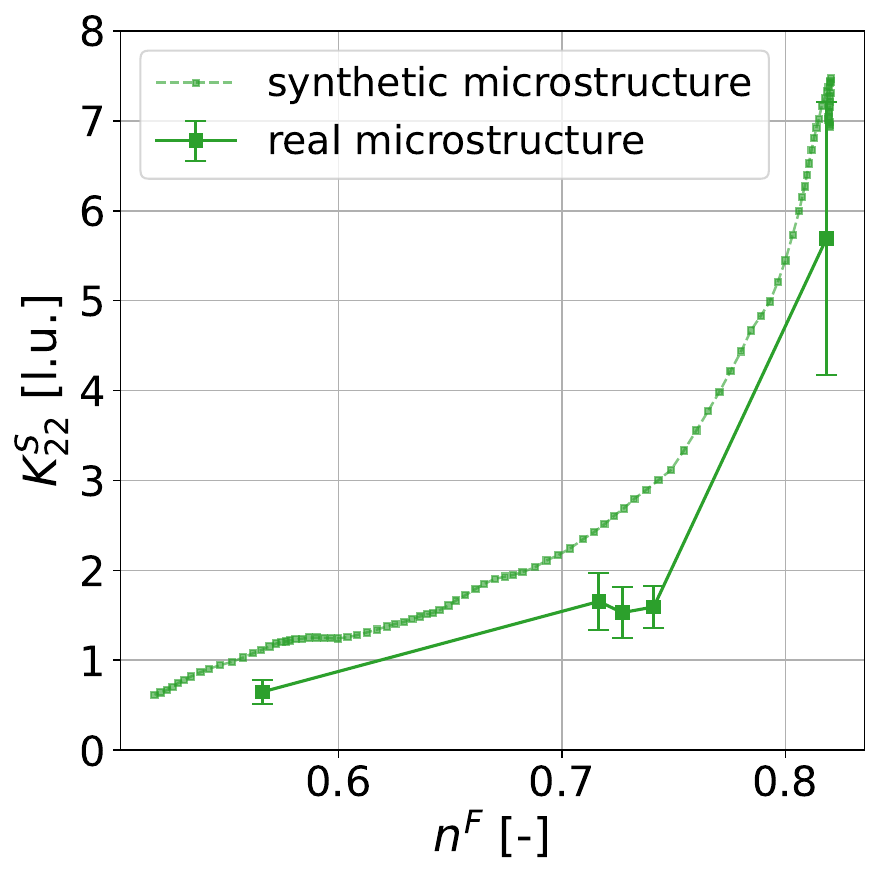}~~~
\includegraphics[width=4.8cm]{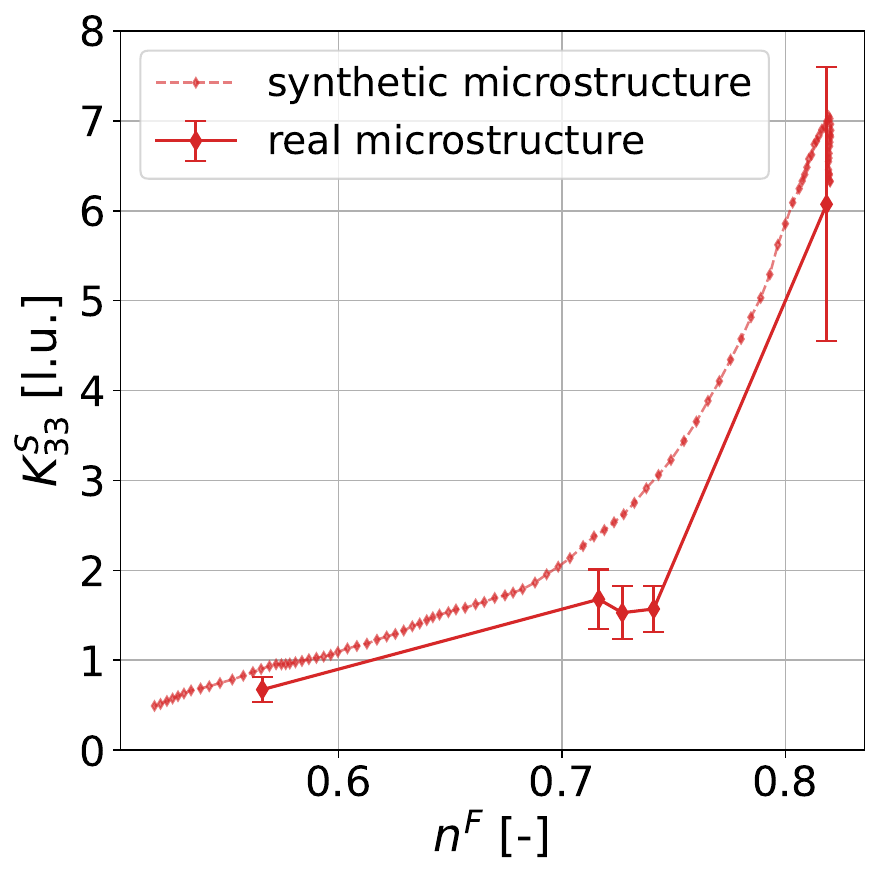}
\end{center}
\caption{
Comparison between synthetic and real microstructures of the intrinsic permeability components as functions of $n^F$. The dashed lines represent the permeability values from synthetic microstructures obtained using $slerp$ from Figure~\ref{Fig:realLatentInterpolation}. The solid lines with error bars indicate the mean and standard deviation of permeability from real microstructures of the training dataset according to the mean porosity.}
\label{Fig:ComparisionofRealKSvsIntKs}
\end{figure}

Figure~\ref{Fig:ComparisionofRealKSvsIntKs} presents a comparison of intrinsic permeability in each direction between synthetic microstructures generated from $slerp$ as shown in Figure~\ref{Fig:realLatentInterpolation} and real microstructures from CT scan. The average intrinsic permeability and porosity of whole dataset used to plot the solid line of the real microstructures are computed from the training dataset as the classical homogenization technique. The dashed and solid curves exhibit similar trends with respect to porosity, indicating consistency between the synthetic and real data. The proposed pVAE effectively captures the relationship between intrinsic permeability and porosity under compression, even when passing through the "dead zone" in the latent space.

%

\section{Conclusions and future aspects}
\label{sec:Conclusions}
This study developed a property-variational autoencoder (pVAE) framework and applied it to both a synthetic academic dataset and a real-world $\mu$-CT dataset in order to explore the generative framework and inverse design of porous metamaterials.
A convolutional neural network (CNN) model is applied to predict flow properties directly from image data, including data generated via pVAE.
The results demonstrate that the pVAE effectively captures the underlying microstructural features, forming a meaningful latent space that facilitates both interpolation and extrapolation. This enables the generation of microstructures beyond the training dataset, showcasing the framework's potential for inverse design and material discovery.

For the synthetic academic dataset, the latent space exhibited smoothness, continuity, and interpretability, making it a valuable tool for exploring microstructural designs. 
The inverse design framework, driven by effective permeability, demonstrated its potential to generate microstructures with target properties. 
However, the pVAE in this dataset is biased toward permeability prediction, as reflected by the higher $R^2$. This likely results from the model’s stronger ability to encode and reconstruct connectivity-driven features in the latent space, which are more directly related to permeability than to porosity. Additionally, structural variations with similar porosity but different flow paths add uncertainty to porosity prediction, making it a more ill-posed problem.
For the real-world $\mu$-CT dataset, the pVAE effectively reconstructed microstructures, forming a clustered latent space that reflected the dataset’s limited size and variability. Despite these constraints, spherical linear interpolation (\textit{slerp}) demonstrated the model’s capacity to generate microstructures beyond the training distribution, indicating its potential for extrapolation. This highlights the pVAE’s promise for future applications in inverse design, particularly when combined with larger and more diverse datasets to enhance its generalization ability. 
Future research will focus on expanding the design space for porous metamaterials by developing a more comprehensive dataset, particularly incorporating heterogeneous and complex-channel microstructures. A key priority will be integrating diverse material properties, including mechanical characteristics, to better capture the sensitivity of structure-property relationships. Additionally, embedding physics-based constraints within the pVAE framework will be investigated to enhance structure-property mapping accuracy and reduce reliance on large datasets. This approach aims to improve the framework’s capability to generate realistic microstructures, address inverse design challenges, and mitigate computational costs associated with extensive data requirements.
\section*{Conflict of interest}
The authors declare that there is no conflict of interest.

\section*{Data availability statement}
The datasets and codes used in this study will be made open-access. 

\appendix

\section{LBM for single-phase fluid flow}
\label{appx:OneFluid_LBMTheor}
In this study, we employ lattice Boltzmann method (LBM) simulations to compute the average fluid velocity within each 3D sample under an applied pressure gradient. This enables the inverse calculation of the intrinsic permeability tensor using Darcy's law, as described in \cite{chaaban2020upscaling, HeiderDakheelEhlers2024CNN}. 
A brief introduction to single-phase flow simulations using LBM is provided below, while more detailed explanations and references can be found in \cite{chaaban2020upscaling,chaaban2022_TwoPhaseLBMtpm}.

The LBM employs a grid-based approach to solve the Boltzmann equation \cite{boltzmann2012lectures}. It begins by defining the velocity distribution function $ f(\mathbf{x}, \boldsymbol{\xi}, t)$, which represents the probability of finding a fluid particle at a given position $\mathbf{x}$ and time $t$ with a discrete velocity $\boldsymbol{\xi}$. The Boltzmann equation then governs the evolution of $f(\mathbf{x}, \boldsymbol{\xi}, t)$ in both space and time. 
This evolution, driven by the exchange of momentum and energy among fluid particles, occurs through two key processes; streaming and collision as
\begin{equation} 
\label{Boltzmann}
\dfrac{df}{dt} \Big|_{\text{streaming}} = \dfrac{df}{dt} \Big|_{\text{collision}}\,\quad\text{with}\quad\quad
\underbrace{\dfrac{\partial f}{\partial t} + 
\Vxi \cdot \dfrac{\partial f}{\partial \mathbf{x}}}_{\text{streaming operator}} = \underbrace{\vphantom {\dfrac{\partial f}{\partial t} + \mathbf{v} \cdot \dfrac{\partial f}{\partial \mathbf{x}}} \Omega \, (f)\,.}_{\text{collision operator}}
\end{equation}
As described in \citet{kruger2017lattice}, the distribution function $f(\mathbf{x}, \boldsymbol{\xi}, t)$ is related to macroscopic variables such as the fluid density $\rho^\FR$ and the fluid velocity $ \mathbf{v}_F$ through its moments. This relationship is established using the following integrals: 
%
\begin{equation}
\begin{array}{l}
\Ds
\rho^\FR(\mathbf{x},t)\approx\rho_l\,(\mathbf{x},t) = \int{f(\mathbf{x},\Vxi,t)} \, d\Vxi \quad \text{and} \quad
\Vv_F(\mathbf{x},t)\approx\Vu_l\,(\mathbf{x},t) = \dfrac{1}{\rho_l}\int{\Vxi\,f(\mathbf{x},\Vxi,t)} \, d\Vxi\,.
\end{array}
\end{equation}

%
For spatial discretization in three dimensions, a fluid particle can propagate along 19 discrete velocity directions, a scheme known as D3Q19. These directions are defined as:
\begin{eqnarray}
\mathbf{e}_i = \left\{\begin{array}{lcl}
(0,0,0) & \quad & i=0 \\
(\pm1,0,0), (0,\pm1,0), (0,0,\pm1) & \quad & i=1, 2,\dotsc, 6\\
(\pm1,\pm1,0), (\pm1,0,\pm1), (0,\pm1,\pm1) & \quad & i=7, 8,\dotsc, 18, \\
\end{array}\right.
\end{eqnarray}
where $\mathbf{e}_i$ is the direction of the velocity vectors $\Vxi_i = c \, \mathbf{e}_i\,$ given in terms of $c$ as the ratio of the distance between the nodes $\Delta x$ to the time-step size $\Delta \,t$\,.
Regarding the collision operator $\Omega \, (f)$ in (\ref{Boltzmann}), the Bhatnagar-Gross-Krook (BGK) \cite{PhysRev.94.511} model is used since it is easy to implement and has been widely used in LBM fluid flow simulation \cite{wolf2004lattice}. 
In particular, the BGK collision operator $\Omega_{\text{BGK}}$ is expressed as
\begin{eqnarray}
\label{eq_BGK}
\Omega_{\text{BGK}} = - \frac{f_i - f_i^{eq}}{\tau}  \quad \text{with} \quad \tau := \frac{1}{2} + \nu_l \, c_s^{-2}\,.
\end{eqnarray}
Herein, the relaxation time $\tau$ depends on the lattice fluid viscosity $\nu_l$ and lattice speed of sound $c_s=1/\sqrt{3}$\,. 
The BGK model facilitates the relaxation of the distribution functions $f_i$ toward equilibrium distributions $f_i^{eq}$ at a collision frequency $\tau^{-1}$. The formulation of $f_i^{eq}$ is expressed as follows
\begin{eqnarray}
\label{eq_equil}
f_i^{eq} = w_i \, \rho_l \, \bigg(1 + \dfrac{\mathbf{e}_i \cdot \mathbf{u}_l}{{c_s}^2} + \dfrac{(\mathbf{e}_i \cdot \mathbf{u}_l)^2}{2{c_s}^4} - \dfrac{(\mathbf{u}_l \cdot \mathbf{u}_l)^2}{2{c_s}^2}\bigg)\,,
\end{eqnarray}
where $w_i$ presents the lattice weights, i.e.
\begin{eqnarray}
w_i = \left\{\begin{array}{lcl}
1/3 & \quad & i=0 \\
1/18 & \quad & i=1, 2,\dotsc, 6 \\
1/36 & \quad & i=7, 8,\dotsc, 18\,.
\end{array}\right.
\end{eqnarray}
The distribution functions are updated through the following equation:
\begin{eqnarray}
\underbrace{f_i(\mathbf{x} + \Vxi_i \Delta t, t + \Delta t) - f_i(\mathbf{x}, t)}_{\text{streaming}} = \underbrace{\Omega_{\text{BGK}}}_{\text{collision}}.
\end{eqnarray}
 %

The time integration scheme of the Lattice Boltzmann Method (LBM) consists of two main steps performed explicitly at each time step: collision and streaming.

\begin{enumerate}
    \item {\textbf{Collision step:} The particle distribution functions \( f_i(\mathbf{x}, t) \) at each lattice node are relaxed towards equilibrium \( f_i^{eq}(\mathbf{x}, t) \) using the BGK operator:
    \begin{equation}
    f_i^*(\mathbf{x}, t) = f_i(\mathbf{x}, t) - \frac{1}{\tau} \left( f_i(\mathbf{x}, t) - f_i^{eq}(\mathbf{x}, t) \right),
    \label{eq:collision}
    \end{equation}
    where \( \tau \) is the relaxation time related to the fluid viscosity.}
    
    \item \textbf{Streaming step:} The post-collision distributions \( f_i^* \) propagate to neighboring lattice nodes along discrete velocities \( \mathbf{c}_i \):
    \begin{equation}
    f_i(\mathbf{x} + \mathbf{c}_i \Delta t, t + \Delta t) = f_i^*(\mathbf{x}, t).
    \label{eq:streaming}
    \end{equation}
\end{enumerate}

The time step \( \Delta t \) and lattice spacing \( \Delta x \) define the lattice velocity scale \( c = \frac{\Delta x}{\Delta t} \). This explicit scheme is first-order accurate in time and widely used for its computational efficiency and stability.

For the boundary conditions (BCs), the Zou-He bounce-back scheme \cite{zou1997pressure} is employed. For a more detailed explanation of the LBM approach for single-phase fluid flow, refer to \cite{chaaban2020upscaling}.

\section{Intrinsic permeability computation}
\label{appx:InrinsicPermComputation}


The proposed pVAE model in this study takes CT images of viscoelastic foam as input and outputs the components of the permeability tensor.  
To incorporate intrinsic permeability components into the database, 3D samples of viscoelastic foam at varying deformation levels are processed using a single-phase LBM solver.  
The primary goal of the LBM simulations is to calculate the average lattice fluid velocity for each prescribed pressure gradient applied across the porous domain along the hydrodynamic axes $\mathbf{x}_1,\,\mathbf{x}_2$ and $\mathbf{x}_3\,$, i.e. $\nabla p_1, \,\nabla p_2 \, \text{and} \, \nabla p_3\,$, respectively. 
The simulation results are used to determine the lattice intrinsic permeability tensor
$\mathbf{K}^S_l$ in lattice units \(\big[\mathrm{l.u.}\big]\). This is done under the assumption that the permeability tensor is symmetric and positive definite \cite{kuhn2015stress}. 

As proposed by \citet{kuhn2015stress}, two fluid flow simulations are performed for each direction using different boundary conditions:  
one with no-slip boundary conditions and another with natural slip boundary conditions on surfaces parallel to the fluid flow direction (see Fig.~\ref{Fig:LBM_illustraion_KS_Diag}, left, for illustration). The rationale behind this approach is that the average velocity in the pressure gradient direction is lower with no-slip boundary conditions compared to natural slip boundary conditions. The difference between these velocities reveals additional fluid flow in the direction orthogonal to the applied pressure gradient.  
The average velocities obtained under no-slip boundary conditions are then used to compute the diagonal components of the permeability tensor based on 1D Darcy’s filtration law as
\begin{equation}
\label{eqnDarcy}
(K_l^S)_{ii}  = 
- \nu_l \, \frac{( u_{l})_{i,\,\text{avg}}}{\nabla\!_i \, p_l}\quad\big[\mathrm{l.u.}\big]\,, \quad \mathrm{with} \quad i=1,\,2,\,3.
\end{equation}
In this, the lattice pressure gradient, presented by $\nabla\!_i \, p_l = \partial p_l / \partial x_i$\, is induced between two opposing surfaces perpendicular to the flow direction to calculate the average lattice fluid velocity $(u_{l})_{i,\,\text{avg}}$\,. 
As for the latter, the unknown non-diagonal elements of the permeability tensor are computed using the average velocity with natural slip boundary conditions following \cite{kuhn2015stress} as
\begin{eqnarray} 
\label{eqnNonDiag}
\begin{bmatrix}
	\nabla\!_2 \, p_l & \nabla\!_3 \, p_l & 0 \\[1mm]
	\nabla\!_1 \, p_l & 0 & \nabla\!_3 \, p_l \\[1mm]
	0 & \nabla\!_1 \, p_l & \nabla\!_2 \, p_l 
\end{bmatrix}
\begin{bmatrix}
(K_l^S)_{12} \\[1mm] (K_l^S)_{13} \\[1mm] (K_l^S)_{23} 
\end{bmatrix}
=
\begin{bmatrix}
- \nu_l \, ( u_{l})_{1,\,\text{avg}} - (K_l^S)_{11} \, \nabla\!_1 \, p_l \\[1mm] 
- \nu_l \, ( u_{l})_{2,\,\text{avg}} - (K_l^S)_{22} \, \nabla\!_2 \, p_l \\[1mm] 
- \nu_l \, ( u_{l})_{3,\,\text{avg}} - (K_l^S)_{33} \, \nabla\!_3 \, p_l
\end{bmatrix}.
\end{eqnarray}
The non-diagonal components computed for viscoelastic foam are much smaller than the diagonal components. Thus, we neglect them for simplicity from the ML model, see, \cite{PhuEtAl2023_PAMM}.
In particular, the symmetric permeability tensor and its simplified diagonal form are expressed as follows
\begin{equation}
    \TK_l^S =
    \begin{bmatrix}
	(K_l^S)_{11} \;& (K_l^S)_{12} \;& (K_l^S)_{13} \\[1mm]
	(K_l^S)_{21} \;& (K_l^S)_{22} \;& (K_l^S)_{23} \\[1mm]
	(K_l^S)_{31} \;& (K_l^S)_{32} \;&(K_l^S)_{33}
    \end{bmatrix}(\overline{\Ve}_i\otimes\overline{\Ve}_j)
    \approx
    \begin{bmatrix}
	(K_l^S)_{11} & 0 & 0 \\[1mm]
	0 & (K_l^S)_{22} & 0 \\[1mm]
	0 & 0 &(K_l^S)_{33}
    \end{bmatrix}(\overline{\Ve}_i\otimes\overline{\Ve}_j)\,.
\end{equation}
Here, $\overline{\Ve}_i$, $\overline{\Ve}_j$ represent the cartesian basis vectors with $i ,j\in\{1,2,3\}$ and $\otimes$ is the dyadic product (tensor product).
The macroscopic intrinsic permeability tensor is derived from the lattice permeability tensor as follows
\begin{equation}\label{eqnConversion}
K_{ij}^S = (K_l^S)_{ij} \, (\delta x_i\,\delta x_j) \quad\text{in}\quad \big[\mathrm{m}^2\big] \,,
\end{equation}
where $\delta x_i$ and $\delta x_j$ characterize the spatial resolution of the $\bm{\mu}$-CT images in the $i$ and $j$ directions, respectively. 
\begin{acknowledgements}
The authors gratefully acknowledge financial support from the {\it International Research Training Group (IRTG) 2657}: Computational Mechanics Techniques in High Dimensions, funded by the German Research Foundation (DFG) with Project-ID 433082294.  
\end{acknowledgements}

%
%

\bibliographystyle{spbasic}      
\bibliography{main}
%
%

\end{document}